\documentclass[11pt]{article}
\usepackage[table]{xcolor}

\usepackage{acl}

\usepackage{times}
\usepackage{latexsym}
\usepackage[T1]{fontenc}

\usepackage[utf8]{inputenc}

\usepackage{microtype}
\usepackage{pgfplotstable} 
\usepackage{collcell} 
\usepackage{inconsolata}
\usepackage{graphicx}
\usepackage{subcaption}
\usepackage{CJK}
\usepackage{amsmath}
\usepackage{multirow}
\usepackage{listings}
\usepackage{float}
\usepackage{geometry}
\usepackage{caption}
\usepackage{tabularx}
\usepackage{amssymb}
\usepackage{booktabs}
\usepackage{arydshln}
\usepackage[absolute]{textpos}
\usepackage{enumitem}
\setitemize{itemsep=0pt,parsep=0pt,topsep=0pt,partopsep=0pt}

\usepackage{scalerel}
\usepackage{cancel}
\usepackage{colortbl}
\usepackage{moresize}

\title{\includegraphics[height=1em]{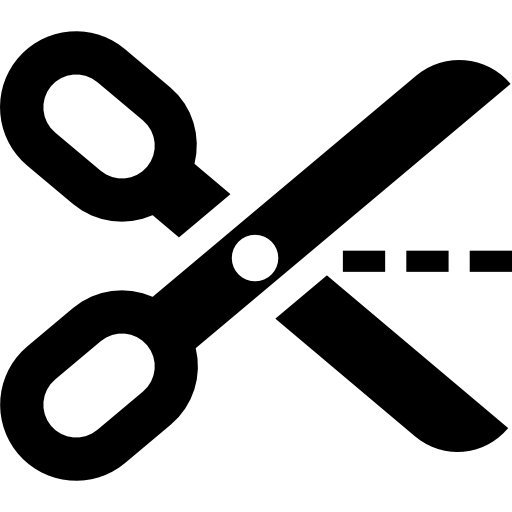} \includegraphics[height=1em]{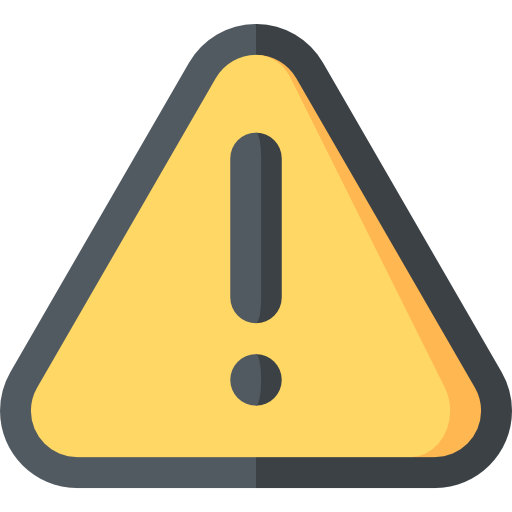} 
 Cutting Off the Head Ends the Conflict: A Mechanism for\\ Interpreting and Mitigating Knowledge Conflicts in Language Models}

\author{
\textbf{Zhuoran Jin}$^{1, 2}$, 
\textbf{Pengfei Cao}$^{1, 2}$,
\textbf{Hongbang Yuan}$^{1, 2}$,
\textbf{Yubo Chen}$^{1, 2}$,\\
\textbf{Jiexin Xu}$^{3}$\textbf{,}
\textbf{Huaijun Li}$^{3}$\textbf{,}
\textbf{Xiaojian Jiang}$^{3}$\textbf{,}
\textbf{Kang Liu}$^{1, 2}$\textbf{,}
\textbf{Jun Zhao}$^{1, 2}$\\
$^1$  School of Artificial Intelligence, University of Chinese Academy of Sciences, Beijing, China\\
$^2$ The Laboratory of Cognition and Decision Intelligence for Complex Systems,\\
Institute of Automation, Chinese Academy of Sciences, Beijing, China \\
$^3$ China Merchants Bank\\
\texttt{\{zhuoran.jin, pengfei.cao, yubo.chen, kliu, jzhao\}@nlpr.ia.ac.cn}\\
}

\begin{document}
\maketitle
\begin{abstract}
Recently, retrieval augmentation and tool augmentation have demonstrated a remarkable capability to expand the \textbf{internal memory} boundaries of language models (LMs) by providing \textbf{external context}.
However, internal memory and external context inevitably clash, leading to \textbf{knowledge conflicts} within LMs.
In this paper, we aim to interpret the mechanism of knowledge conflicts through the lens of information flow, and then mitigate conflicts by precise interventions at the pivotal point.
We find there are some attention heads with opposite effects in the later layers, where \textbf{memory heads} can recall knowledge from internal memory, and \textbf{context heads} can retrieve knowledge from external context.
Moreover, we reveal that the pivotal point at which knowledge conflicts emerge in LMs is the integration of inconsistent information flows by memory heads and context heads.
Inspired by the insights, we propose a novel method called \textbf{P}runing \textbf{H}ead via \textbf{P}at\textbf{H} \textbf{P}atc\textbf{H}ing (\textbf{PH3}), which can efficiently mitigate knowledge conflicts by pruning conflicting attention heads without updating model parameters.
PH3 can flexibly control eight LMs to use internal memory ($\uparrow$ 44.0\%) or external context ($\uparrow$ 38.5\%).
Moreover, PH3 can also improve the performance of LMs on open-domain QA tasks.
We also conduct extensive experiments to demonstrate the cross-model, cross-relation, and cross-format generalization of our method.

\end{abstract}

\section{Introduction}

Language models (LMs) \citep{GPT3, touvron2023llama, GPT4} have memorized a substantial amount of factual knowledge during pre-training, and stored the knowledge within their parameters as \textbf{internal memory} (\textit{i.e.}, \textbf{parametric knowledge}) \citep{meng2022locating}.
During the inference phase, LMs rely on their internal memory to understand and generate text. However, the internal memory may be limited or outdated, making LMs prone to producing factually incorrect content.

To alleviate the problem, one promising solution is to employ additional retrievers or tools to augment LMs by providing \textbf{external context} (\textit{i.e.}, \textbf{non-parametric knowledge}).
Nevertheless, internal memory and external context can often contradict each other, which is known as \textbf{knowledge conflicts} \citep{longpre-etal-2021-entity, chen-etal-2022-rich, xie2023adaptive, yu-etal-2023-characterizing}.
Recent works have mainly investigated the behavior and preference of LMs, attempting to determine whether these models are more inclined towards internal memory or external context when faced with knowledge conflicts.
However, there is a limited understanding of the underlying mechanism of knowledge conflicts.
Insights into the mechanism will facilitate precise interventions at the pivotal point to mitigate knowledge conflicts, which can not only empower LMs to more reliably adhere to internal memory (\textit{e.g.}, ignoring misleading external context) but also enhance faithfulness in generating text based on external context (\textit{e.g.}, correcting outdated memory).

\begin{figure*}[t]
    \centering
    \includegraphics[clip=true,width=0.91\textwidth]{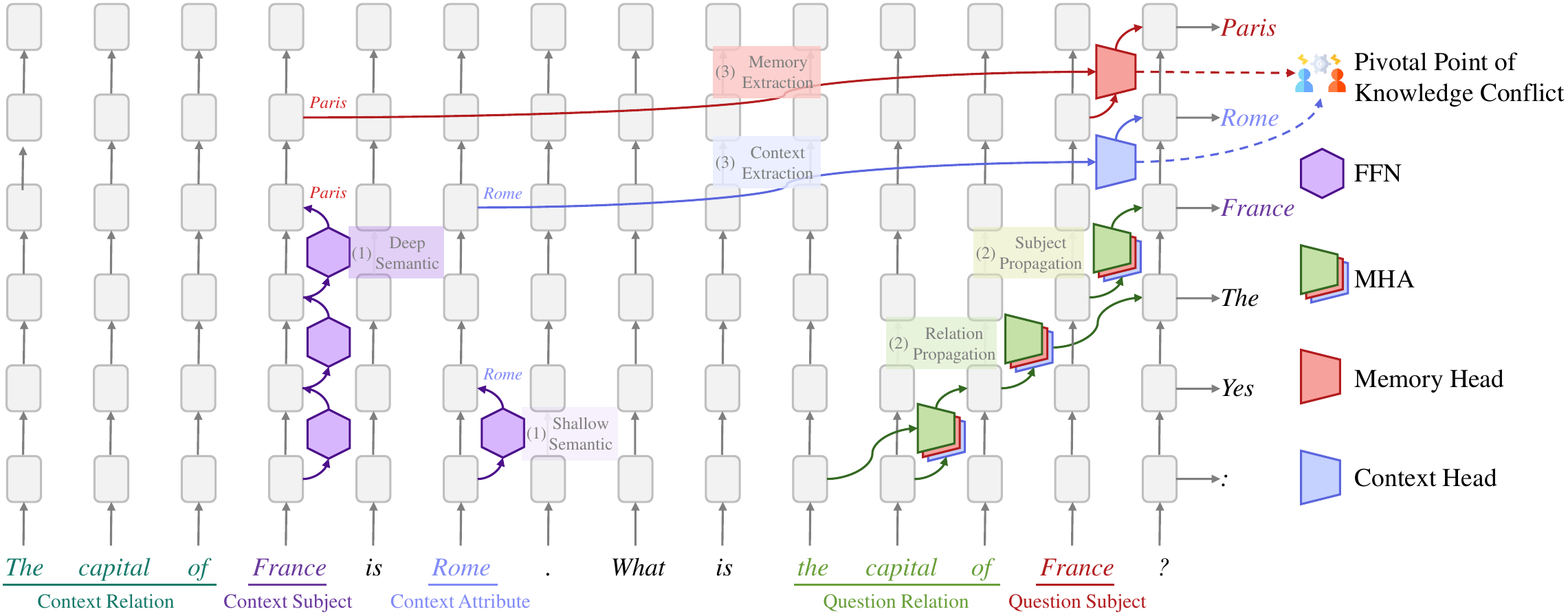}

    \caption{An illustration of the mechanism of knowledge conflicts in LMs: (1) Enriching the semantic information of context subject and context attribute; (2) Propagating question information to the last token through MHAs; (3) Extracting attribute information through memory attention heads and context attention heads at later layers. }

    \label{example}
            \vspace{-12pt}

\end{figure*}

In this paper, we reveal that the pivotal point at which knowledge conflicts emerge in LMs is the integration of inconsistent information flows by various attention heads in later layers.
To investigate this, we consider a simple factual recall task (\textit{i.e.}, subject attribute prediction) inspired by the work of \citet{yu-etal-2023-characterizing}.
As illustrated in Figure \ref{example}, given the question (\textit{i.e.}, ``\textit{What is the capital of France?}'') and the conflicting external context (\textit{i.e.}, ``\textit{The capital of France is Rome.}''), the model can either use internal memory (\textit{i.e.}, ``\textit{Paris}'') or external context (\textit{i.e.}, ``\textit{Rome}'') to predict the subject's attribute.
Following this, we present a set of ``\textit{top-down}'' analyses to locate the pivotal point where conflicts emerge and to identify the model components that are significant in knowledge conflicts, which primarily involves the following three steps:

\textbf{Step 1}: We start by answering the \textit{first} question ``\textit{What function do model components serve in knowledge conflicts?}''.
We knock out the activations to examine the functionality of \textit{multi-head attention} (MHA) blocks and \textit{feed-forward network} (FFN) blocks.
We find that FFNs enrich the semantic information of input elements in early layers, while MHAs play an important role in passing information to the last token in later layers;
\textbf{Step 2}: Based on this, the \textit{second} question naturally arises, namely ``\textit{When and where do MHAs pass information to the last token?}''.
We investigate the MHAs by knocking out the attention weights from the last token to other input elements.
Results reveal that the question information is first propagated to the last token, and then the last token extracts attribute information from the subject and the attribute in the context;
\textbf{Step 3}: Inspired by this, we aim to answer the \textit{final} question ``\textit{How do MHAs extract attribute information under knowledge conflicts?}''.
We find that some attention heads in late MHAs play opposite roles, where \textbf{memory heads} can recall attributes from internal memory, and \textbf{context heads} can retrieve attributes from external context.
According to our findings, the mechanism by which LMs use both internal memory and external context can be summarized as three stages in Figure \ref{example}: (1) Enriching semantic information; (2) Propagating question information; and (3) Extracting attribute information, where knowledge conflicts arise at the third stage, due to the inconsistent information flows between memory heads and context heads.

Inspired by our insights into knowledge conflicts, we propose a minimally-invasive control method called \textbf{P}runing \textbf{H}ead via \textbf{P}at\textbf{H} \textbf{P}atc\textbf{H}ing (\textbf{PH3}), which can efficiently mitigate knowledge conflicts by intervening on attention heads without updating model parameters.
First, we use the \textbf{path patching} \cite{goldowsky2023localizing, wang2023interpretability} technique to localize important memory heads and context heads. 
Our method can avoid the noise interference of other heads, enabling a more accurate calculation of the importance score for the target head.
Then, we perform \textbf{structured pruning} on those negative attention heads to mitigate conflicts. In this way, our method can flexibly control LMs to use internal memory or external context.
Experimental results on the \texttt{World Capital} dataset show that our method can not only reliably and consistently increase the average internal memory usage rate of eight LMs by \textbf{44.0\%} (from 49.7\% to 93.7\%) but also increase the external context usage rate by \textbf{38.5\%} (from 50.3\% to 88.8\%).
PH3 also enables LMs to generate answers more faithfully according to retrieved passages in open-domain QA tasks.
We conduct extensive experiments to demonstrate the cross-model (\textit{e.g.}, from GPT series to LLaMA2 series), cross-relation (\textit{e.g.}, from \texttt{World Capital} to \texttt{Official Language}), and cross-format (\textit{e.g.}, from triple format to document format) generalization.
Our contributions are summarized as follows:
\begin{itemize}

\item We perform an exploration into the mechanism of interpreting knowledge conflicts, and reveal that memory heads and context heads at later layers can cause knowledge conflicts when inconsistent information flows merge.

\item We propose a novel method called \textbf{P}runing \textbf{H}ead via \textbf{P}at\textbf{H} \textbf{P}atc\textbf{H}ing (\textbf{PH3}), which can efficiently mitigate knowledge conflicts by pruning those conflicting attention heads.

\item We demonstrate that our PH3 can flexibly control LMs to use internal memory ($\uparrow$ 44.0\%) or external context ($\uparrow$ 38.5\%).
We also prove the cross-model, cross-relation, and cross-format generalization ability of our method.

\end{itemize}

\section{Background}

In this work, we mainly focus on the autoregressive transformer-based language models.
Given a sequence of input tokens $x = \left[x_{1}, \cdots, x_{N}\right]$, the LM $\mathcal{G}$ first embeds each token $x_{i}$ into a vector $\mathbf{x}_i^0 \in \mathbb{R}^d$ using an embedding matrix $E \in \mathbb{R}^{|\mathcal{V}| \times d}$, over a vocabulary $\mathcal{V}$.
The input embeddings are processed by $L$ transformer layers.
Each layer consists of an MHA and an FFN. Formally, the hidden state $\mathbf{x}_i^\ell$ of token $x_i$ at layer $\ell$ is calculated as:
\begin{equation}
\mathbf{x}_i^{\ell}=\mathbf{x}_i^{\ell-1}+\mathbf{a}_i^{\ell}+\mathbf{m}_i^{\ell},
\end{equation}
where $\mathbf{a}_i^{\ell}$ and $\mathbf{m}_i^{\ell}$ are the outputs from the MHA block and the FFN block in the $\ell$-th layer.
Then, the vocabulary head $\phi(\cdot)$ and the softmax function $\sigma(\cdot)$ predict the output probability:
\begin{equation}
\mathbf{p}^{L}_{i}=\sigma\left(\phi\left(\mathbf{x}_i^{L}\right)\right).
\end{equation}

\paragraph{MHA.} 
A MHA block consists of $M$ attention heads, which are capable of aggregating global information from the hidden states \cite{halawi2023overthinking, wang-etal-2023-label}.
An individual attention head $h$ in layer $\ell$ consists of three learnable matrices, $\mathbf{W}_Q^{\ell, h}, \mathbf{W}_K^{\ell, h}, \mathbf{W}_V^{\ell, h} \in \mathbb{R}^{d \times \frac{d}{M}}$.
Formally, for the input $\mathbf{X}^{\ell-1}=\left[\mathbf{x}_{1}^{\ell-1},\cdots,\mathbf{x}_{N}^{\ell-1}\right]$ in layer $\ell$:
\begin{gather}
\mathbf{A}^{\ell}=\left[\mathbf{H}^{\ell,1}; \cdots ; \mathbf{H}^{\ell,M}\right] \mathbf{W}_o^{\ell},\\
\mathbf{H}^{\ell,h}=\mathbf{s}^{\ell,h} \mathbf{X}^{\ell-1} \mathbf{W}_V^{\ell, j},\\
\mathbf{s}^{\ell,h}=\sigma\left(\frac{\left(\mathbf{X}^{\ell-1} \mathbf{W}_Q^{\ell, h}\right)\left(\mathbf{X}^{\ell-1} \mathbf{W}_K^{\ell, h}\right)^T}{\sqrt{d/M}}\right)
\end{gather}
where $\mathbf{A}^{\ell}=\left[\mathbf{a}_{1}^{\ell},\cdots,\mathbf{a}_{N}^{\ell}\right]$ is the MHA block's output. 
$\mathbf{W}_O^{\ell, h} \in \mathbb{R}^{d \times d}$ is a learnable output matrix.

\paragraph{FFN.} A FFN block can work as a key-value memory to store factual knowledge \cite{geva-etal-2021-transformer}, enriching the hidden states of token $i$:
\begin{equation}
\mathbf{m}_{i}^{\ell}=f\left( \left(\mathbf{x}_{i}^{\ell-1}+\mathbf{a}_{i}^{\ell}\right)\mathbf{W}_1^{\ell}\right)\mathbf{W}_2^{\ell}.
\end{equation}

\begin{figure*}[t]
        \vspace{-10pt}
    \captionsetup{skip=4pt}

    \centering
    \begin{subfigure}[b]{0.37\linewidth}
              \captionsetup{skip=0pt}

        \includegraphics[width=\textwidth]{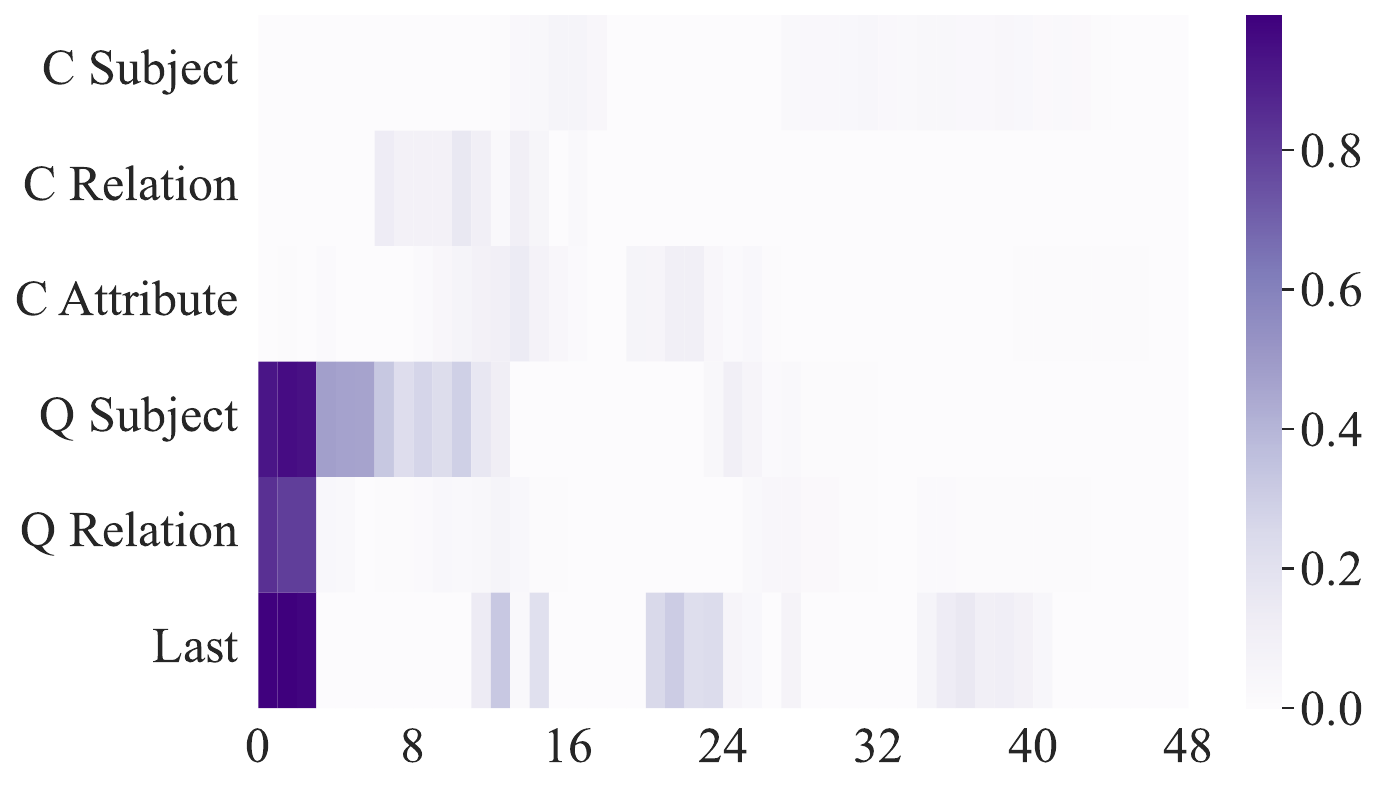}
        \caption{Effect of FFNs on internal memory.}
        \label{effect1}
    \end{subfigure}
    \begin{subfigure}[b]{0.31\linewidth}
              \captionsetup{skip=0pt}

        \includegraphics[width=\textwidth]{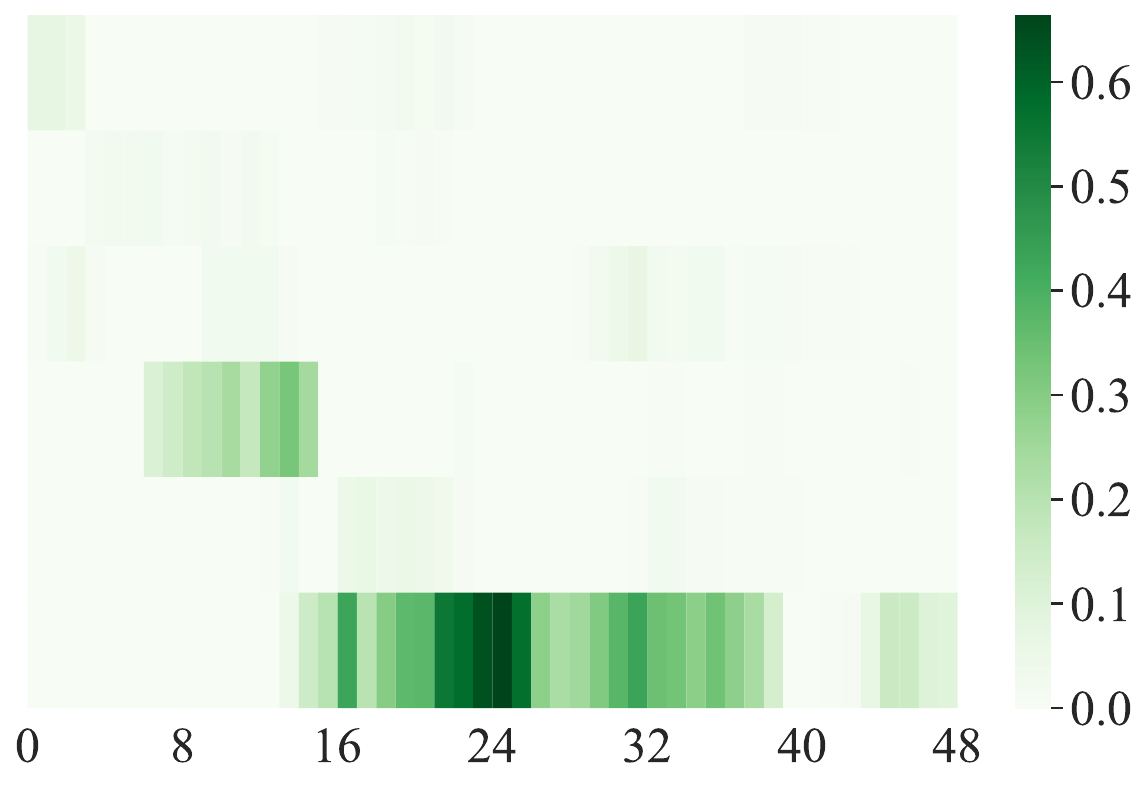}
        \caption{Effect of MHAs on internal memory.}
        \label{effect2}
    \end{subfigure}
    \begin{subfigure}[b]{0.305\linewidth}
              \captionsetup{skip=0pt}

        \includegraphics[width=\textwidth]{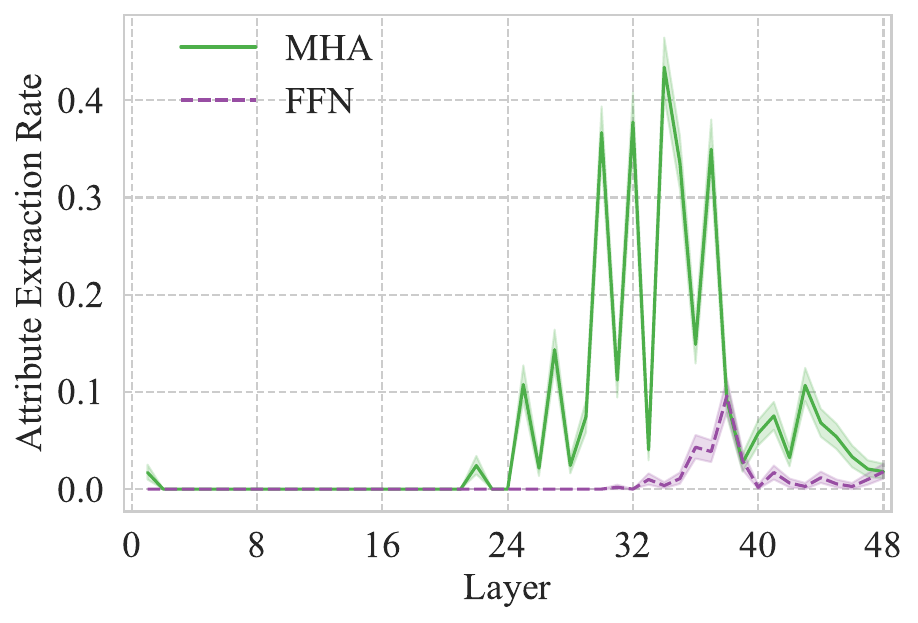}
        \caption{Extraction rate of internal memory.}
        \label{effect3}
    \end{subfigure}

    \begin{subfigure}[b]{0.37\linewidth}
              \captionsetup{skip=0pt}

        \includegraphics[width=\textwidth]{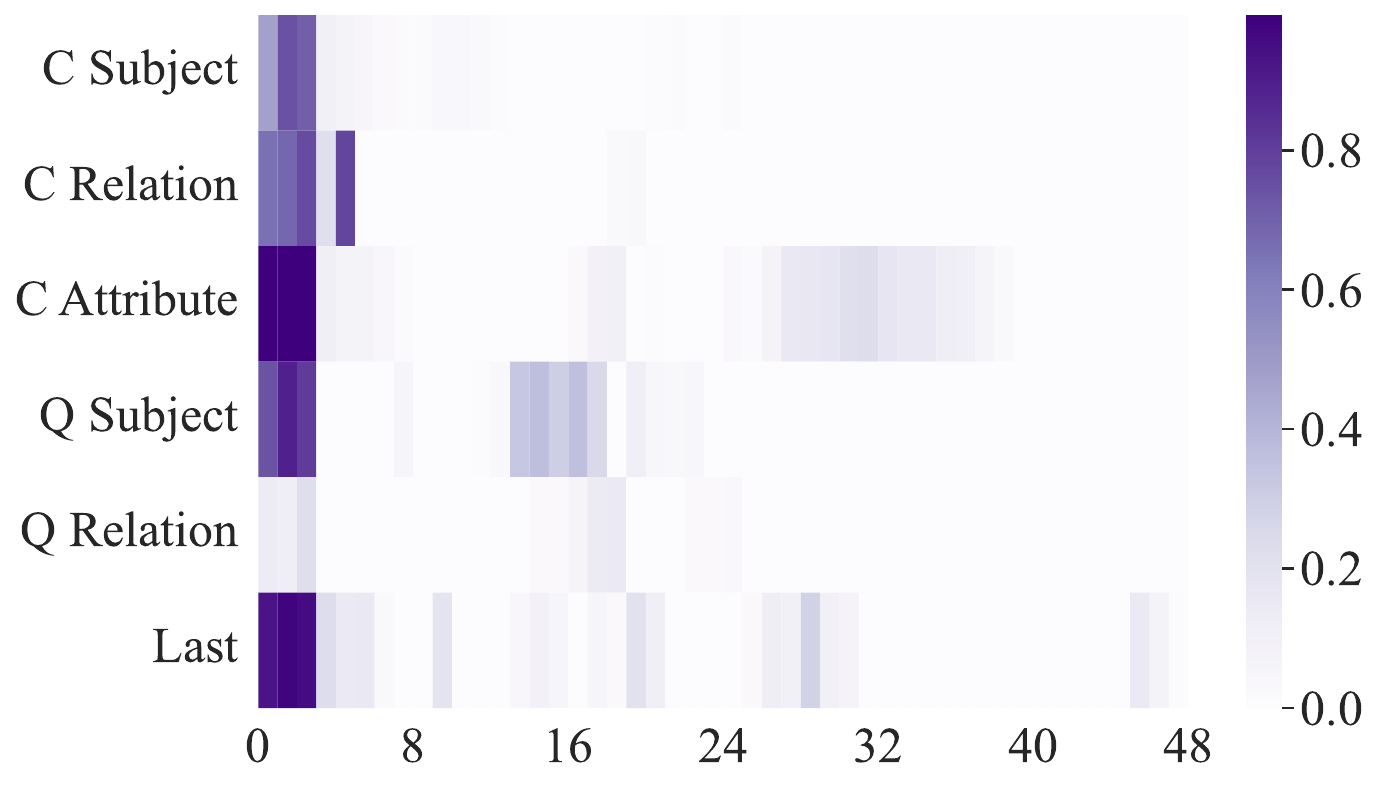}
        \caption{Effect of FFNs on external context.}
        \label{effect4}
    \end{subfigure}
    \begin{subfigure}[b]{0.31\linewidth}
              \captionsetup{skip=0pt}

        \includegraphics[width=\textwidth]{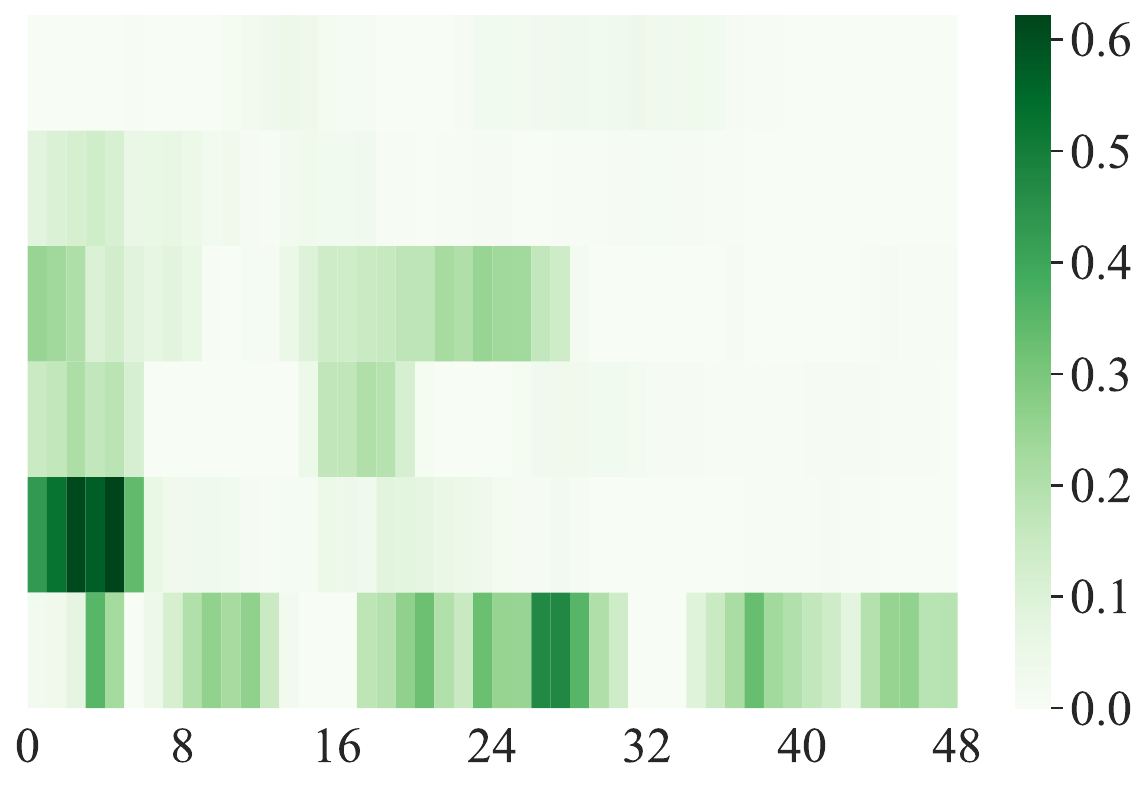}
        \caption{Effect of MHAs on external context.}
        \label{effect5}
    \end{subfigure}
    \begin{subfigure}[b]{0.305\linewidth}
              \captionsetup{skip=0pt}

        \includegraphics[width=\textwidth]{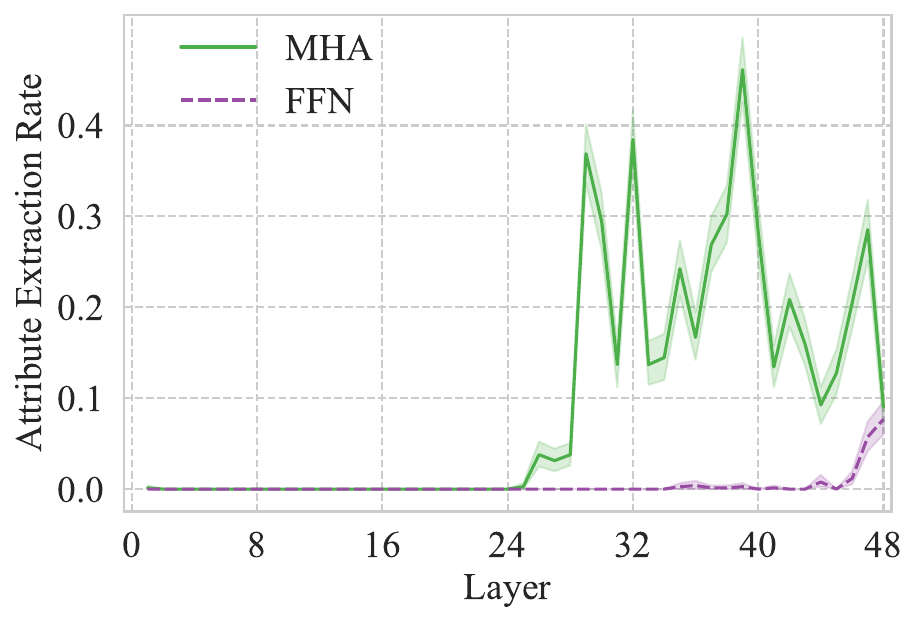}
        \caption{Extraction rate of external context.}
        \label{effect6}
    \end{subfigure}
    \caption{Effect of model components (FFNs and MHAs) in GPT-2 XL on the final prediction probability. Figures \ref{effect1} and \ref{effect2} (Figures \ref{effect4} and \ref{effect5}) show the effect of different model components and input elements when the model predicts based on internal memory (external context). The deeper color indicates the greater the impact of knocking out this part on the original prediction probability. Figure \ref{effect3} (Figure \ref{effect6}) shows the effect of MHAs and FFNs on the last token's attribute extraction rate when the model predicts based on internal memory (external context).}
        \vspace{-12pt}

            \label{effect}
\end{figure*}

\section{Experimental Setup}
\subsection{Tasks}
In this paper, we conduct controlled experiments to construct knowledge conflicts, wherein the internal memory is factual while the external context is counterfactual.
To avoid the LM being influenced by other irrelevant factors (\textit{i.e.}, reasoning ability), we adopt a simple factual recall task \cite{geva-etal-2023-dissecting}, which requires predicting the corresponding attribute $a_{m}$ based on the given subject $s$ and relation $r$.
Building on previous work \cite{yu-etal-2023-characterizing}, we use the \texttt{World Capital} dataset to interpret this problem in \S \ref{Interpreting}, where the LM needs to predict the capital city of the country based on the question $q$:
\begin{center}
\scalebox{1.0}{
    \fbox{
    \shortstack[c]{
    \textit{Q: What is the capital of} \{$s$\}\textit{? A:}
    }
    }}
\end{center}
We retain those questions that the LM can correctly predict the factual attributes $a_{m}$ based on internal memory, then provide the counterfactual attributes $a_{c}$ in the external context $c$ to construct conflicts:
\begin{center}
\scalebox{1.0}{
    \fbox{
    \shortstack[c]{
    \textit{The capital of} \{$s$\} \textit{is} \{$a_{c}$\}\textit{.} \{$q$\}}
    }}
\end{center}
To mitigate knowledge conflicts, we further construct three datasets for verifying the generalization of our method in \S \ref{Mitigating}, including the \texttt{Official Language}, \texttt{Country}, and \texttt{Continent} datasets.
We also generate a more complex \texttt{World Capital D} dataset based on the \texttt{World Capital} dataset, using \texttt{gpt-3.5-turbo} to rewrite the external context from triplet form into document form.
More details about these datasets are shown in Appendix \ref{Details}.

\subsection{Models}

We analyze two GPT-series LMs: GPT-2 XL \cite{gpt2} and GPT-J \cite{gpt-j} in \S \ref{Interpreting}.
Additionally, we also validate the effectiveness of our method on six LMs: OPT-1.3B, OPT-2.7B \cite{opt}, Pythia-6.9B, Pythia-12B \cite{pythia}, LLaMA2-7B and LLaMA2-13B \cite{touvron2023llama} in \S \ref{Mitigating}.

\section{Interpreting Knowledge Conflicts}
\label{Interpreting}

We utilize a ``\textit{top-down}'' analysis approach to locate the pivotal point where conflicts emerge and to identify the model components that are significant in knowledge conflicts.
We start by examining the functionality of model components by knocking out activations, and reveal that MHAs in the middle and late layers play a crucial role in passing information to the last token (\S \ref{4.2}).
Then, we further investigate MHAs by knocking out the attention weights.
We find the question information is first passed to the last token, then the last token extracts information from the subject and the attribute in the context (\S \ref{4.3}).
Last, we discover that some attention heads in later MHAs play opposite roles in conflicts, where memory heads can recall knowledge from internal memory,
and context heads can retrieve knowledge from external context (\S \ref{4.4}).

\subsection{Examining Component Functionality}
\label{4.2}

\begin{figure*}[t]
\vspace{-10pt}

              \captionsetup{skip=4pt}

    \centering
    \begin{subfigure}[b]{0.325\linewidth}
              \captionsetup{skip=0pt}

        \includegraphics[width=\textwidth]{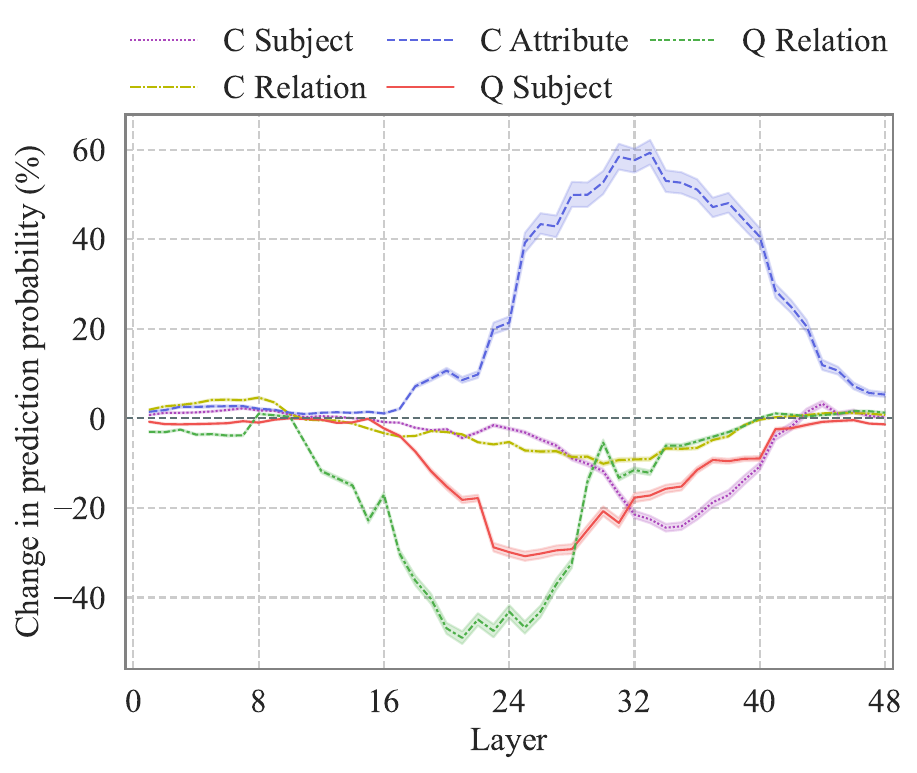}
        \caption{Prediction based on internal memory.}
            \label{flow1}
    \end{subfigure}
    \begin{subfigure}[b]{0.325\linewidth}
              \captionsetup{skip=0pt}

        \includegraphics[width=\textwidth]{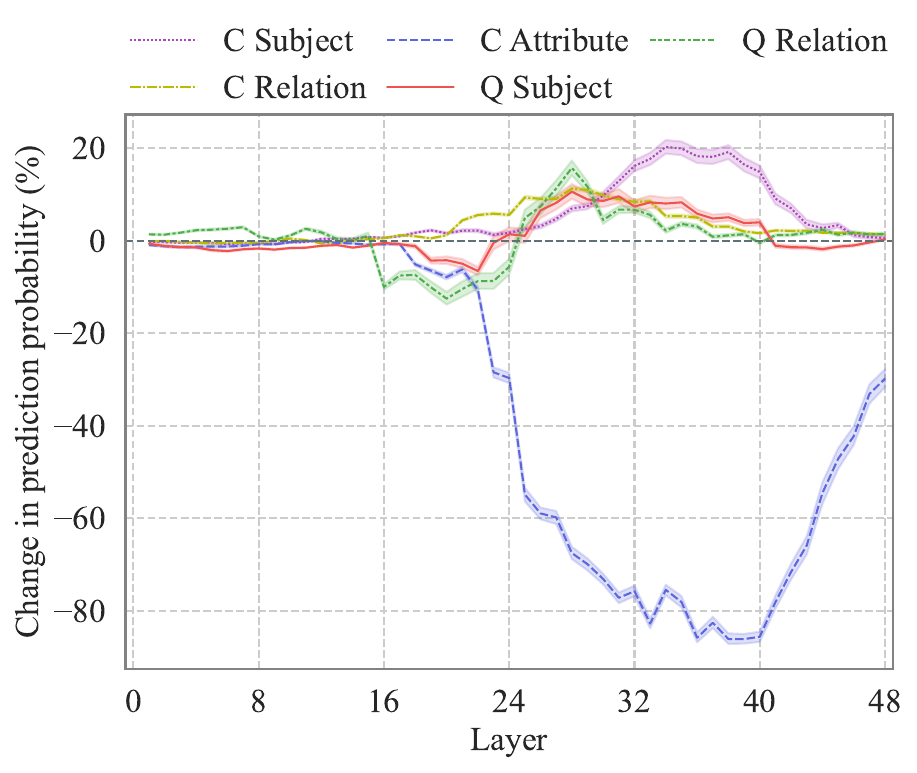}
        \caption{Prediction based on external context.}
            \label{flow2}
    \end{subfigure}
    \begin{subfigure}[b]{0.33\linewidth}
              \captionsetup{skip=0pt}

        \includegraphics[width=\textwidth]{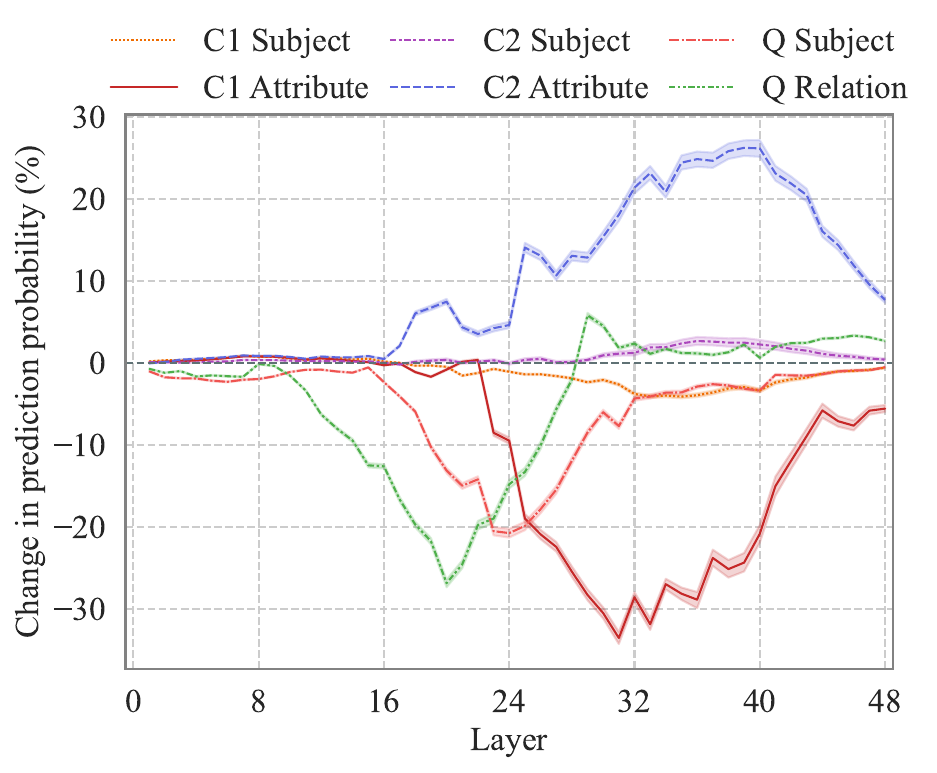}
        \caption{Prediction based on internal memory.}
            \label{flow3}
    \end{subfigure}

    \caption{Relative change in the prediction probability when blocking the information flow from the input elements to the last token. Figures \ref{flow1} and \ref{flow2} only provide conflicting context. Figure \ref{flow3} provides both supporting and conflicting context to internal memory, C1 denotes the supporting context, and C2 denotes the conflicting context.}
            \label{flow}
        \vspace{-12pt}

\end{figure*}

We start by exploring the functionality of model components (including FFNs and MHAs across various layers) in knowledge conflicts.

\paragraph{Experiment 1: Knocking Out Component.} We examine which component in the transformer layer is critical for the attribute prediction by knocking out activations.
Then, we divide the input into six elements for analysis: context subject $s_{c}$, context relation $r_{c}$, context attribute $a_{c}$, question subject $s_{q}$, question relation $r_{q}$, and the last token $x_{N}$.
To measure the impact on the final prediction results, we zero-out the updates to the specified input element from the MHA and FFN blocks within each layer.
For example, to intervene in the update of the $\ell$-th MHA (FFN) to the input element $s_{c}$, we set $\mathbf{a}_{i}^{\ell'}=\mathbf{0}$ ($\mathbf{m}_{i}^{\ell'}=\mathbf{0}$) for $i$ in the token range of $s_{c}$ and $\ell' = \operatorname{max}\left(1,\ell-W/2\right),\cdots,\operatorname{min}\left(L,\ell+W/2\right)$, where $W$ denotes the window size.
We define the effect of a model component as the change in the original prediction probability after knocking it out.

\paragraph{Results.}

Figure \ref{effect} illustrates the effect of model components (FFNs and MHAs) in GPT-2 XL with the window size $W=5$.
Our observation reveals that destroying the FFN blocks in the early layers has a significant effect on the prediction probability while destroying the FFN blocks at the late layers shows minimal or no impact (Figures \ref{effect1} and \ref{effect4}). 
Moreover, the MHA blocks at the middle and late layers are crucial for the last token (Figures \ref{effect2} and \ref{effect5}).
A possible explanation of the model’s behavior on the factual recall task is that \textit{the early FFNs first enrich the semantic information of input elements, and then the enriched semantic information about attributes is extracted to the last token via late MHAs, where knowledge conflicts may arise at the later stage}.
To verify this hypothesis, we will examine the attribute extraction function of MHAs.

\paragraph{Experiment 2: Extracting Attributes via MHAs.} We adopt the extraction rate \cite{geva-etal-2023-dissecting} to examine the attribute extraction function of MHAs.
We apply the early exit \cite{schuster-etal-2021-consistent, geva-etal-2022-transformer} to project the MHA update $\mathbf{a}_{N}^{\ell}$ for the last token $x_{N}$ over the vocabulary.
Then we check whether the top token $t^{\ell}$ of each update aligns with the attribute $t^{*}$ predicted at the final layer $L$:
\begin{gather}
t^*=\arg \max \left(\mathbf{p}_N^L\right),\\
t^{\ell}=\arg \max \left(\sigma\left(\phi\left( \mathbf{a}_N^{\ell}\right)\right)\right).
\end{gather}
We consider that the MHA correctly performs attribute extraction when $t^{*}=t^{\ell}$.
For comparison, we also examine the extraction rate of FFNs.

\paragraph{Results.} As illustrated in Figures \ref{effect3} and \ref{effect6}, it is evident that \textit{the attribute extraction rate of MHAs significantly exceeds that of FFNs}. Moreover, attribute extraction mainly takes place at the 24-48 layers.
Results for GPT-J show similar trends in Appendix \ref{GPT-J}.
The above findings motivate us to conduct an in-depth study on the information flows of MHAs from input elements to the last token.

\subsection{Tracing Information Flow}
\label{4.3}
The analysis presented above confirms that the last token extracts attribute information for prediction through MHA blocks.
Following this, we explore the order and importance of the information flow from the various elements to the last token.

\paragraph{Experiment 3: Blocking Information Flow.}

We localize the information propagation from the input elements (including $s_{c}$, $r_{c}$, $a_{c}$, $s_{q}$ and $r_{q}$) to the last token by knocking out attention edges between them.
For example, to block the information flow from the input element $s_{c}$ to the last token $x_{N}$ in the layer $\ell$, we set the attention weight $s^{\ell,h}\left[N,i\right]=0$ for $i$ in the token range of $s_{c}$, $h=1,\cdots,M$, and $\ell' = \operatorname{max}\left(1,\ell-W/2\right),\cdots,\operatorname{min}\left(L,\ell+W/2\right)$.
In this way, we can restrict the last token from attending to the target element.
If blocking the information propagation between them has a significant impact on the original prediction probability, this indicates that it is a crucial information flow.

\paragraph{Results.} Figure \ref{flow} illustrates the information flow in GPT-2 XL with the window size $W=9$.
We can observe that in the early to middle layers, blocking the attention to the question relation leads to a decrease in the prediction probability.
Similarly, in the subsequent layers, blocking the attention to the question subject also results in a decrease in the prediction probability.
This suggests that the critical relation and subject information in the question are sequentially transmitted to the last token.

Then, in the middle to late layers, blocking the attention to the context subject and context attribute has the opposite effect on the final prediction probability.
Taking Figure \ref{flow1} as an example (when the model predicts the attribute based on internal memory), blocking the attention to the context attribute can improve the prediction probability, however, blocking the attention to the context subject can reduce the prediction probability.
This suggests that the last token can extract the internal knowledge from the context subject, and extract the external knowledge from the context attribute.
In addition, the last token also extracts a certain degree of internal knowledge from the question subject.
Results for GPT-J show consistent trends in Appendix \ref{GPT-J}.

Overall, this shows that there are two specific stages in the process of information flow passing to the last token: (1) \textit{the question information is first passed to the last token}; (2) \textit{the last token extracts or copies the attribute from the context subject or the context attribute}.
In the later stage, knowledge conflicts arise during the process of merging inconsistent information flows from MHAs.

\begin{figure*}[t]              \captionsetup{skip=4pt}

\vspace{-10pt}

    \centering
    \begin{subfigure}[b]{0.327\linewidth}
              \captionsetup{skip=0pt}

        \includegraphics[width=\textwidth]{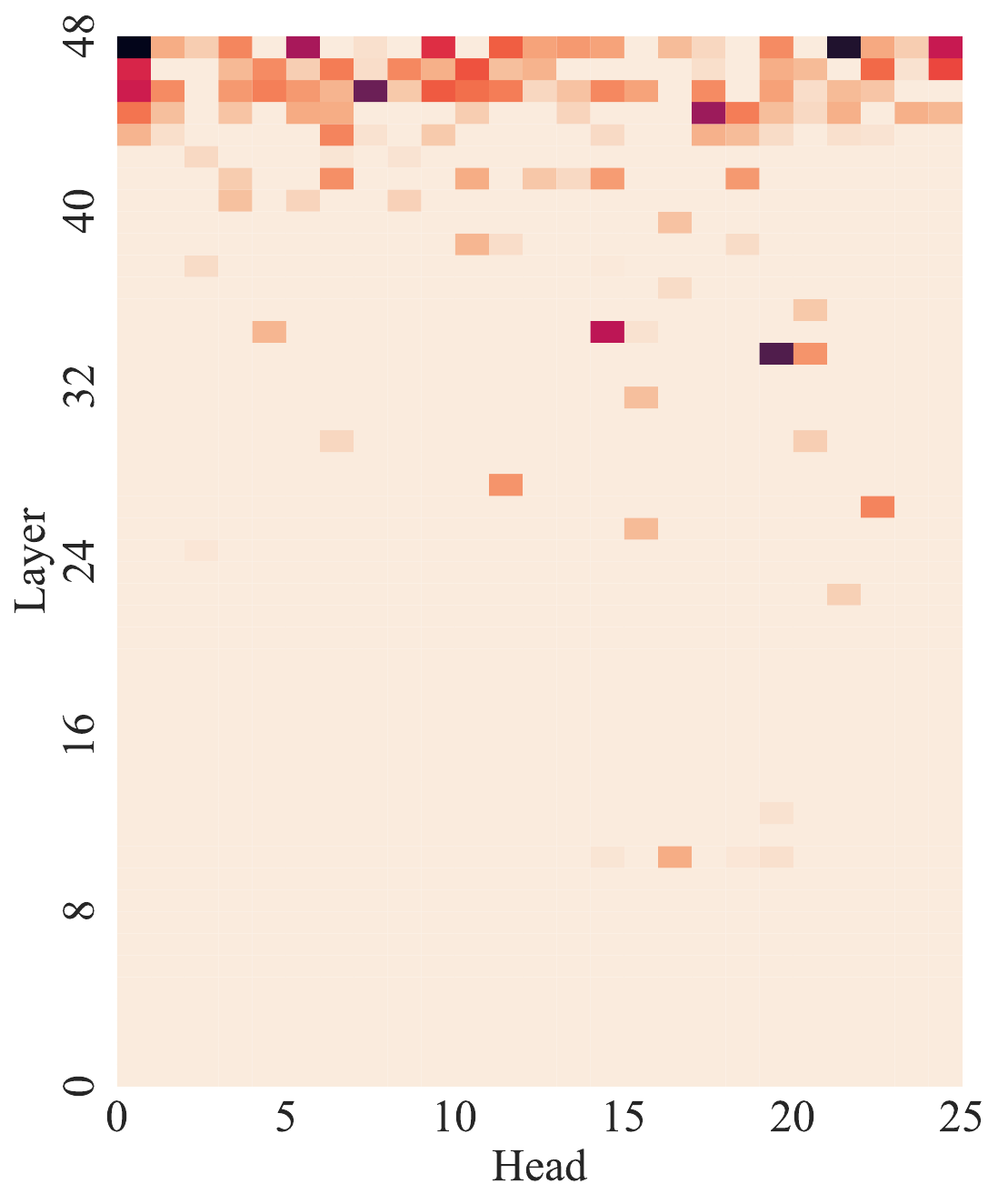}
        \caption{Memory head.}
            \label{head1}
    \end{subfigure}
    \begin{subfigure}[b]{0.370\linewidth}
              \captionsetup{skip=0pt}

        \includegraphics[width=\textwidth]{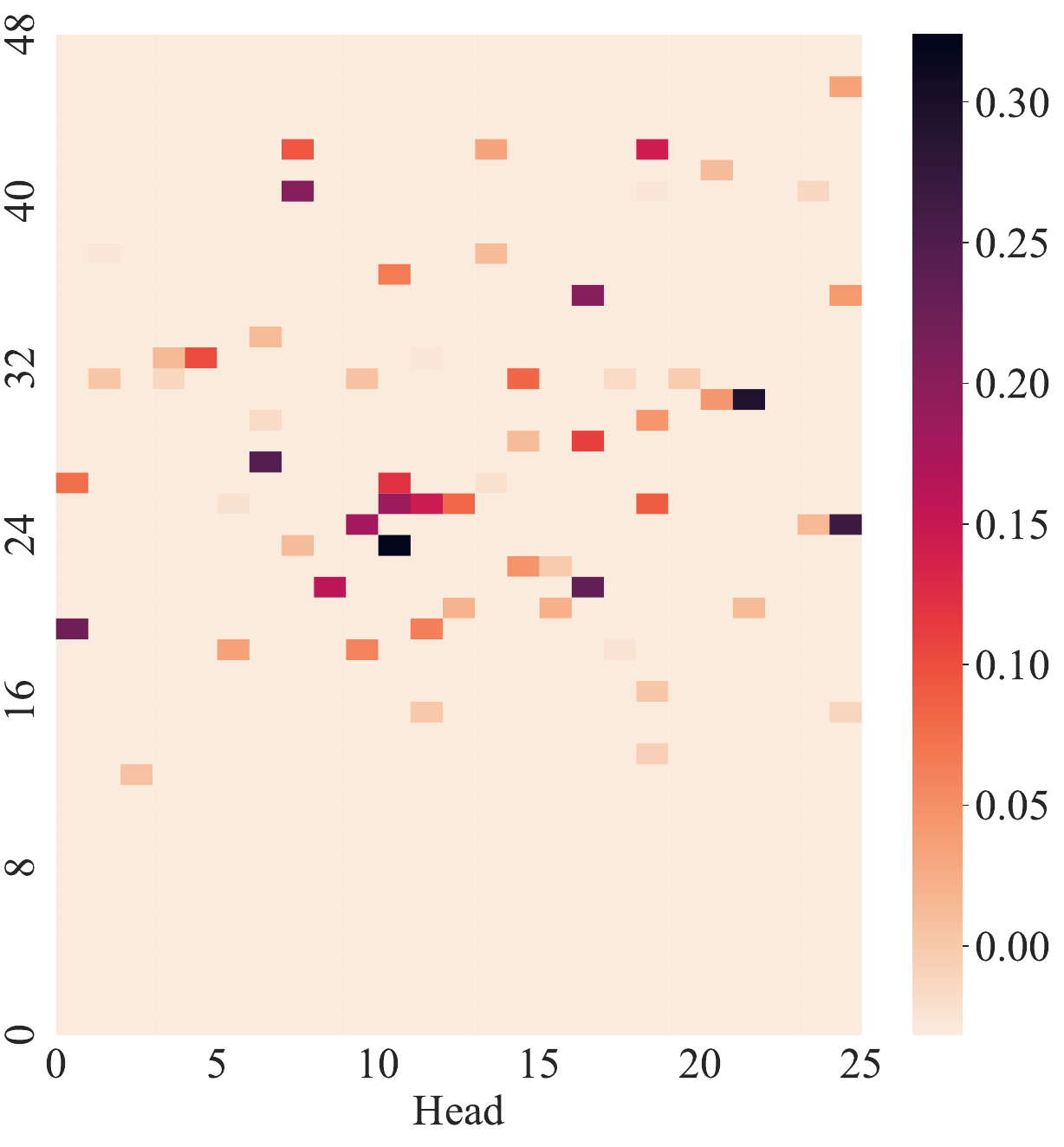}
        \caption{Context head.}
            \label{head2}
    \end{subfigure}
    \begin{subfigure}[b]{0.29\linewidth}
              \captionsetup{skip=0pt}

        \includegraphics[width=\textwidth]{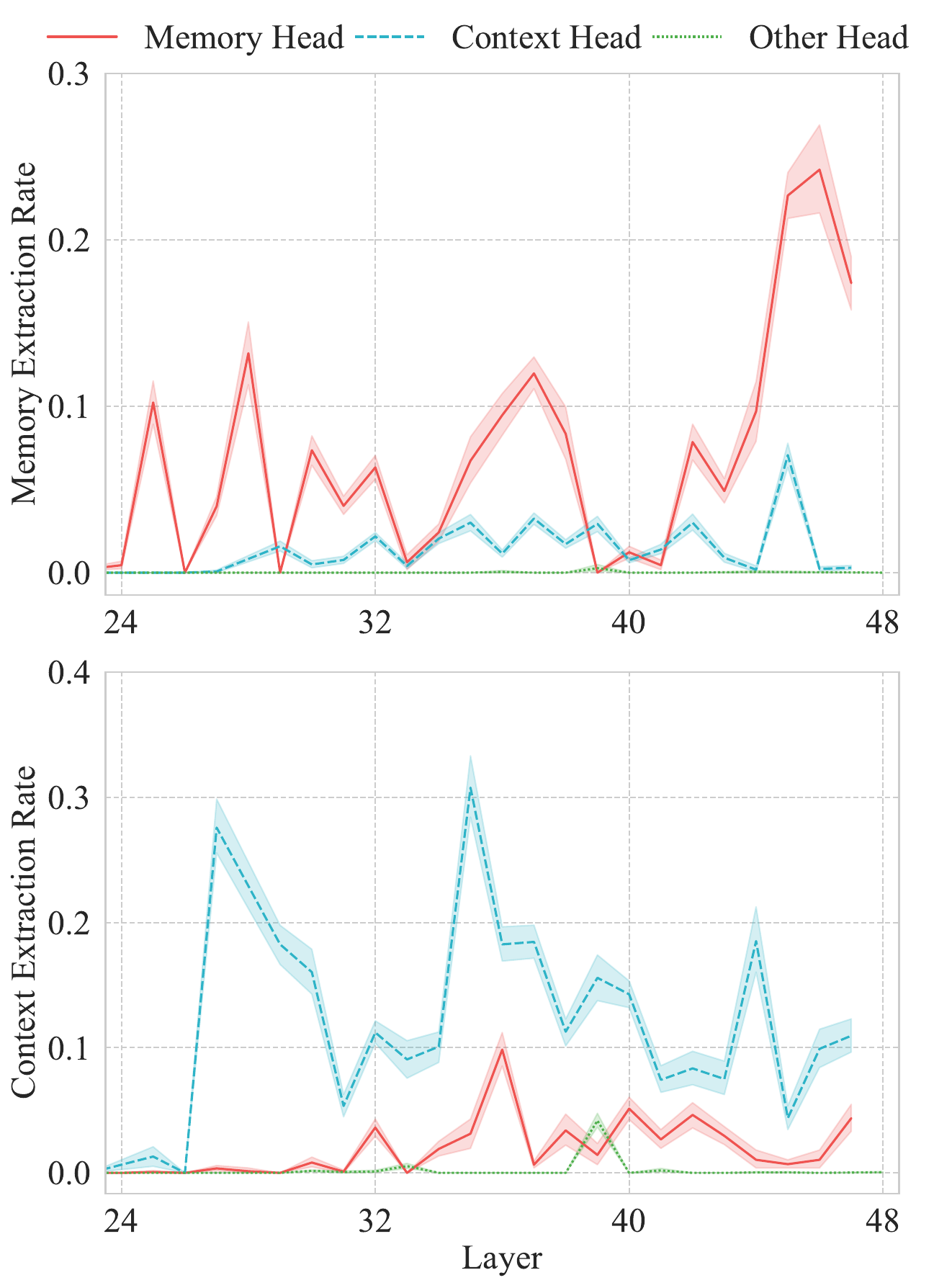}
        \caption{Extraction rate of attention heads.}
            \label{head3}
    \end{subfigure}

    \caption{Memory heads and context heads in GPT-2 XL. Figure \ref{head1} shows the important score heatmap for predicting based on internal memory. Figure \ref{head2} shows the important score heatmap for predicting based on external context. Figure \ref{head3} illustrates the memory and context attribute extraction rate of different attention heads. }
            \label{head}
        \vspace{-12pt}

\end{figure*}

\paragraph{Experiment 4: Extending to Conflicts between Contexts.}

We extend our analysis to a more complex scenario in which the model is presented with both supporting context and conflicting context relative to internal memory.
Supporting context and conflicting context contain $a_{m}$ and $a_{c}$ respectively:
\begin{center}
\scalebox{1.0}{
    \fbox{
    \shortstack[c]{
    \textit{C1: The capital of} \{$s$\} \textit{is} \{$a_{m}$\}\textit{.} \\
    \textit{C2: The capital of} \{$s$\} \textit{is} \{$a_{c}$\}\textit{.} \{$q$\}}
    }}
\end{center}
We find that GPT-2 XL prefers to choose attributes consistent with internal memory 97.6\% of the time.
Hence, we only analyze the cases where the model makes predictions based on its internal memory.

\paragraph{Results.}

As illustrated in Figure \ref{flow3}, we can observe that the question information is first passed to the last token in the first stage, which is consistent with the trend of a single conflicting context.
In the second stage, a notable distinction is that \textit{the model no longer extracts the memory attribute from the subject; instead, it opts for a more straightforward approach of copying the memory attribute from the context}.
The above findings indicate that there exists a mechanism within MHAs capable of distinguishing and selecting between internal knowledge and external knowledge. This motivates us to conduct further analysis of MHAs.

\subsection{Looking Deeper into Attention Heads}
\label{4.4}

Attention heads serve as the fundamental component of an MHA block.
For example, GPT-2 XL contains a total of 1,200 attention heads.
This motivates us to conduct an investigation into the role of attention heads in handling knowledge conflicts.
\paragraph{Experiment 5: Discovering Important Heads.} To discover the attention heads that are crucial for predicting memory attributes or context attributes, we compute the gradient-based importance score \cite{NEURIPS2019_2c601ad9, bansal-etal-2023-rethinking} for each head.
Given a dataset $\mathcal{D}$ with a set of inputs $x$ and outputs $y$, the importance score of an attention head $h$ captures the expected sensitivity of the model to $h$ and is computed as follows:
\begin{equation}
I^{l,h}(\mathcal{D})=\mathbb{E}_{(x, y)}\left|{\mathbf{H}^{l,h}}^{T} \frac{\partial \mathcal{L}(y \mid x)}{\partial \mathbf{H}^{l,h}}\right|,
\end{equation}
where $\mathcal{L}(\cdot)$ is the loss function of conditional autoregressive generation.
The proxy score of head $h$ for predicting internal memory is calculated as:
\begin{equation}
S^{l,h}_{m}(\mathcal{D}_{m}, \mathcal{D}_{m}^{\prime})=I^{l,h}(\mathcal{D}_{m})-I^{l,h}(\mathcal{D}_{m}^{\prime}),
\end{equation}
where $(x, a_{m}) \in \mathcal{D}_{m}$ denotes the original outputs are memory attributes, $(x, a_{c}) \in \mathcal{D}_{m}^{\prime}$ denotes replacing the original outputs with context attributes.
In this way, we can also calculate the proxy score of head $h$ for predicting external context as:
\begin{equation}
S^{l,h}_{c}(\mathcal{D}_{c}, \mathcal{D}_{c}^{\prime})=I^{l,h}(\mathcal{D}_{c})-I^{l,h}(\mathcal{D}_{c}^{\prime}).
\end{equation}
We compute the proxy score of each head across different layers to discover important heads.

\paragraph{Results.}
As shown in Figure \ref{head1} (Figure \ref{head2}), the deeper color of the red square indicates a more significant contribution from this attention head to the model's predictions based on internal memory (external context).
We can observe that there are a specific number of attention heads within middle-to-late layers that play opposite roles in predicting attributes.
Accordingly, we refer to those heads that contribute to the prediction of memory attributes as \textbf{memory heads}, and those that facilitate predicting context attributes as \textbf{context heads}.
Therefore, we claim that \textit{they may serve in a mutually exclusive capacity during knowledge conflicts}.
The heatmaps of GPT-J are provided in the Appendix \ref{GPT-J}.

\paragraph{Experiment 6: Extracting Specific Attributes via Heads.}
We further analyze the two types of heads discovered above to verify their role in knowledge conflicts.
We rank the attention heads in descending order based on their importance scores, $S^{l,h}_{m}$ for memory and $S^{l,h}_{c}$ for context, subsequently identifying the top-5\% of heads as memory heads and context heads, respectively.
For comparison, we also randomly choose an additional 5\% of the attention heads as other heads.
Then, we examine their memory extraction rate when $t_{\ell} = a_{m}$, and context extraction rate when $t_{\ell} = a_{c}$.

\paragraph{Results.}
As shown in Figure \ref{head3}, memory heads and context heads are responsible for extracting different attribute information to the last token with a significant difference between memory and context extraction rates.
Therefore, we discern that \textit{the pivotal point at which knowledge conflicts emerge in LMs is the integration of inconsistent information flows by memory heads and context heads}.

\section{Mitigating Knowledge Conflicts}
\label{Mitigating}

Building on the above insights, we propose a novel method called \textbf{P}runing \textbf{H}ead via \textbf{P}at\textbf{H} \textbf{P}atc\textbf{H}ing (\textbf{PH3}) to efficiently mitigate knowledge conflicts by intervening on attention heads without the need to update model parameters (\S \ref{Method}). Then, we conduct extensive experiments to show that our method can flexibly control LMs to use internal memory or external context (\S \ref{Experiment}). 
Moreover, we analyze the generalization capability of our method  (\S \ref{Analysis}).

\subsection{Method}
\label{Method}

Our method consists of two stages, first identifying the important heads through path patching, then intervening on these heads via structured pruning.

\paragraph{Localizing Memory Heads and Context Heads via Path Patching.}
When we use the gradient-based method in \S \ref{4.4} to estimate the importance score of the target head $h$, it is subject to interference from other heads.
The calculated gradients may not fully reflect the contribution of the target head, but rather a mixture of the influences from other heads.
Therefore, we adopt the path patching technique \cite{goldowsky2023localizing, wang2023interpretability} to analyze the causal relationship between the head $h$ and the output attribute (including $a_{m}$ and $a_{c}$) in conflicts.
To calculate the important score $S_{c}^{\ell,h}$ of the target head $h$, our path patching method consists of three steps shown in Figure \ref{method}:
\begin{itemize}

\item[1.]  Run on the original input $x \in \mathcal{D}_{c}$ to record the original activations of all heads;
\item[2.]  Run on the corrupted input $\cancel{x}$ to record the corrupted activations of all heads, where $\cancel{x}$ is:
\begin{center}
\scalebox{1.0}{
    \fbox{
    \shortstack[c]{
    \textit{The capital of} \{$s$\} \textit{is} $\langle$unk$\rangle$\textit{.} \{$q$\}}
    }}
\end{center}
where $\langle$unk$\rangle$ is the special token;
\item[3.] Run on the original input $x$, while keeping all the heads frozen to their activations on $x$, except for the target head $h$ whose activation is set on $\cancel{x}$.
Then measure the important score as the change of output logits.

\end{itemize}
The important score $S_{c}^{\ell,h}$ of head $h$ is computed as:
\begin{equation}
\begin{aligned}
S^{l,h}_{c}(\mathcal{D}_{c})=\mathbb{E}_{(x)}[\left(\mathbb{P}_{x}(a_{c})-\mathbb{P}_{x}(a_{m})\right) \\ -\left(\mathbb{P}_{\cancel{x}}(a_{c})-\mathbb{P}_{\cancel{x}}(a_{m})\right)].
\end{aligned}
\end{equation}
We adopt similar steps to calculate the importance score $S_{m}^{\ell,h}$ of the target head $h$ for memory attribute prediction in Appendix \ref{Amethod}.
We also provide the importance score heatmaps of memory and context heads for various models in Appendix \ref{Aheat}, and our method can clearly distinguish between them.

\begin{table*}[t]
\vspace{-13pt}
\centering
\resizebox{0.76\textwidth}{!}{
\begin{tabular}{cllcccccccccc}
\toprule
                                 & \multicolumn{2}{l}{}                                  & \multicolumn{2}{c}{\textbf{\texttt{World}}}   & \multicolumn{2}{c}{\textbf{\texttt{World}}}     & \multicolumn{2}{c}{\textbf{\texttt{Official}}} & \multicolumn{2}{c}{}                                   & \multicolumn{2}{c}{}                                     \\
                                 & \multicolumn{2}{l}{}                                  & \multicolumn{2}{c}{\textbf{\texttt{Capital}}} & \multicolumn{2}{c}{\textbf{\texttt{Capital D}}} & \multicolumn{2}{c}{\textbf{\texttt{Language}}} & \multicolumn{2}{c}{\multirow{-2}{*}{\textbf{\texttt{Country}}}} & \multicolumn{2}{c}{\multirow{-2}{*}{ \textbf{\texttt{Continent}}}} \\ \cmidrule{4-13} 

\multirow{-3}{*}{\textbf{Model}} & \multicolumn{2}{c}{\multirow{-3}{*}{\textbf{Method}}} & \textbf{RM}          & \textbf{RC}         & \textbf{RM}           & \textbf{RC}          & \textbf{RM}            & \textbf{RC}           & \textbf{RM}        & \textbf{RC}       &  \textbf{RM}   &  \textbf{RC}  \\ \midrule
                                 \rowcolor[RGB]{246, 246, 246} \cellcolor{white} &  \multicolumn{2}{c}{Base}                               & 59.2              & 40.8             & 47.2               & 52.8              & 42.2              & 57.8              & 37.2                      & 62.8                      & 41.5                       & 58.5         \\ \cmidrule{2-13} 
                                 \rowcolor[RGB]{255, 231, 231} \cellcolor{white}&       \cellcolor{white}       &              Prompt                 &  12.5              & 81.2             & 21.3               & 71.2              & 20.3              & 74.4              & 24.5                      & 75.5                      & 16.2                       & 43.4         \\
                                 \rowcolor[RGB]{255, 231, 231} \cellcolor{white}&    \cellcolor{white}     $\uparrow$ Memory                     & Gradient               & 72.4              & 9.8              & 78.6               & 10.5              & 41.5              & 40.5              & 39.1                      & 60.2                      & 42.7                       & 46.6                                \\
                                 \rowcolor[RGB]{255, 197, 197} \cellcolor{white}&    \cellcolor{white}       & PH3 (Ours)                  & \textbf{97.9}     & \textbf{0.6}     & \textbf{93.3}      & \textbf{2.5}      & \textbf{74.4}     & \textbf{9.8}      & \textbf{50.9}             & \textbf{36.3}             & \textbf{53.1}              & \textbf{38.1} \\ \cmidrule{2-13} 
                                \rowcolor[RGB]{238, 239, 255} \cellcolor{white}GPT-2 XL &        \cellcolor{white}                      & Prompt                 & 9.3               & 87.5             & 18.9               & 75.2              & 17.1              & 80.7              & 18.5                      & 81.4                      & 25.5                       & 58.3          \\
                                 \rowcolor[RGB]{238, 239, 255} \cellcolor{white}&       \cellcolor{white}                       & CAD                    & 25.0              & 65.6             & 12.5               & 63.6              & \textbf{9.1}      & 80.5              & 27.2                      & 72.5                      & 22.9                       & 60.4                                \\
                                 \rowcolor[RGB]{238, 239, 255} \cellcolor{white}&      \cellcolor{white}              $\uparrow$ Context           & Gradient               & 44.4              & 49.0             & 28.0               & 58.7              & 29.6              & 59.5              & 36.4                      & 63.4                      & \textbf{18.4}              & 51.5                                 \\
                                 \rowcolor[RGB]{212, 216, 255} \cellcolor{white}&          \cellcolor{white}                    & PH3 (Ours)                   & 27.5              & 68.9             & 7.7                & 91.3              & 20.7              & 74.8              & 22.7                      & 75.7                      & 27.7                       & \textbf{66.7}                                \\
    \rowcolor[RGB]{212, 216, 255}  \cellcolor{white}&     \cellcolor{white}     & \ \ \ \  + Prompt          & \textbf{3.6}      & \textbf{95.1}    & \textbf{5.2}       & \textbf{94.4}     & 9.6      & \textbf{88.7}     & \textbf{12.6}             & \textbf{86.0}             & 20.9              & 63.8                       \\ \midrule
                                 \rowcolor[RGB]{246, 246, 246} \cellcolor{white}& \multicolumn{2}{c}{Base}                              & 37.5              & 62.5             & 43.1               & 56.9              & 41.5              & 58.5              & 54.0                      & 46.0                      & 43.2                       & 56.8                                \\ \cmidrule{2-13} 
                                 \rowcolor[RGB]{255, 231, 231} \cellcolor{white}&                             \cellcolor{white} & Prompt                 & 29.8              & 67.1             & 31.6               & 62.1              & 23.1              & 69.1              & 22.4                      & 77.6                      & 12.5                       & 86.0                                \\
                                 \rowcolor[RGB]{255, 231, 231} \cellcolor{white}&   \cellcolor{white} $\uparrow$ Memory                           & Gradient                & 67.9              & 8.3              & 67.6               & \textbf{6.7}      & 39.4              & 53.4              & 54.4                      & 45.5                      & 57.2                       & 30.0                                \\
                                 \rowcolor[RGB]{255, 197, 197} \cellcolor{white} &    \cellcolor{white}     & PH3 (Ours)                   & \textbf{93.3}     & \textbf{1.6}     & \textbf{76.5}      & 10.8              & \textbf{63.3}     & \textbf{25.3}     & \textbf{58.9}             & \textbf{40.5}             & \textbf{75.1}              & \textbf{17.6}                        \\ \cmidrule{2-13} 
                                \rowcolor[RGB]{238, 239, 255} \cellcolor{white} GPT-J&   \cellcolor{white}                           & Prompt                 & 31.9              & 64.5             & 15.8               & 76.4              & 16.2              & 70.6              & 17.9                      & 82.1                      & 7.2                        & \textbf{91.4}                                 \\
                                 \rowcolor[RGB]{238, 239, 255} \cellcolor{white}&  \cellcolor{white}                            & CAD                    & 2.5               & 89.9             & 13.4               & 68.2              & 4.7               & 89.9              & 17.0                      & 81.8                      & 13.0                       & 80.3                                \\
                                \rowcolor[RGB]{238, 239, 255} \cellcolor{white} & \cellcolor{white} $\uparrow$ Context                            & Gradient               & 6.1               & 88.3             & 7.9                & 67.8              & 5.7               & 76.4              & 29.2                      & 70.5                      & 36.8                       & 60.7                                \\
                                \rowcolor[RGB]{212, 216, 255} \cellcolor{white}&  \cellcolor{white}                            & PH3 (Ours)                  & 0.2               & 99.3             & \textbf{0.1}       & \textbf{98.4}     & 2.3               & \textbf{90.6}     & 9.5                       & 86.7                      & 8.0                        & 64.9                                \\
 \rowcolor[RGB]{212, 216, 255}   \cellcolor{white}      &      \cellcolor{white}     & \ \ \ \  + Prompt         & \textbf{0.1}      & \textbf{99.5}    & 0.2                & 97.8              & \textbf{2.0}      & 81.9     & \textbf{1.4}              & \textbf{98.6}             & \textbf{1.4}               & 90.9                      \\ \midrule
                                 \rowcolor[RGB]{246, 246, 246}  \cellcolor{white}&  \multicolumn{2}{c}{Base}                              & 46.3              & 53.7             & 95.5               & 4.0               & 18.8              & 80.3              & 52.9                      & 46.8                      & 30.9                       & 69.1                                \\ \cmidrule{2-13} 
                                 \rowcolor[RGB]{255, 231, 231} \cellcolor{white}&  \cellcolor{white}                            &  Prompt                  & 36.0              & 63.2             & 96.0               & 3.7               & 40.0              & 59.1              & 68.2                      & 31.6                      & 77.4                       & 22.6                                 \\
                                 \rowcolor[RGB]{255, 231, 231}  \cellcolor{white}&   \cellcolor{white}                     $\uparrow$ Memory        & Gradient               & 81.0              & 5.8              & 95.1               & 1.6               & 50.1              & 47.4              & 60.0                      & 38.0                      & 64.5                       & 24.5                                \\
                                 \rowcolor[RGB]{255, 197, 197} \cellcolor{white}&      \cellcolor{white}     &   PH3 (Ours)                   & \textbf{98.1}     & \textbf{1.2}     & \textbf{98.0}      & \textbf{1.3}      & \textbf{73.7}     & \textbf{17.8}     & \textbf{76.9}             & \textbf{20.6}             & \textbf{90.5}              & \textbf{8.8}                                \\ \cmidrule{2-13} 
                                 \rowcolor[RGB]{238, 239, 255} \cellcolor{white} LLaMA2-7B&     \cellcolor{white}                         & Prompt               & 3.2               & 96.6             & 92.4               & 2.2               & 25.5              & 73.8              & 58.2                      & 41.5                      & 19.2                       & 80.3                               \\
                                \rowcolor[RGB]{238, 239, 255} \cellcolor{white}&  \cellcolor{white}                            & CAD                    & 1.4               & 95.5             & 29.1               & 70.6              & \textbf{0.0}      & \textbf{100.0}    & 13.6                      & 86.1                      & 0.2                        & 98.2                               \\
                                 \rowcolor[RGB]{238, 239, 255} \cellcolor{white}& \cellcolor{white} $\uparrow$ Context                            & Gradient               & 23.6              & 63.2             & 40.1               & 58.8              & 25.7              & 74.6              & 17.6                      & 82.2                      & 27.1                       & 72.9                                \\
                                 \rowcolor[RGB]{212, 216, 255}\cellcolor{white}& \cellcolor{white}                             & PH3 (Ours)                    & 1.6               & 97.4             & 19.1               & 73.4              & 0.1               & 99.9              & 5.2                       & 94.7                      & 0.5                        & 99.4                                \\
     \rowcolor[RGB]{212, 216, 255} \cellcolor{white}&     \cellcolor{white}      & \ \ \ \  + Prompt          & \textbf{0.4}      & \textbf{98.8}    & \textbf{10.6}      & \textbf{85.3}     & \textbf{0.0}      & \textbf{100.0}    & \textbf{2.8}              & \textbf{97.0}             & \textbf{0.0}               & \textbf{100.0}                                      \\ \bottomrule
\end{tabular}
}
\caption{Experimental results of GPT-2 XL, GPT-J and LLaMA2-7B on five datasets. Bolds denote the best results.}
\vspace{-12pt}
\label{exp}
\end{table*}

\paragraph{Pruning Attention Heads to Mitigate Knowledge Conflicts.}

By ranking all the attention heads in ascending order based on the importance score $S_{c}^{l,h}$ ($S^{l,h}_{m}$), we can prune the top-$k$\% attention heads that negatively impact the model's capability to predict context (memory) attributes, thereby enhancing the model's ability to utilize external context (internal memory).
To prune a head $h$ in layer $\ell$ in practice, we set $\mathbf{H}^{\ell,h}$ to be the zero matrix.

\subsection{Experiment}
\label{Experiment}

\paragraph{Setups.}

We evaluate our method on five datasets, including \texttt{World Capital}, \texttt{World Capital D}, \texttt{Official Language}, \texttt{Country}, and \texttt{Continent}.
To verify the generalization of PH3, we only calculate the importance scores of the attention heads on the \texttt{World Capital} dataset, and then directly evaluate PH3 on other datasets.
We also select 1,000 test samples from an open-domain QA dataset NQ \cite{NQ}, providing the LM with the top-5 retrieved passages, and ensuring that at least one relevant passage is among them.
We validate the effectiveness of PH3 on eight LMs.

\paragraph{Metrics.} 
We use the internal memory usage rate $RM=\frac{f_{m}}{f_{m}+f_{c}+f_{o}}$ and the external context usage rate $RC=\frac{f_{c}}{f_{m}+f_{c}+f_{o}}$ to assess how effectively the method controls the reliance of LMs on either internal memory or external context, where $f_{m}$ is the frequency of relying on internal memory, $f_{c}$ is the frequency of relying on external context, and $f_{o}$ is the frequency of other answers.
For the open-domain QA task, we use Recall to evaluate whether the model can provide correct answers based on the retrieved passages following \citet{adlakha2023evaluating}.

\paragraph{Baselines.} We compare with the following baselines: (1) \textbf{Prompt}: We instruct the LM to generate answers based on internal memory or external context through specific prompts;
(2) \textbf{CAD}: \citet{shi2023trusting} leverage contrastive decoding \cite{li-etal-2023-contrastive} to encourage the LM to attend to its context during generation; (3) \textbf{Gradient}: We replace our path patching method with the gradient-based method to discover the attention heads.
We select the optimal pruning rate $k$ on the development set for both Gradient and PH3.
More details about hyperparameter settings are in Appendix \ref{AHyperparameter}.

\paragraph{Results.} 
Table \ref{exp} shows the results of GPT-2 XL, GPT-J and LLaMA2-7B, and more results of other models are in Table \ref{exp1}.
Throughout our experiments, we note the following key observations:

(1) PH3 significantly outperforms other baselines. Experimental results show that PH3 can not only increase the average internal memory usage rate of eight LMs by 44.0\%, but also increase the average external context usage rate by 38.5\%.
When PH3 is combined with Prompt, it can more effectively control the LMs to use external context.

(2) As shown in Table \ref{NQ}, PH3 can also achieve an average 6.2\% Recall improvement on open-domain QA tasks.
By pruning a small number of negative context heads, PH3 can make LMs generate answers more faithfully based on retrieved passages.

(3) Although Prompt and CAD can effectively increase the external context usage rate, there are limitations. CAD cannot directly enhance internal memory, and Prompt may even have the opposite effect.
In contrast, our method offers a viable solution to enhance the internal memory usage rate.

\subsection{Analysis}
\label{Analysis}

We conduct a thorough analysis of the generalization ability of PH3.
For cross-model generalization, PH3 is effective across a wide range of models.
This shows that our method is not limited to small models, but can also be adopted on relatively large models, including the popular LLaMA2 series.
For cross-relation generalization, by intervening on the attention heads discovered on \texttt{World Capital}, our method can also well resolve knowledge conflicts on other relation types.
This indicates that PH3 does not identify attention heads specific to a certain type of relation. Instead, it identifies universal memory and context heads.
For cross-format generalization, PH3 can transfer well from triple-form context to document-form context.
This indicates that our method does not merely remember the relative positions of elements in context, but is capable of understanding the external context.
Compared to the Gradient, our method has demonstrated superior generalizability.
We also analyze the impact of the number of pruning heads in Appendix \ref{ANumber}.

\section{Conclusion}

In this paper, we perform an exploration into the mechanism of interpreting knowledge conflicts and reveal that memory and context heads in later layers can cause knowledge conflicts when merging inconsistent information flows.
Based on our insights, we propose a novel method called \textbf{P}runing \textbf{H}ead via \textbf{P}at\textbf{H} \textbf{P}atc\textbf{H}ing (\textbf{PH3}), which can mitigate knowledge conflicts by pruning those conflicting attention heads.
We prove that PH3 can flexibly control LMs to use internal memory or external context.
We also demonstrate the cross-model, cross-relation, and cross-format generalization.

\section*{Limitations}

For further study, we conclude some limitations of our work as follows:

\begin{itemize} 

\item Similar to previous works on mechanism interpretability that adopt tasks such as antonym generation \cite{todd2023function}, fact recall \citep{meng2022locating, geva-etal-2023-dissecting}, arithmetic operation \citep{hanna2023how, stolfo-etal-2023-mechanistic}, and text classification \citep{bansal-etal-2023-rethinking, wang-etal-2023-label}, our work also selects a relatively simpler task to interpret the mechanism behind knowledge conflicts.
Simple tasks enable us to better control variables and minimize external distractions.
In the future, we plan to extend our analysis to more complex and realistic scenarios, such as where irrelevant information is present within the external context, or where the model needs to reason with both internal and external knowledge.

\item Although our research has delved into the attention heads in LMs, there may be more basic elements involved in knowledge conflicts.
Furthermore, the memory and context heads we have discovered may not only be responsible for extracting knowledge from internal memory or external context.
These heads may also have other functions, such as helping the model capture global dependencies of input texts.
By pruning these heads, the original capabilities of the model may be affected.
Therefore, we will further explore mitigating knowledge conflicts through more subtle intervention methods.

\end{itemize}

In summary, the mechanism behind knowledge conflicts remains a largely unexplored area, and we hope our work can offer some useful insights for further research.

\section*{Ethics Statement}

To enhance the reproducibility of our research, we will make all source code and datasets publicly available upon the acceptance of this paper.
Our work focuses on uncovering the mechanisms behind knowledge conflicts in LM, thereby better controlling the model in retrieval augmentation and tool augmentation.
Through effective intervention, our method can make the LM more controllable and trustworthy.
On the one hand, it can prevent prompt injections from attacking the model, and on the other hand, it can correct the biased knowledge that the model learned during pre-training.
Nonetheless, the impact of head pruning on the model's original capabilities remains unexplored.
These factors should be taken into careful consideration for future research.

\bibliography{anthology,custom}

\clearpage

\appendix

\section{Related Work}
\label{related}

\subsection{Investigating Knowledge Conflict}

Previous research \citep{longpre-etal-2021-entity, chen-etal-2022-rich, yu-etal-2023-characterizing,  xie2023adaptive, wang2023resolving, neeman-etal-2023-disentqa, jin2024tug} on knowledge conflicts primarily seek to answer the question: \textit{do language models prefer internal memory or external context?}
\citet{yu-etal-2023-characterizing} find that language models are more inclined to internal memory as the frequency of a fact in the pre-training corpus increases.
\citet{xie2023adaptive} demonstrate that large language models (LLMs) are highly receptive to external conflicting evidence.
They also reveal that when both supportive and contradictory evidence to their internal memory are present, LLMs show a strong confirmation bias and tend to cling to their parametric memory.
The above observed phenomena contribute to a better understanding of knowledge conflicts.
However, the underlying mechanism of knowledge conflicts remains unclear.
We observe that knowledge conflicts arise when the late attention heads integrate different information flows from internal memory and external context.

\subsection{Resolving Knowledge Conflict}
Existing work \cite{shi2023trusting, zhou-etal-2023-context,li-etal-2023-large, yu-etal-2023-characterizing, Qian2023MergeCE} has conducted preliminary exploration into the mitigation of knowledge conflicts.
\citet{shi2023trusting} propose a simple method to encourage the LM to attend to the external context via contrastive decoding \cite{li-etal-2023-contrastive}.
\citet{yu-etal-2023-characterizing} use head attribution to identify individual attention heads that either promote the memorized answer or the in-context answer, then scale the value vector of these heads to increase the rate of the in-context answers.
Our work is inspired by their exploration of attention heads, and we propose further analysis to improve understanding of the way knowledge conflicts are formed.
Furthermore, while most existing methods \citep{shi2023trusting, yu-etal-2023-characterizing} primarily focus on improving the model's faithfulness to the context, enabling the model to adhere to its internal memory remains a challenging task.

\subsection{Mechanistic Interpretability}

Recently, there has been a growing interest in the mechanistic interpretability \citep{cammarata2020thread, elhage2021mathematical} of parametric knowledge in LMs, with efforts focusing on reverse engineering the computational processes of model parameters.
\citet{dai-etal-2022-knowledge} use a knowledge attribution method \cite{Hao_Dong_Wei_Xu_2021} to identify the knowledge neurons in FFNs.
\citet{meng2022locating} reveal that FFNs at a range of middle layers can recall facts by using the causal mediation analysis method \cite{NEURIPS2020_92650b2e}.
\citet{geva-etal-2023-dissecting} find that knowledge extraction is typically done via attention heads.
Besides, there are some works investigating LMs in mathematical reasoning \cite{hanna2023how, stolfo-etal-2023-mechanistic} and in-context learning \cite{hendel-etal-2023-context, olsson2022context, bansal-etal-2023-rethinking}.
Besides, there are some studies \citep{yang2023bias, sakarvadia-etal-2023-memory, sakarvadia2023attention, zhang2024best} focused on interpreting attention heads in LMs.
Our work is highly inspired by previous wisdom in mechanistic interpretability, focusing on interpreting and mitigating knowledge conflicts in LMs.

\section{Implement Details}
\label{Details}

\subsection{Datasets}
We construct \texttt{Official Language}, \texttt{Country}, and \texttt{Continent} datasets by sampling knowledge
triples from Wikidata.
The \texttt{Official Language} dataset requires the LM to predict the official language of the given city or country:
\begin{center}
\scalebox{1.0}{
    \fbox{
    \shortstack[c]{
    \textit{The official language of} \{$s$\} \textit{is} \{$a_{c}$\}\textit{.}
    \\\textit{Q: What is the official language of} \{$s$\}\textit{? A:}
    }
    }}
\end{center}
The \texttt{Country} dataset requires the LM to predict the country to which the given city belongs:
\begin{center}
\scalebox{1.0}{
    \fbox{
    \shortstack[c]{
        \textit{The city} \{$s$\} \textit{is located in} \{$a_{c}$\}\textit{.}\\
    \textit{Q: Which country is the city} \{$s$\}\textit{ in ? A:}
    }
    }}
\end{center}
The \texttt{Continent} dataset requires the LM to predict the continent on which the given country is located:
\begin{center}
\scalebox{1.0}{
    \fbox{
    \shortstack[c]{
            \{$s$\} \textit{is in the continent of} \{$a_{c}$\}\textit{.}\\
    \textit{Q: Which continent is} \{$s$\}\textit{ located in ? A:}
    }
    }}
\end{center}
We also generate a more complex \texttt{World Capital D} dataset based on the \texttt{World Capital} dataset, using \texttt{gpt-3.5-turbo} to rewrite the external context from triplet form into document form.

\begin{figure}[t]

    \centering
    \includegraphics[width=0.485\textwidth]{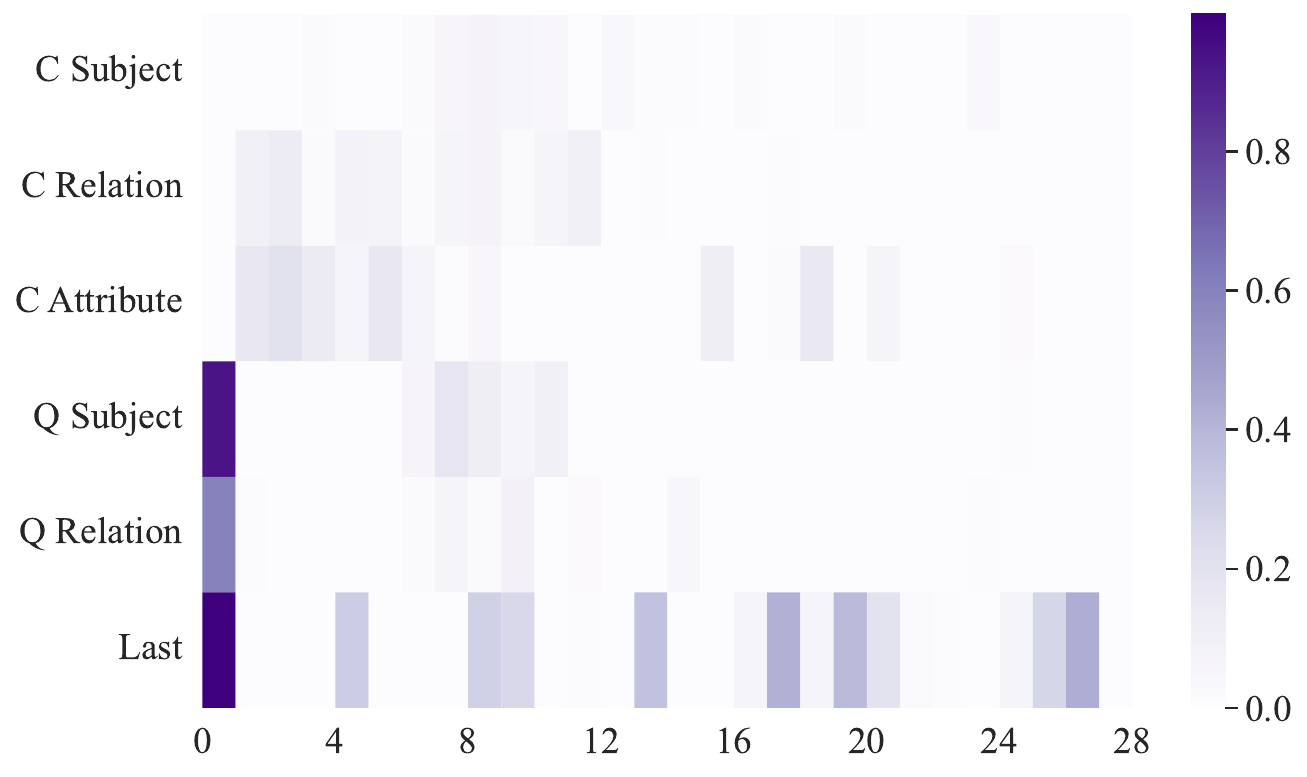}
    \caption{Effect of FFNs in GPT-J on internal memory.}
    \label{destroy-capital-memory-gptj-mlp}

\end{figure}

\begin{figure}[t]

    \centering
    \includegraphics[width=0.49\textwidth]{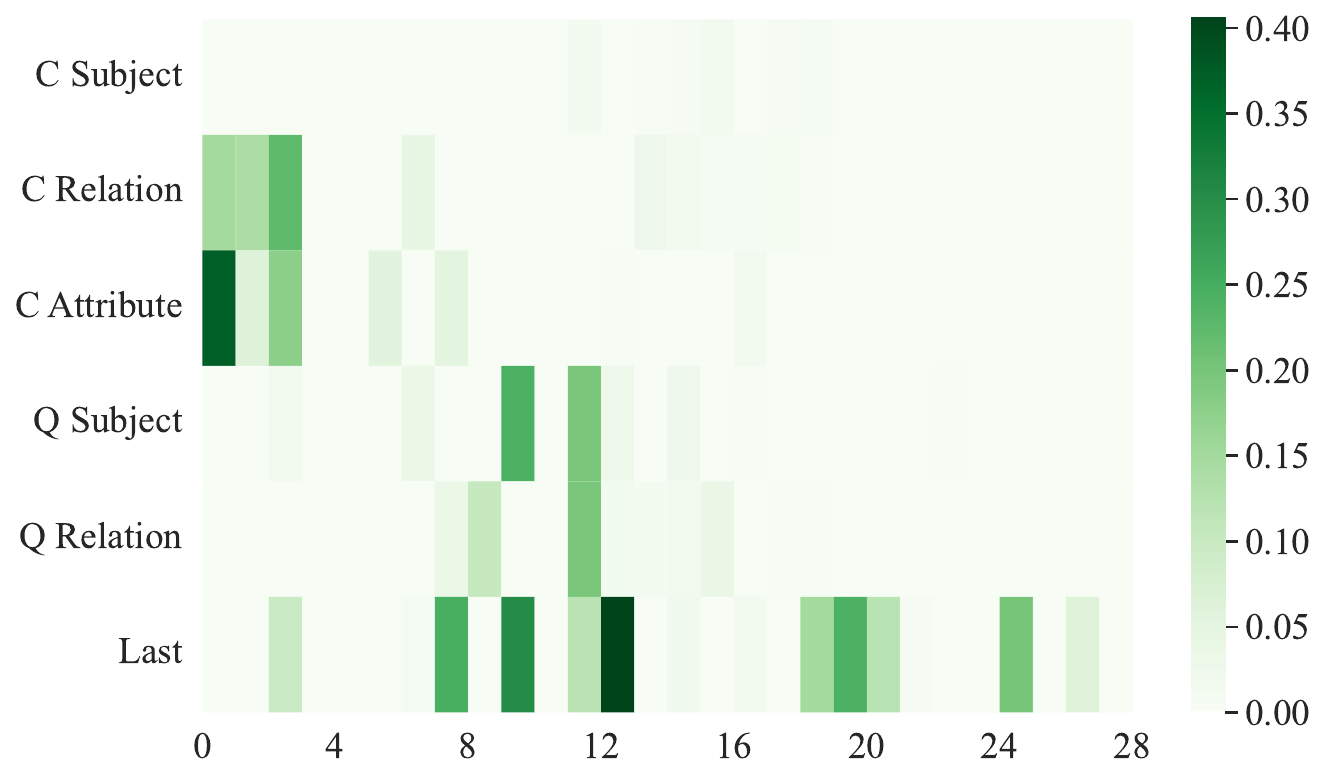}
    \caption{Effect of MHAs in GPT-J on internal memory.}
    \label{destroy-capital-memory-gptj-attn}

\end{figure}

\begin{figure}[t]

    \centering
    \includegraphics[width=0.50\textwidth]{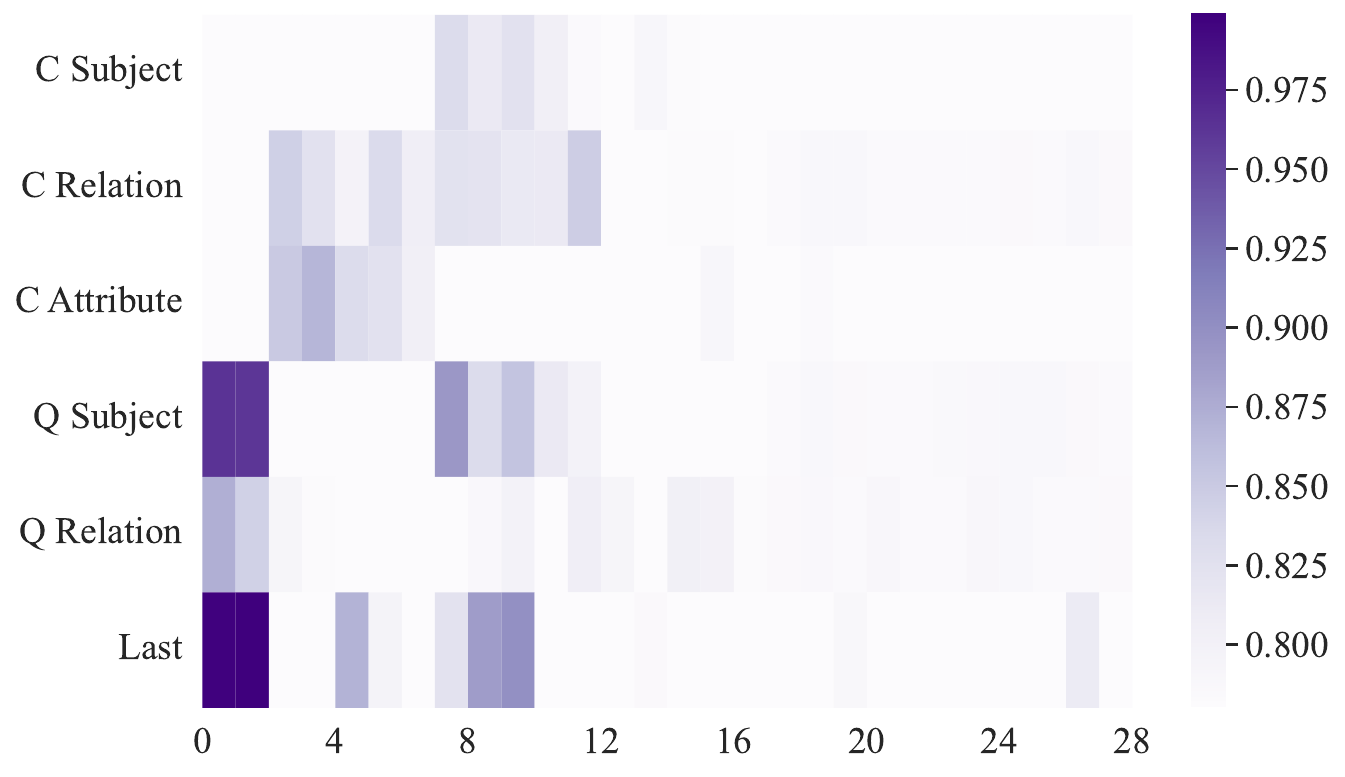}
    \caption{Effect of FFNs in GPT-J on external context.}
    \label{destroy-capital-context-gptj-mlp}

\end{figure}

\begin{figure}[t]

    \centering
    \includegraphics[width=0.50\textwidth]{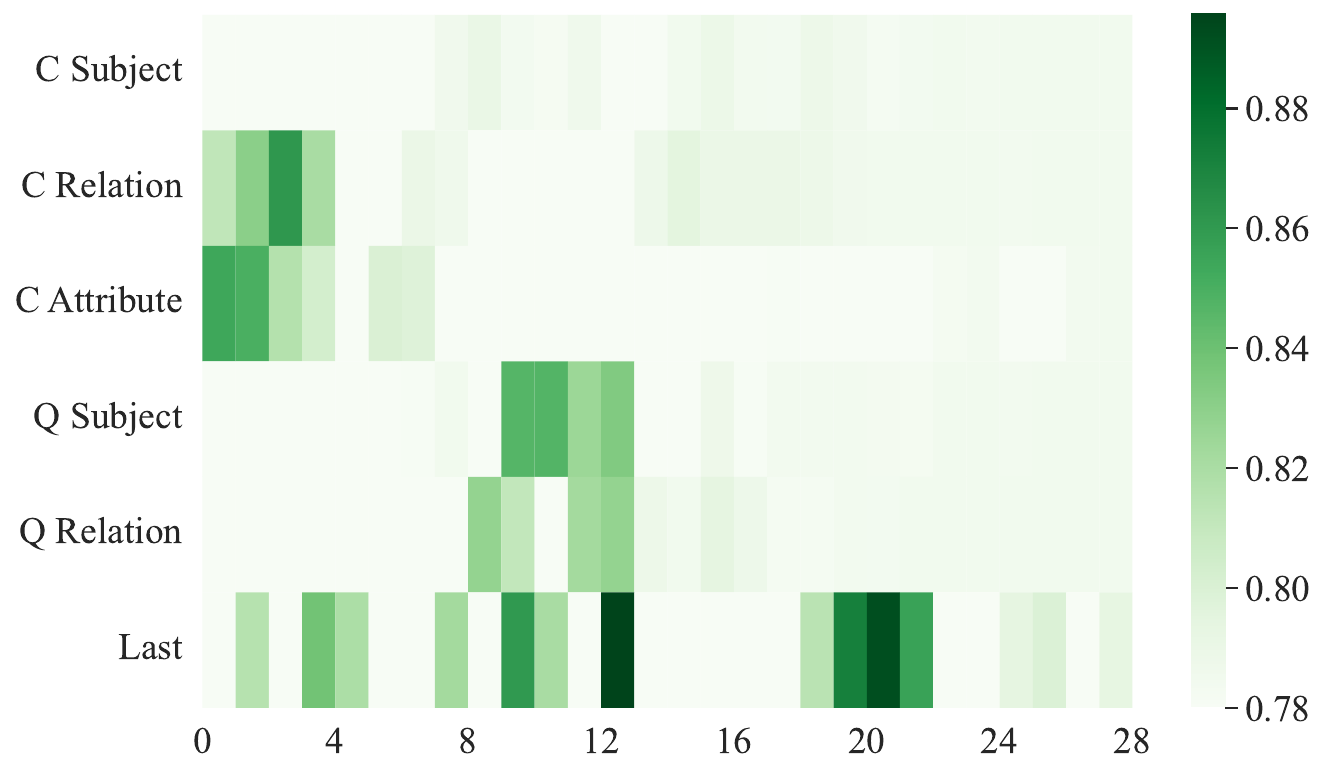}
    \caption{Effect of MHAs in GPT-J on external context.}
    \label{destroy-capital-context-gptj-attn}

\end{figure}


\subsection{Hyperparameter Settings}
\label{AHyperparameter}

Our implementation is based on HuggingFace’s Transformers\footnote{\url{https://github.com/huggingface/transformers/}}, PyTorch\footnote{\url{https://github.com/pytorch/pytorch/}} and baukit\footnote{\url{https://github.com/davidbau/baukit/}}.
For the Prompt method, we use the following prompt to enhance the internal memory:
\begin{center}
\scalebox{1.0}{
    \fbox{
    \shortstack[c]{
        \textit{Please answer the question based on your}\\ \textit{internal memory, ignoring the given context.}
    }
    }}
\end{center}
and we use the following prompt to enhance the external context:
\begin{center}
\scalebox{1.0}{
    \fbox{
    \shortstack[c]{
        \textit{Please answer the question based on the}\\ \textit{given context, ignoring your internal memory.}
    }
    }}
\end{center}
For Gradient and PH3, we select the optimal pruning rate $k \in \{1,3,5,7,9,15\}$ on the development set with 200 samples.
To mitigate knowledge conflicts, setting the pruning rate $k$ of PH3 to $5$ usually achieves excellent results.
For enhancing the open-domain QA capabilities, we usually set the pruning rate $k$ of PH3 to $3$.
Details about the models used in this paper are in Table \ref{model details}.
All experiments are conducted with NVIDIA GeForce RTX A6000 GPUs.


\section{Additional Results for GPT-J}
\label{GPT-J}

\begin{figure}[h]

    \centering
    \includegraphics[width=0.49\textwidth]{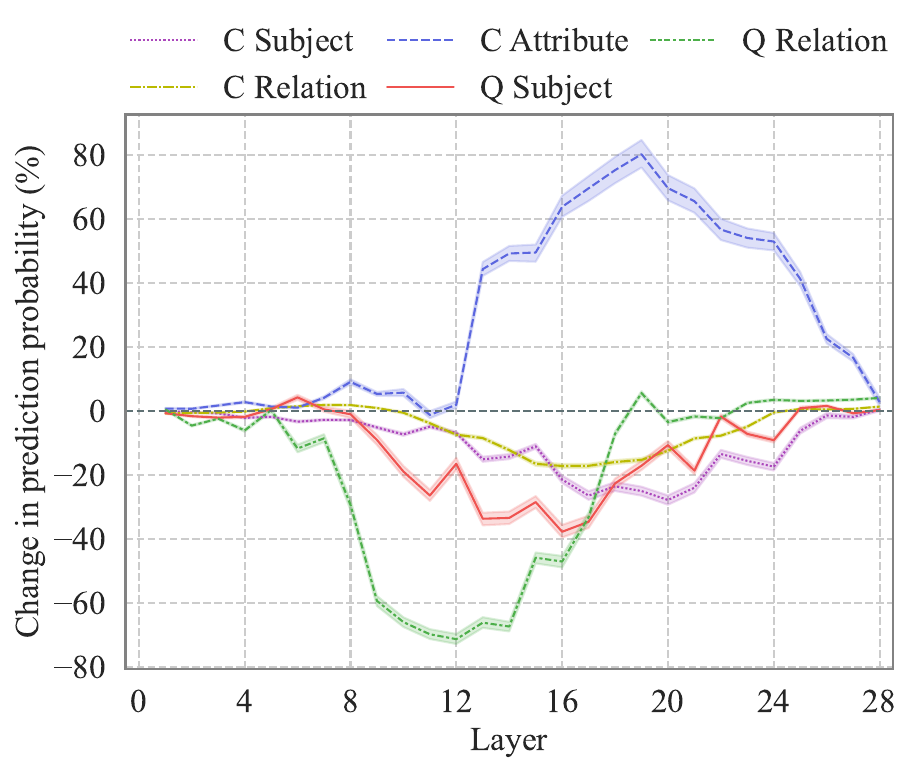}
    \caption{Relative change in the GPT-J's prediction probability based on internal memory.}
    \label{flow-capital-memory-gptj}
\vspace{-12pt}

\end{figure}

\begin{figure}[h]

    \centering
    \includegraphics[width=0.49\textwidth]{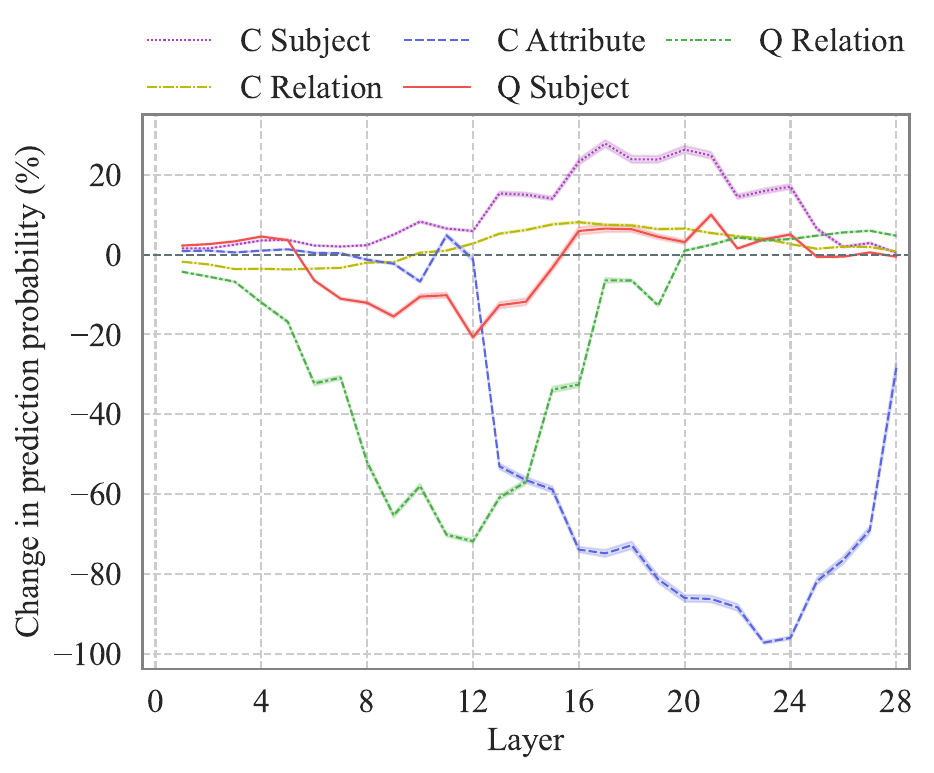}
    \caption{Relative change in the GPT-J's prediction probability based on external context.}
    \label{flow-capital-context-gptj}
\vspace{-12pt}

\end{figure}

\begin{figure}[h]

    \centering
    \includegraphics[width=0.49\textwidth]{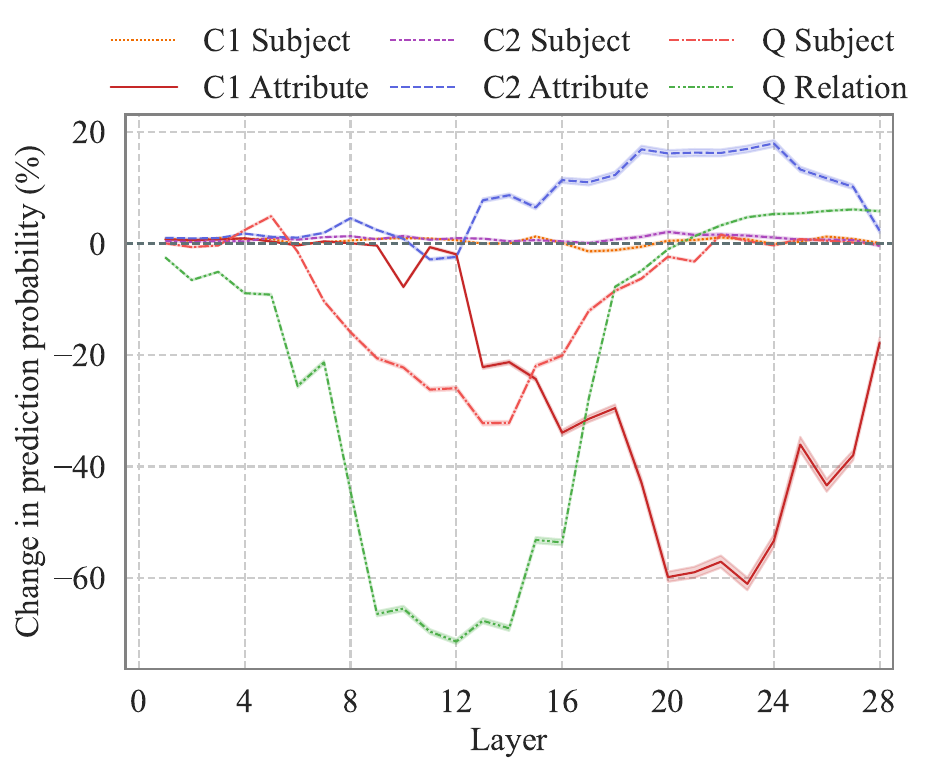}
    \caption{Relative change in the GPT-J's prediction probability based on internal memory when providing both supporting context and conflicting context.}
    \label{flow-capital-memory-gptj-2}
\vspace{-12pt}

\end{figure}

\begin{figure}[h]

    \centering
    \includegraphics[width=0.49\textwidth]{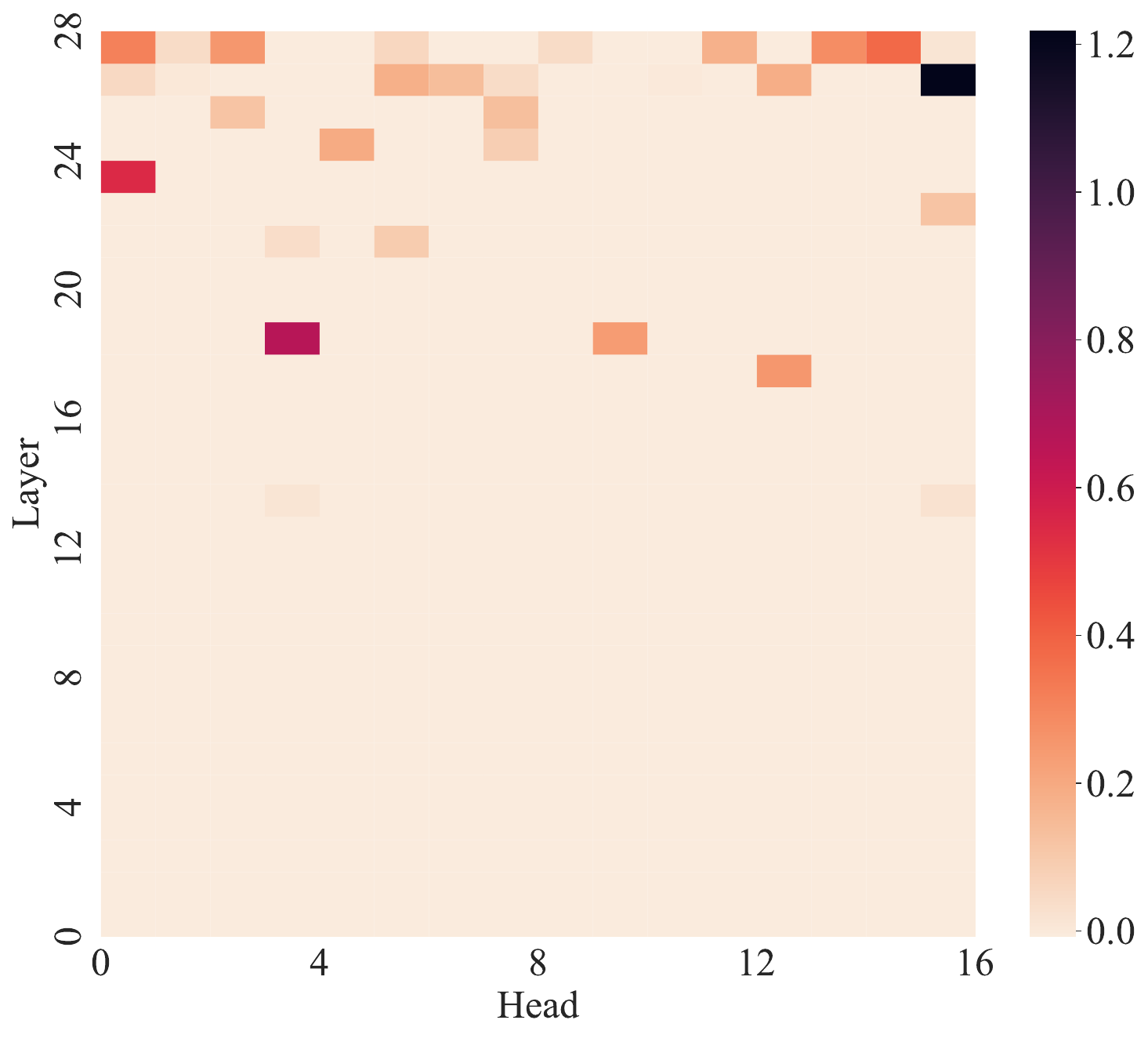}
    \caption{Memory Heads of GPT-J.}
    \label{important-memory-head-capital-gptj}
\vspace{-12pt}

\end{figure}

\begin{figure}[h]

    \centering
    \includegraphics[width=0.49\textwidth]{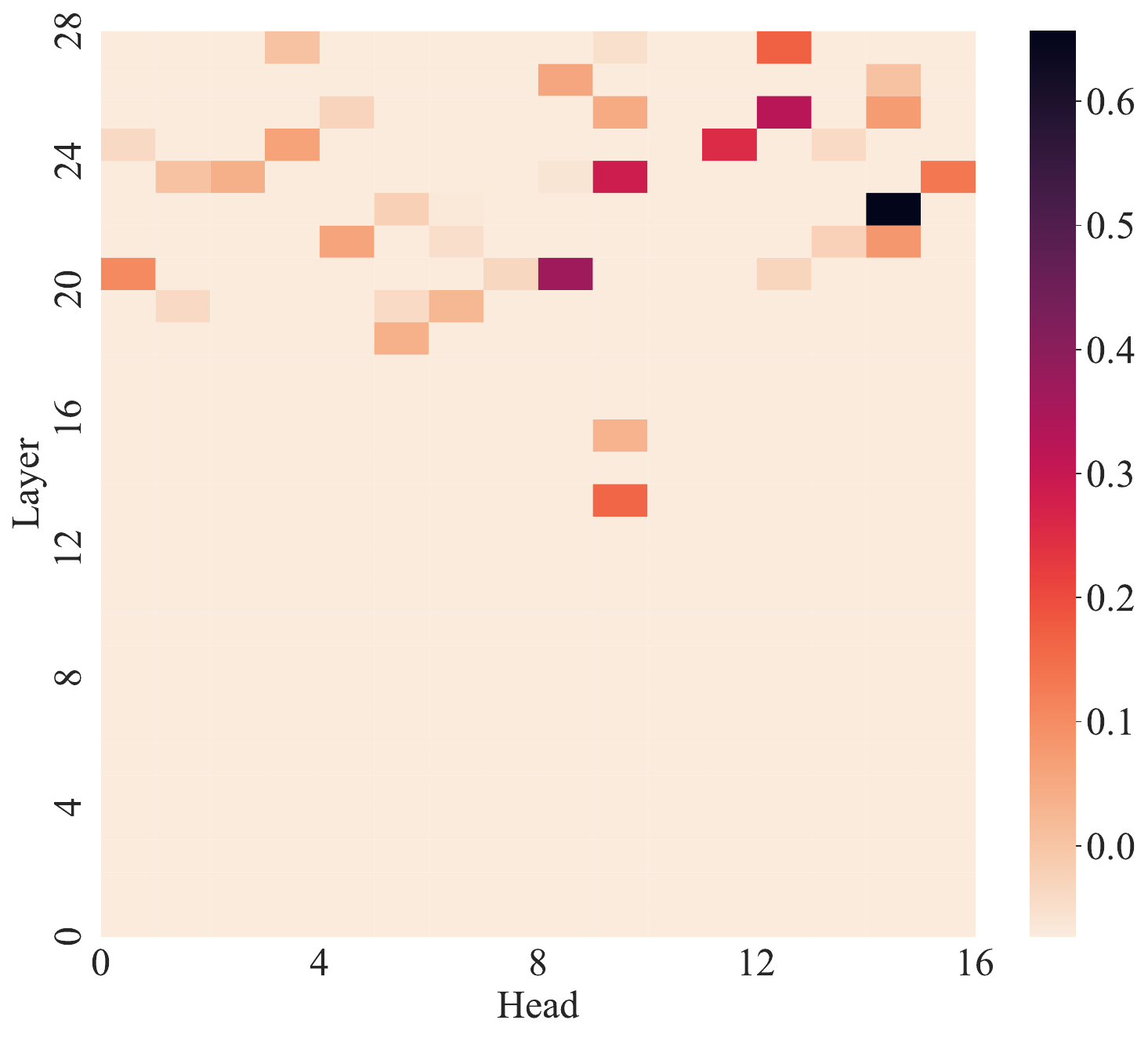}
    \caption{Context Heads of GPT-J.}
    \label{important-context-head-capital-gptj}
\vspace{-12pt}

\end{figure}

We provide here additional results for GPT-J.
Figures \ref{destroy-capital-memory-gptj-mlp} and \ref{destroy-capital-memory-gptj-attn} show the effect of FFNs and MHAs on internal memory, and Figures \ref{destroy-capital-context-gptj-mlp} and \ref{destroy-capital-context-gptj-attn} show the effect of FFNs and MHAs on external context.
Figures \ref{flow-capital-memory-gptj} and \ref{flow-capital-context-gptj} illustrate the information flow in GPT-J with the window size $W=9$.
Figure \ref{flow-capital-memory-gptj-2} shows the information flow in GPT-J when providing both supporting context and conflicting context relative to internal memory.
Figures \ref{important-memory-head-capital-gptj} and \ref{important-context-head-capital-gptj} show the gradient-based important scores of memory heads and context heads in GPT-J.

\section{Method Details}
\label{Amethod}
To calculate the important score $S_{m}^{\ell,h}$ of the target head $h$, our path patching method consists of the following three steps:
\begin{itemize}

\item[1.]  Run on the original input $x \in \mathcal{D}_{m}$ to record the original activations of all heads;
\item[2.]  Run on the corrupted input $\cancel{x}$ to record the corrupted activations of all heads, where $\cancel{x}$ is:
\begin{center}
\scalebox{1.0}{
    \fbox{
    \shortstack[c]{
    \textit{The capital of} $\langle$unk$\rangle$ \textit{is} \{$a_{c}$\}\textit{.}\\ \textit{Q: What is the capital of} $\langle$unk$\rangle$ \textit{? A:}}
    }}
\end{center}
where $\langle$unk$\rangle$ is the special token;
\item[3.] Run on the original input $x$, while keeping all the heads frozen to their activations on $x$, except for the target head $h$ whose activation is set on $\cancel{x}$.
Then measure the important score as the change of output logits.

\end{itemize}
The important score $S_{m}^{\ell,h}$ of head $h$ is computed as:
\begin{equation}
\begin{aligned}
S^{l,h}_{m}(\mathcal{D}_{m})=\mathbb{E}_{(x)}[\left(\mathbb{P}_{x}(a_{m})-\mathbb{P}_{x}(a_{c})\right) \\ -\left(\mathbb{P}_{\cancel{x}}(a_{m})-\mathbb{P}_{\cancel{x}}(a_{c})\right)].
\end{aligned}
\end{equation}

\begin{figure*}[t]
    \centering
        \includegraphics[width=0.98\textwidth]{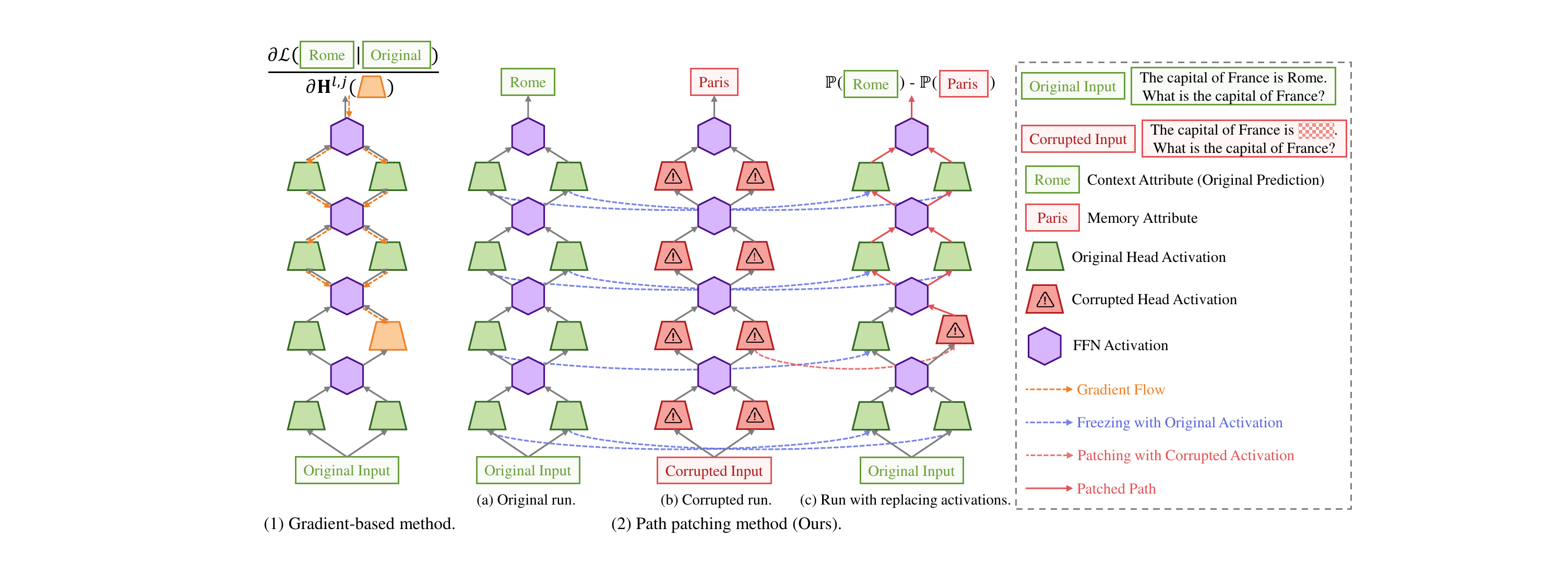}
        \caption{Illustration of gradient-based method and our path patching method.}
        \label{method}

\end{figure*}

\section{Heatmaps of Attention Heads}
\label{Aheat}
\begin{figure}[h]

    \centering
    \includegraphics[width=0.49\textwidth]{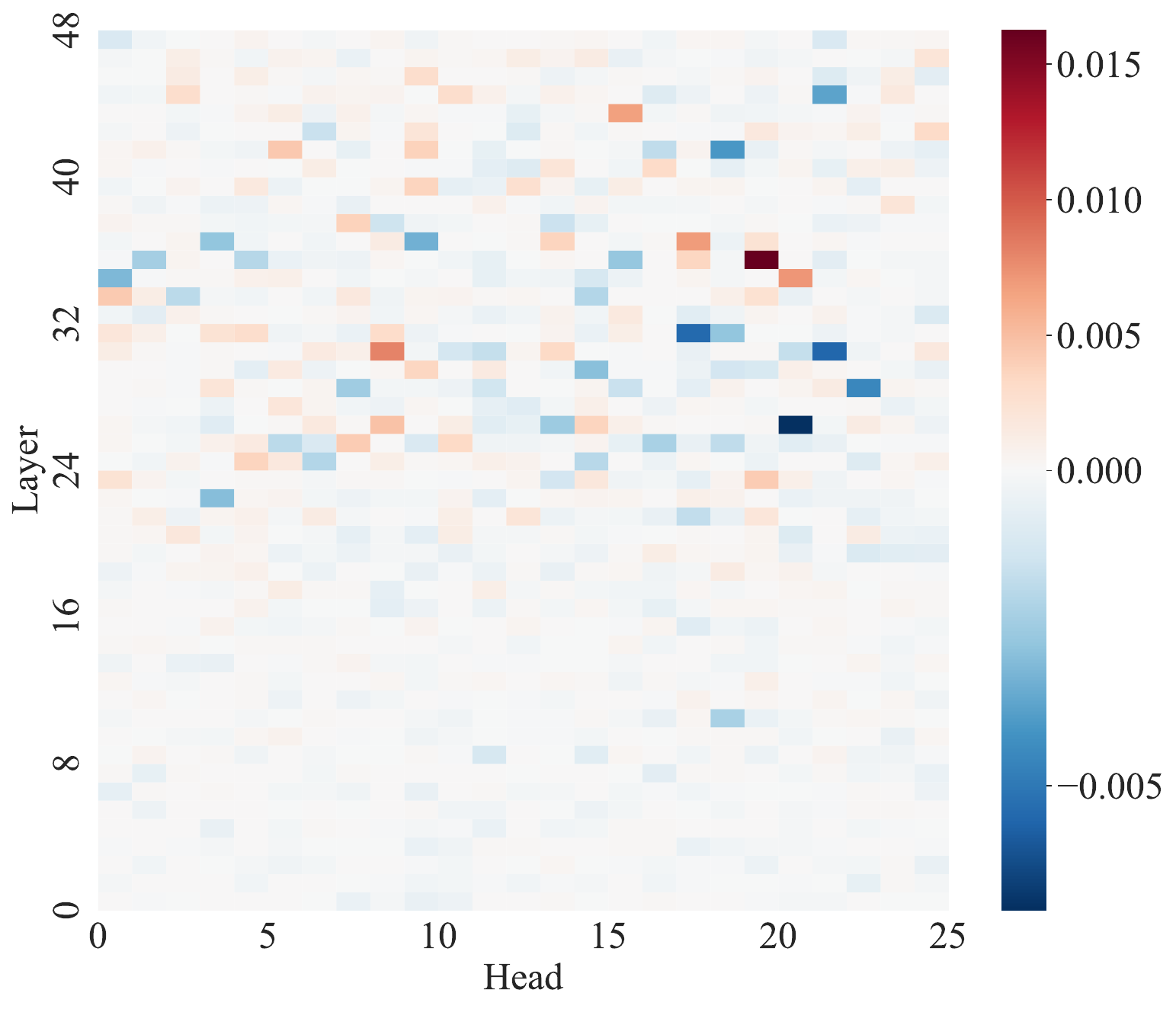}
    \caption{Memory Heads of GPT-2 XL.}
        \label{important-memory-head-capital-gpt2-xl-hb}

\end{figure}

\begin{figure}[h]

    \centering
    \includegraphics[width=0.49\textwidth]{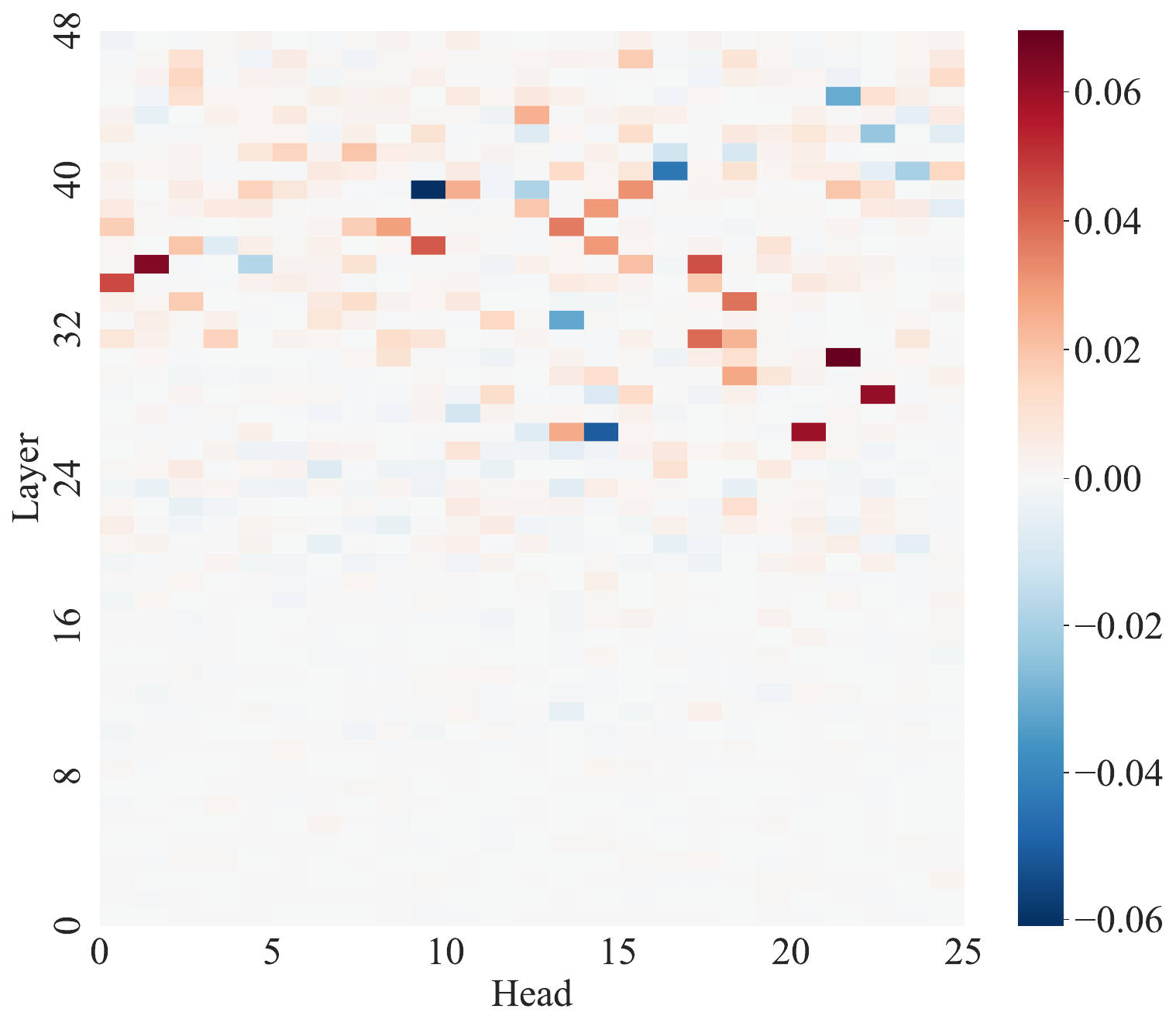}
    \caption{Context Heads of GPT-2 XL.}
        \label{important-context-head-capital-gpt2-xl-hb}

\end{figure}

\begin{figure}[h]

    \centering
    \includegraphics[width=0.49\textwidth]{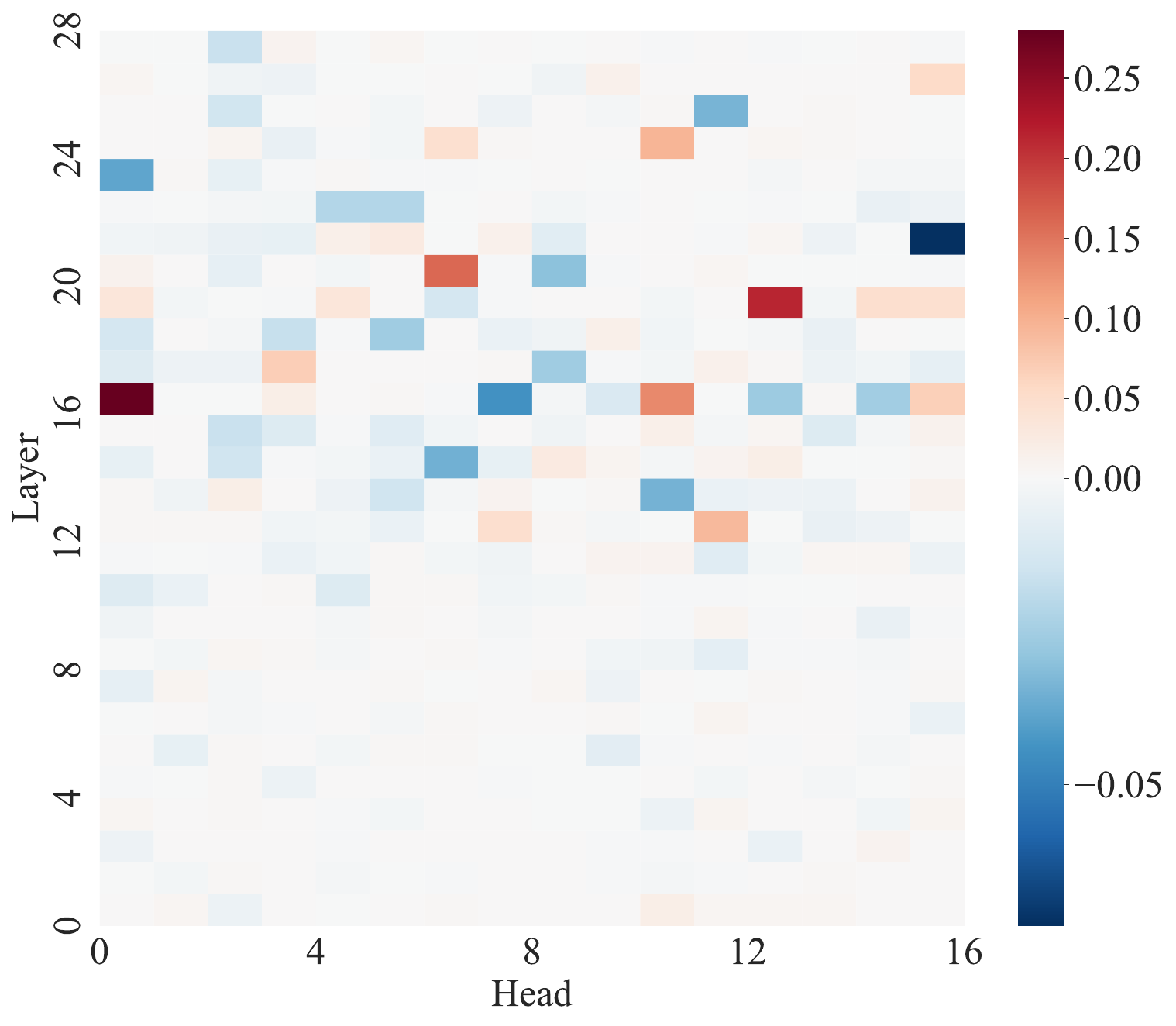}
    \caption{Memory Heads of GPT-J.}
        \label{important-memory-head-capital-gptj-hb}

\end{figure}

\begin{figure}[h]

    \centering
    \includegraphics[width=0.49\textwidth]{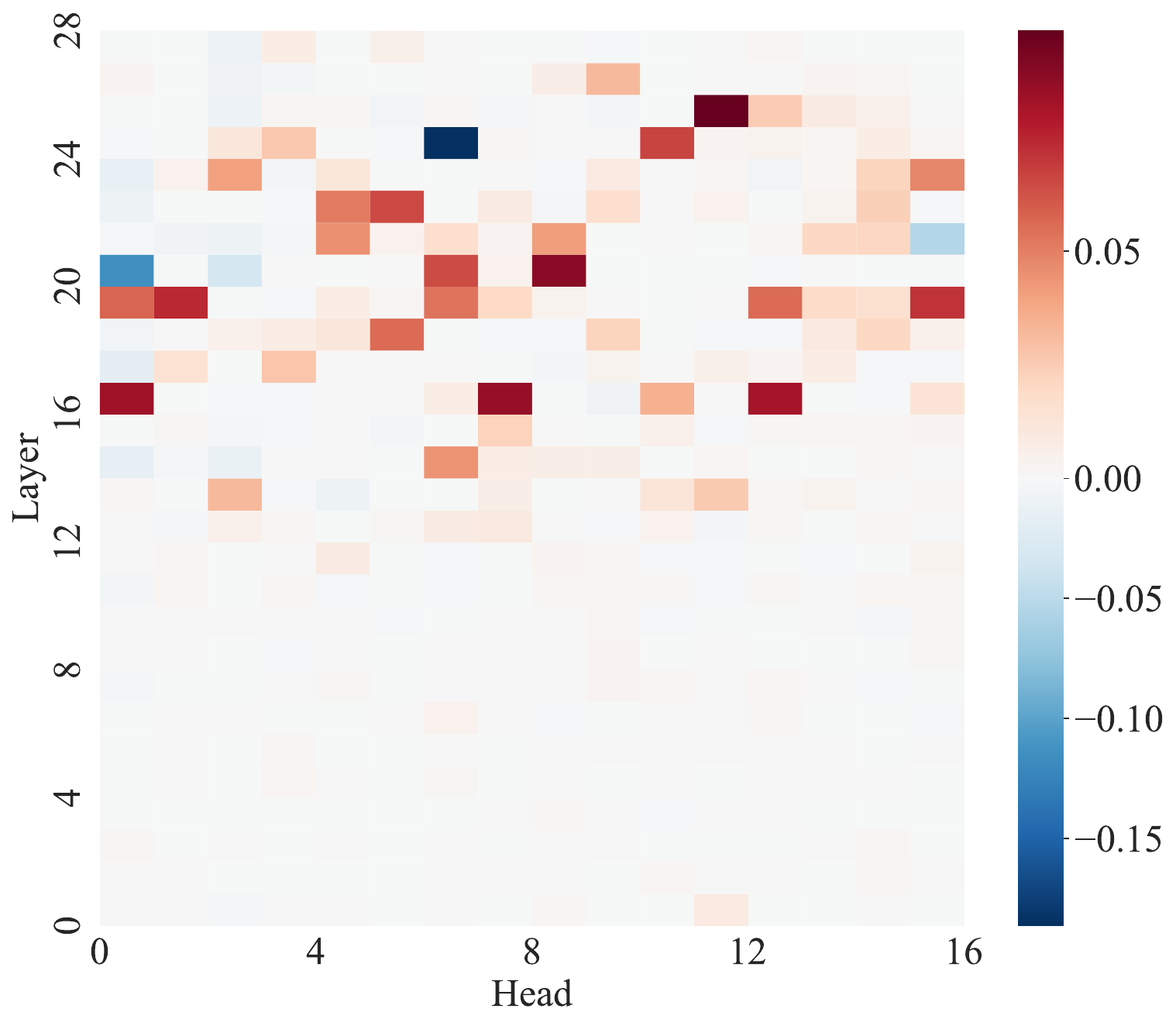}
    \caption{Context Heads of GPT-J.}
        \label{important-context-head-capital-gptj-hb}

\end{figure}

\begin{figure}[h]

    \centering
    \includegraphics[width=0.49\textwidth]{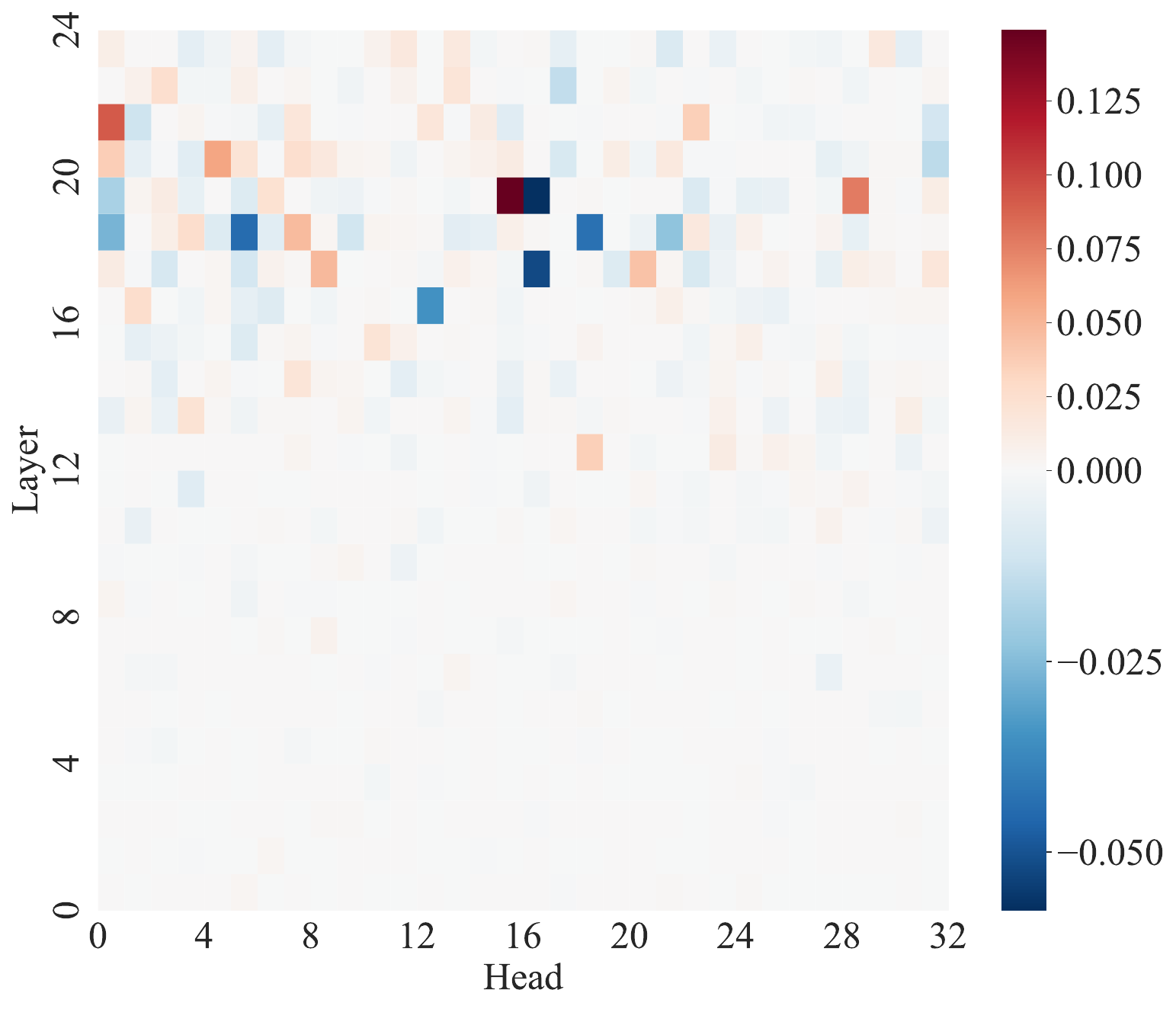}
    \caption{Memory Heads of OPT-1.3B.}
            \label{important-memory-head-capital-opt-1.3b-hb}

\end{figure}

\begin{figure}[h]

    \centering
    \includegraphics[width=0.49\textwidth]{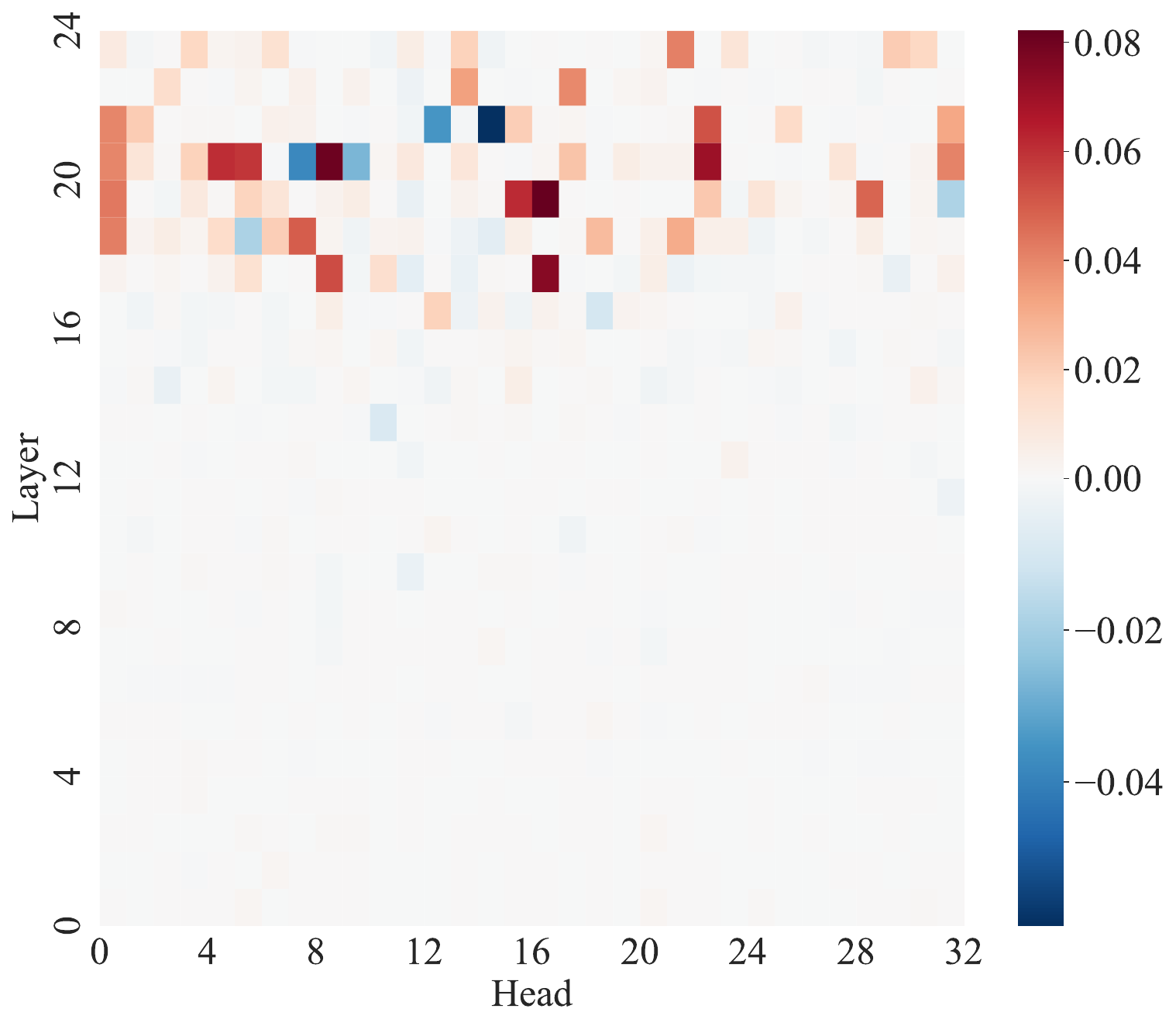}
    \caption{Context Heads of OPT-1.3B.}
            \label{important-context-head-capital-opt-1.3b-hb}

\end{figure}

\begin{figure}[h]

    \centering
    \includegraphics[width=0.49\textwidth]{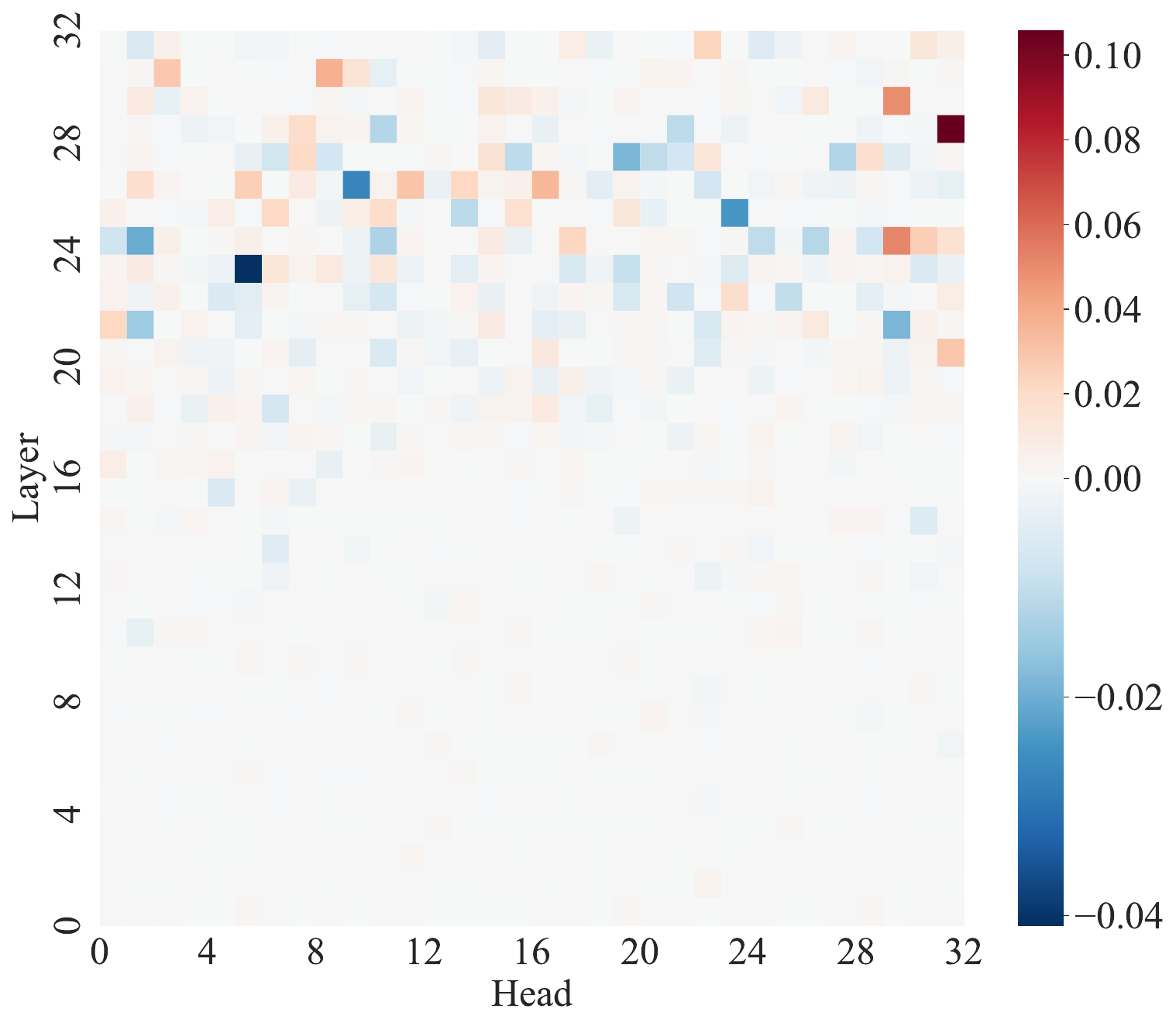}
    \caption{Memory Heads of OPT-2.7B.}
            \label{important-memory-head-capital-opt-2.7b-hb}

\end{figure}

\begin{figure}[h]

    \centering
    \includegraphics[width=0.49\textwidth]{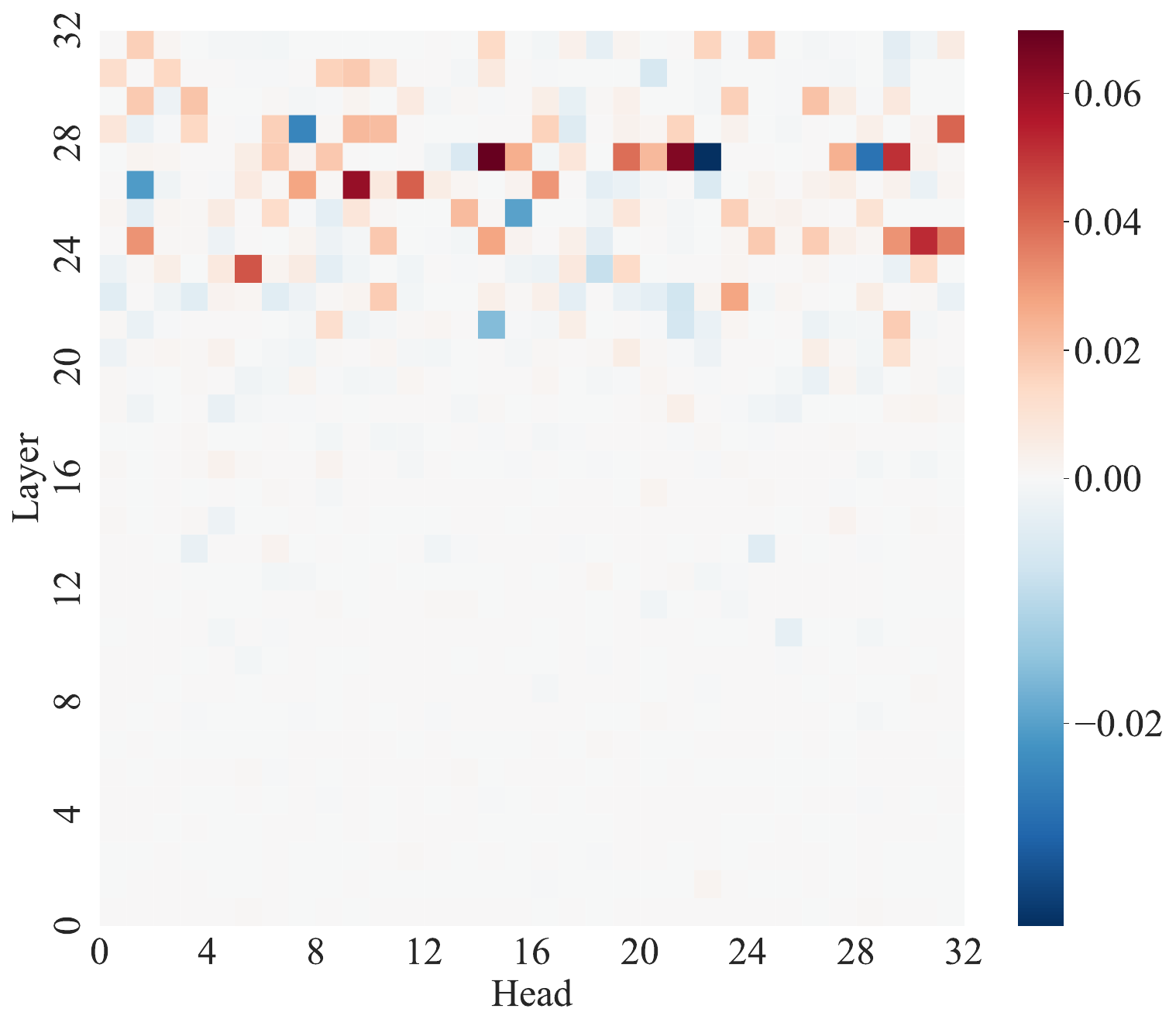}
    \caption{Context Heads of OPT-2.7B.}
            \label{important-context-head-capital-opt-2.7b-hb}

\end{figure}

\begin{figure}[h]

    \centering
    \includegraphics[width=0.49\textwidth]{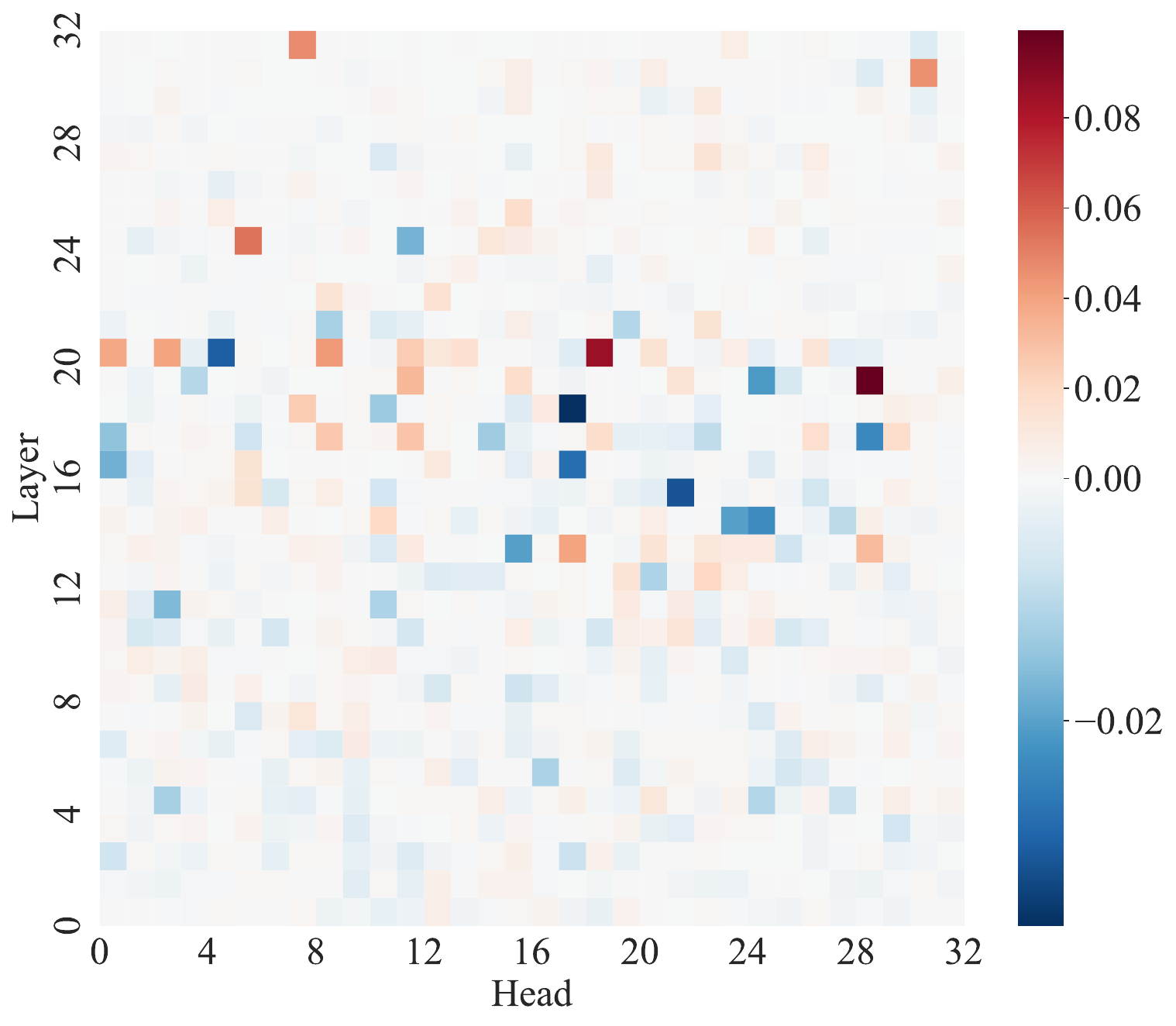}
    \caption{Memory Heads of Pythia-6.9B.}
            \label{important-memory-head-capital-pythia-6.9b-hb}

\end{figure}

\begin{figure}[h]

    \centering
    \includegraphics[width=0.49\textwidth]{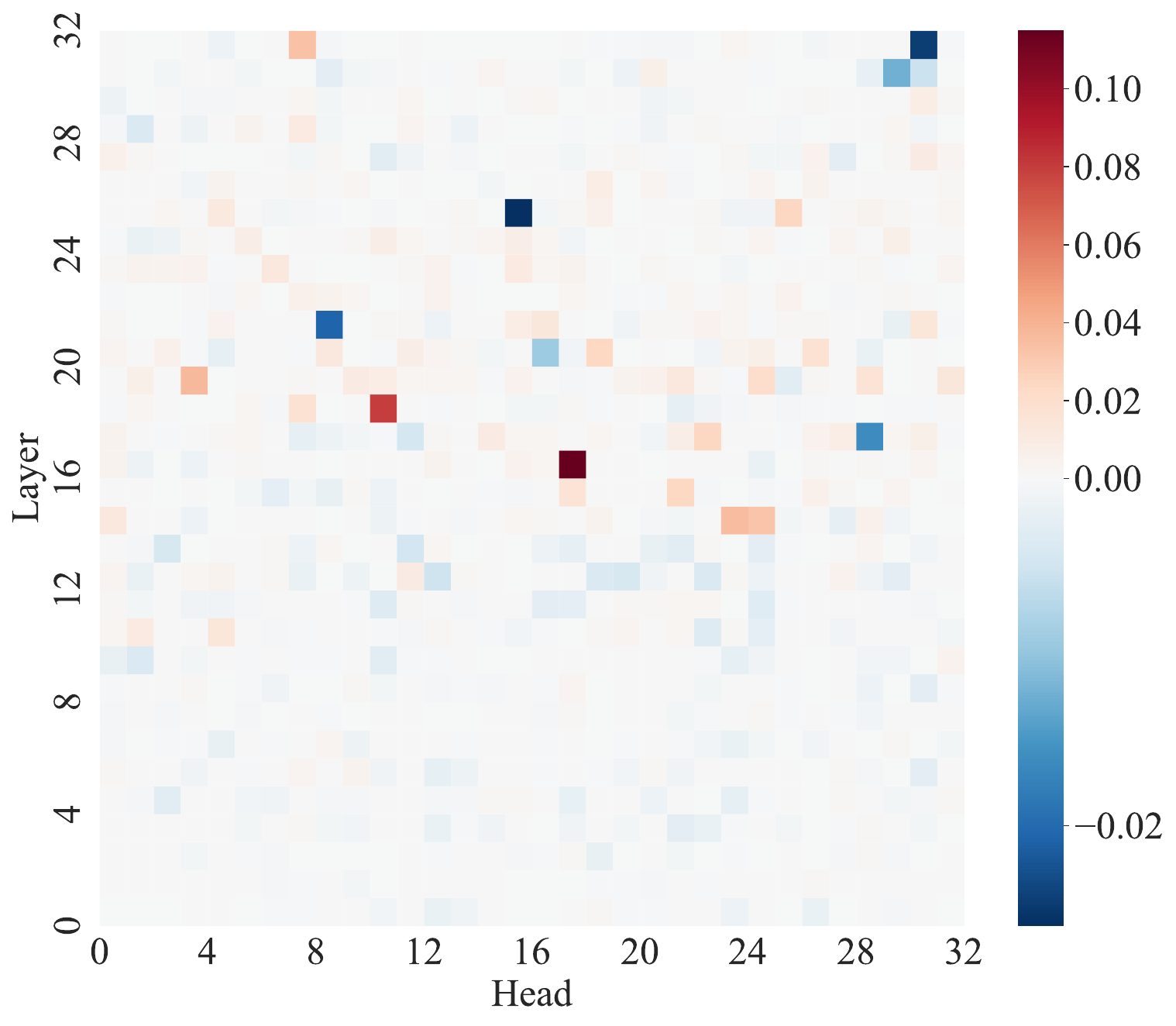}
    \caption{Context Heads of Pythia-6.9B.}
            \label{important-context-head-capital-pythia-6.9b-hb}

\end{figure}

\begin{figure}[h]

    \centering
    \includegraphics[width=0.49\textwidth]{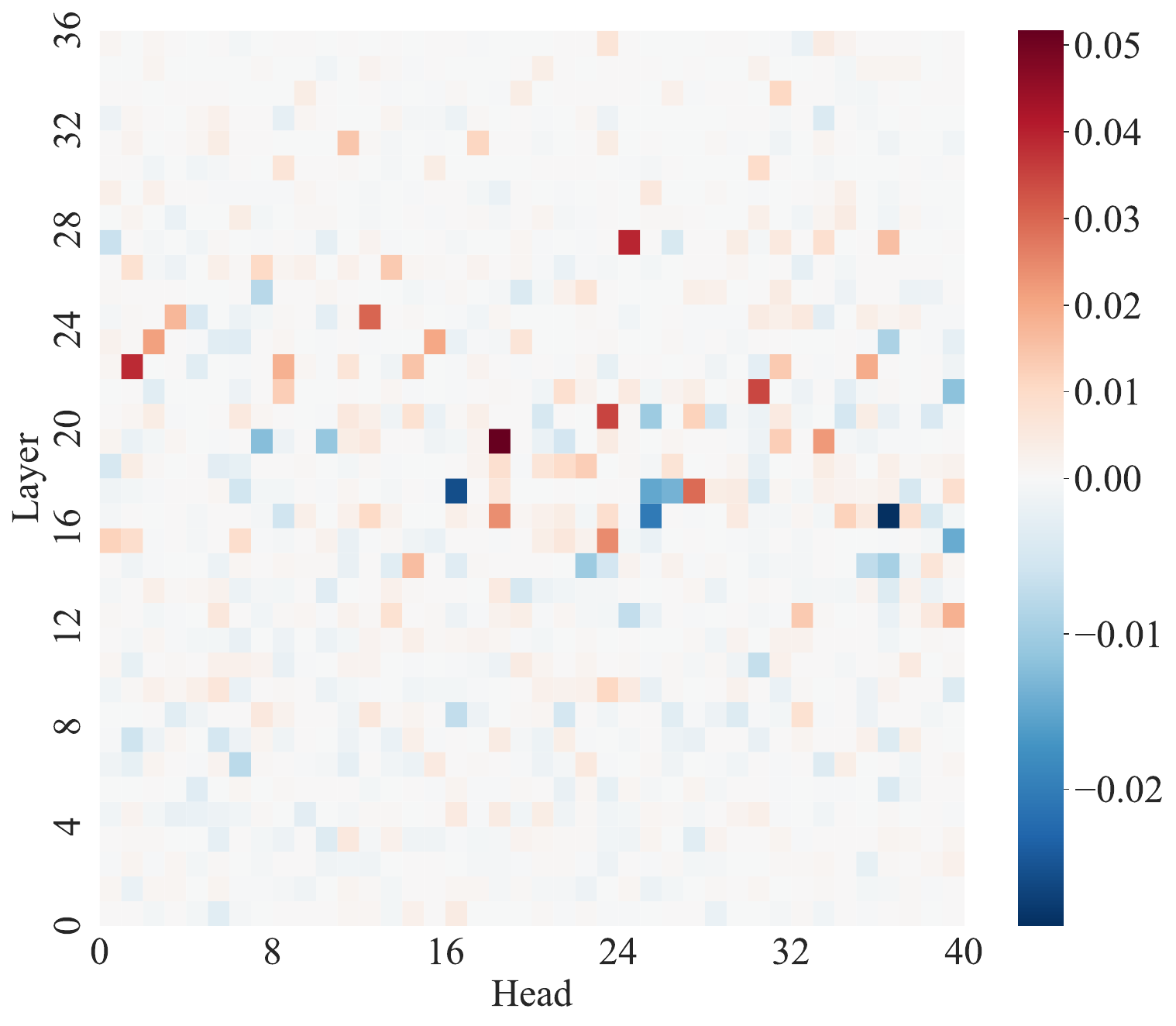}
    \caption{Memory Heads of Pythia-12B.}
            \label{important-memory-head-capital-pythia-12b-hb}

\end{figure}

\begin{figure}[h]

    \centering
    \includegraphics[width=0.49\textwidth]{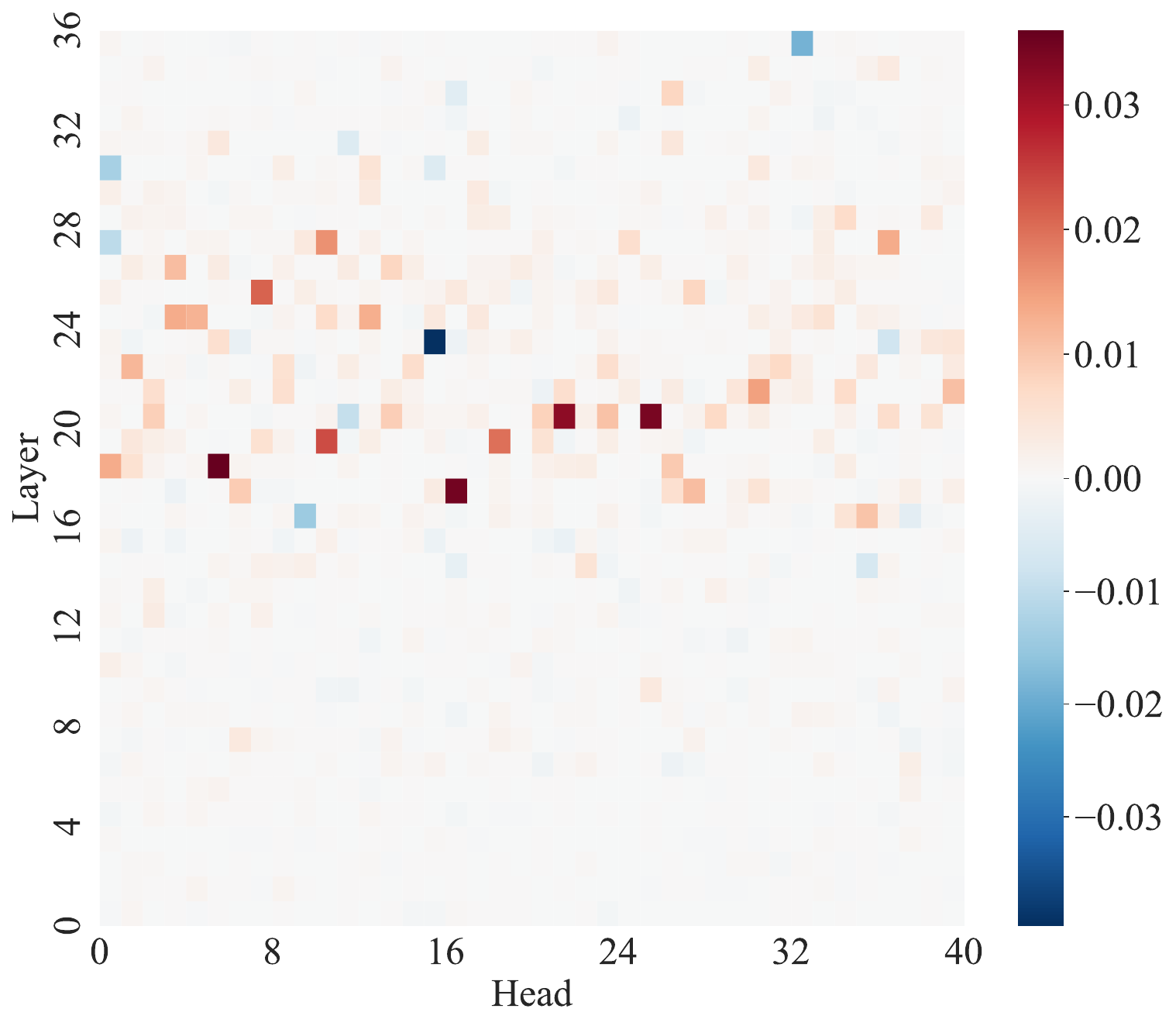}
    \caption{Context Heads of Pythia-12B.}
            \label{important-context-head-capital-pythia-12b-hb}

\end{figure}

\begin{figure}[h]

    \centering
    \includegraphics[width=0.49\textwidth]{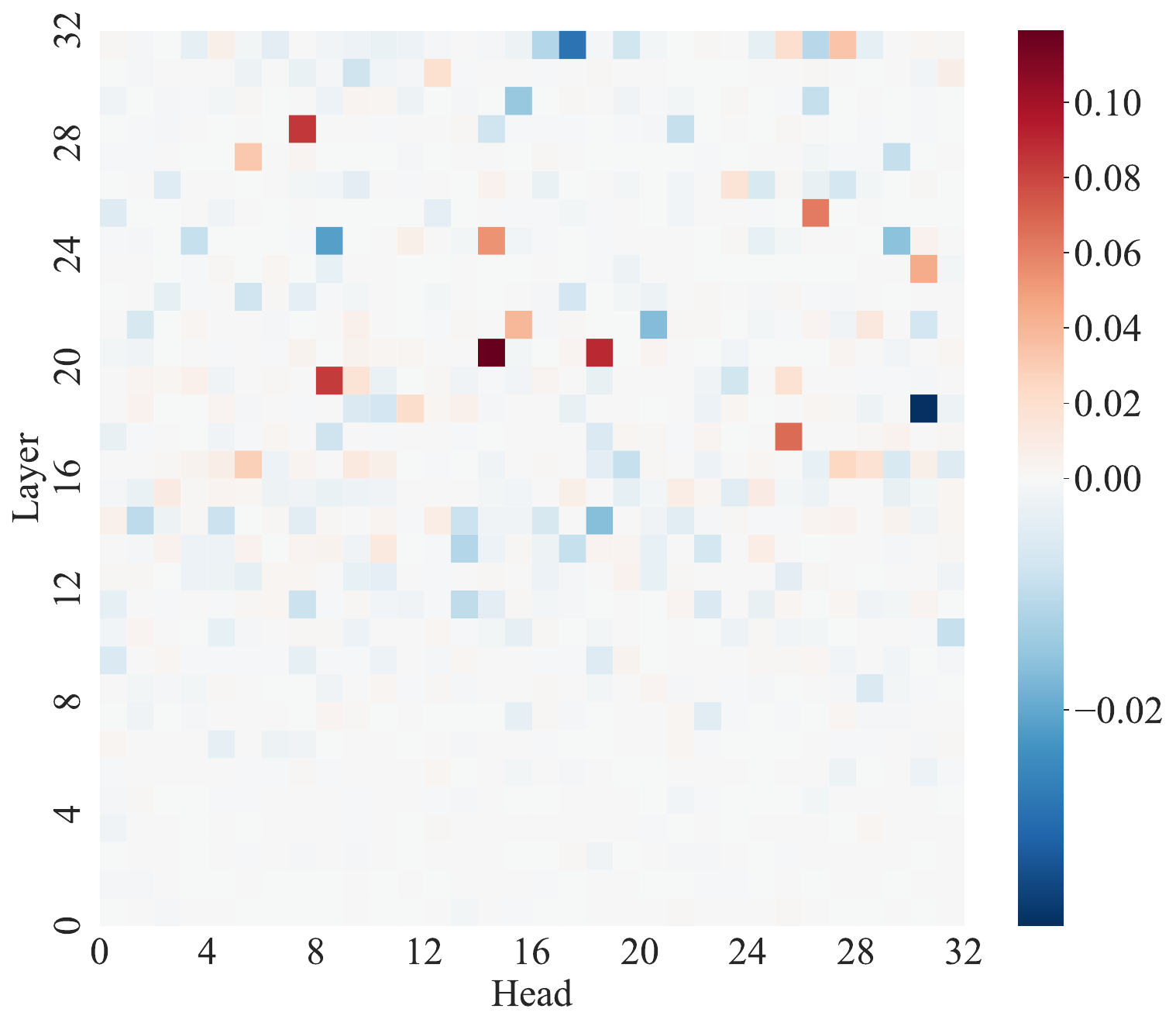}
    \caption{Memory Heads of LLaMA2-7B.}
            \label{important-memory-head-capital-llama2-7b-hb}

\end{figure}

\begin{figure}[h]

    \centering
    \includegraphics[width=0.49\textwidth]{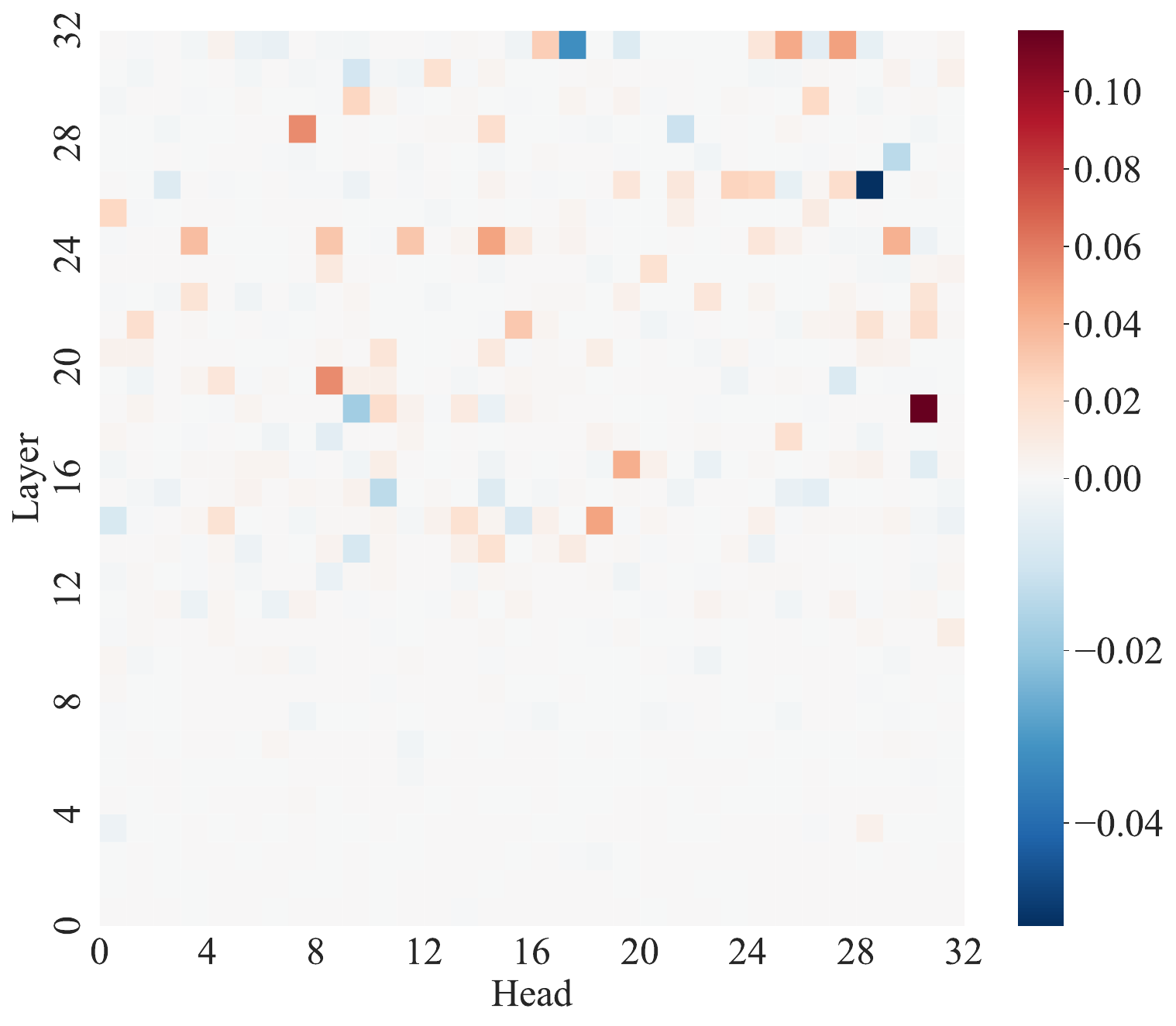}
    \caption{Context Heads of LLaMA2-7B.}
            \label{important-context-head-capital-llama2-7b-hb}

\end{figure}

\begin{figure}[h]

    \centering
    \includegraphics[width=0.49\textwidth]{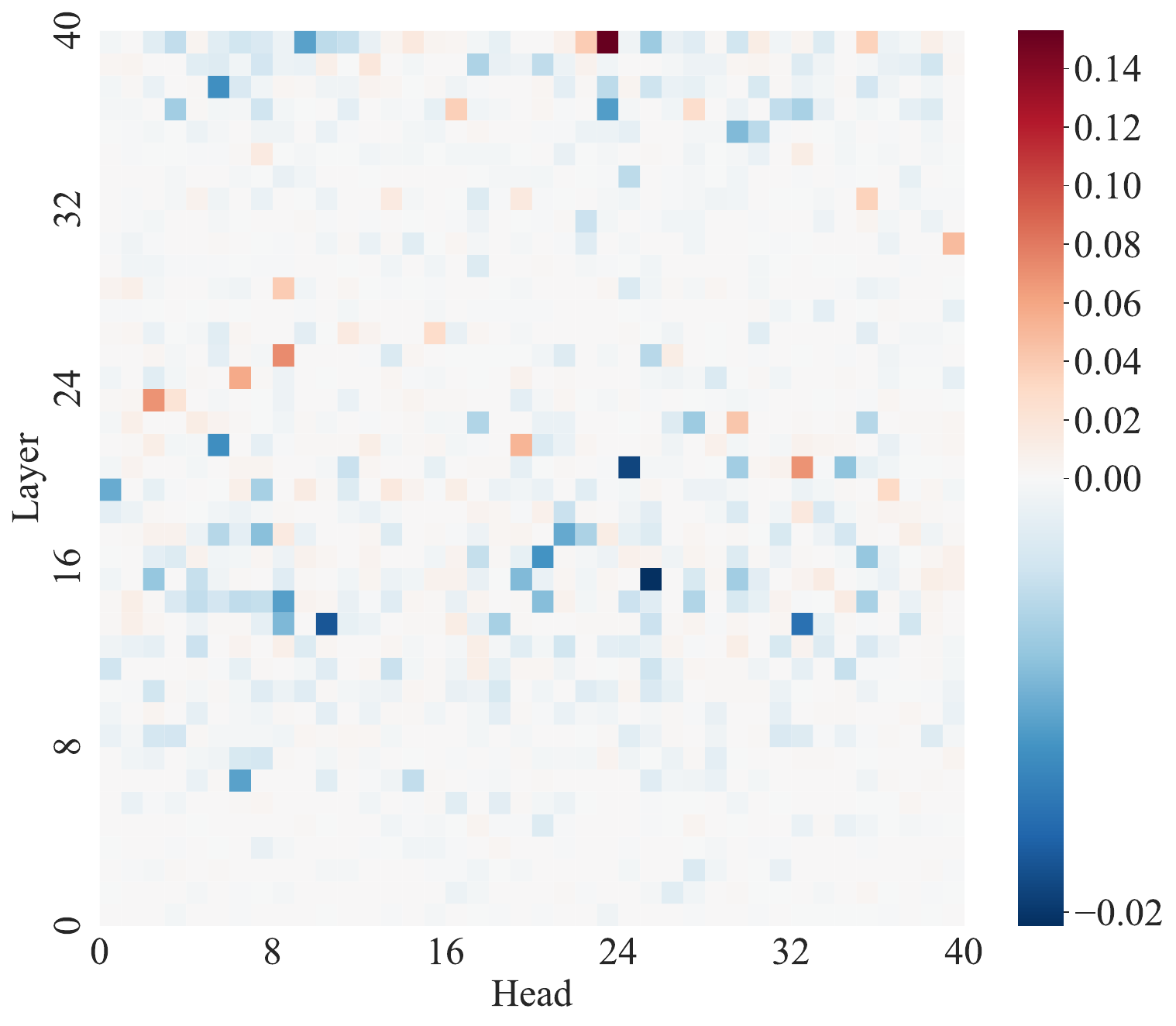}
    \caption{Memory Heads of LLaMA2-13B.}
            \label{important-memory-head-capital-llama2-13b-hb}

\end{figure}

\begin{figure}[h]

    \centering
    \includegraphics[width=0.49\textwidth]{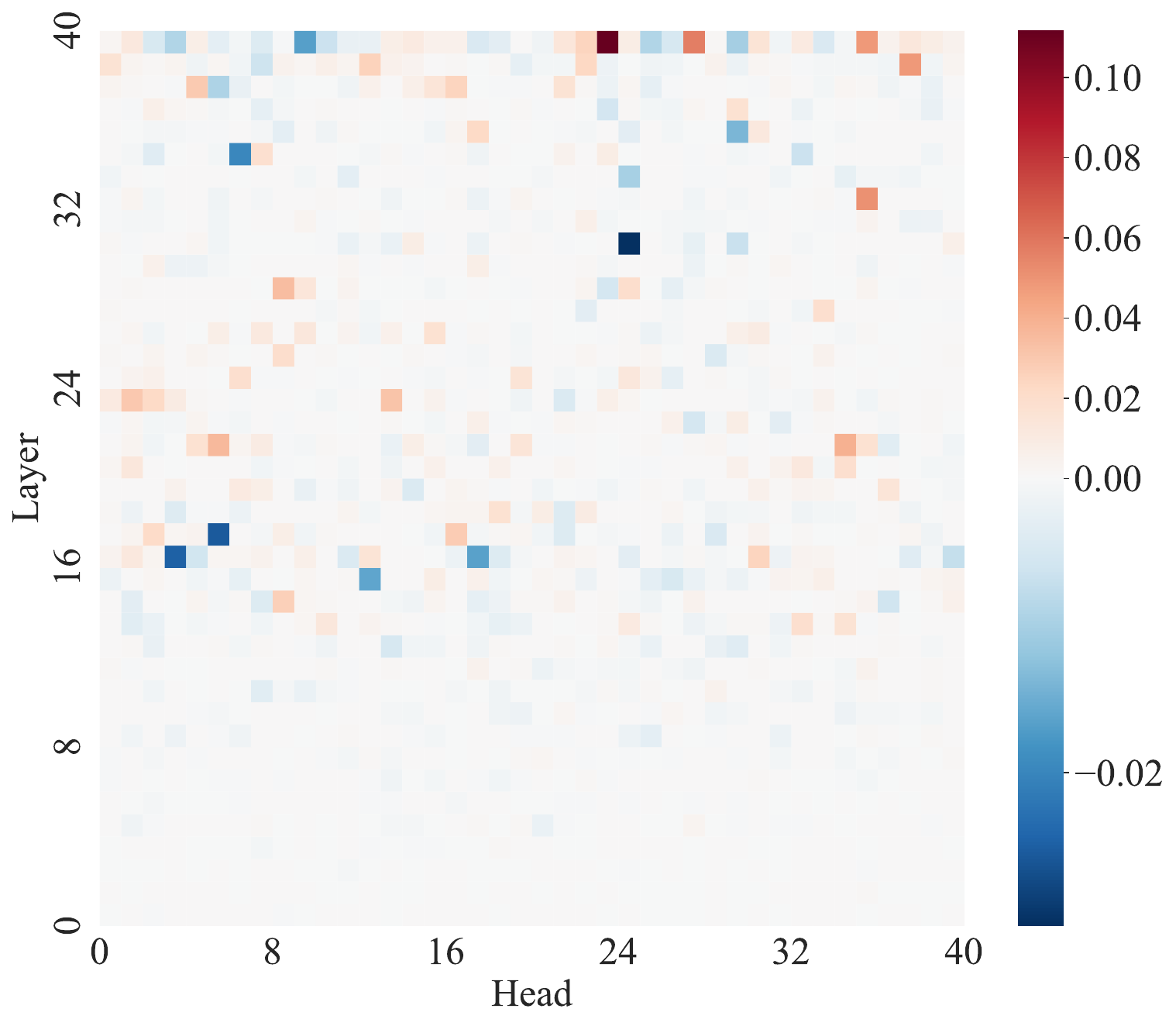}
    \caption{Context Heads of LLaMA2-13B.}
            \label{important-context-head-capital-llama2-13b-hb}

\end{figure}

We calculate the important scores of memory heads and context heads via our path patching method, then provide the heatmaps for GPT-2 XL (Figures \ref{important-memory-head-capital-gpt2-xl-hb} and \ref{important-context-head-capital-gpt2-xl-hb}), GPT-J (Figures \ref{important-memory-head-capital-gptj-hb} and \ref{important-context-head-capital-gptj-hb}), OPT-1.3B (Figures \ref{important-memory-head-capital-opt-1.3b-hb} and \ref{important-context-head-capital-opt-1.3b-hb}), OPT-2.7B (Figures \ref{important-memory-head-capital-opt-2.7b-hb} and \ref{important-context-head-capital-opt-2.7b-hb}), Pythia-6.9B (Figures \ref{important-memory-head-capital-pythia-6.9b-hb} and \ref{important-context-head-capital-pythia-6.9b-hb}), Pythia-12B (Figures \ref{important-memory-head-capital-pythia-12b-hb} and \ref{important-context-head-capital-pythia-12b-hb}), LLaMA2-7B (Figures \ref{important-memory-head-capital-llama2-7b-hb} and \ref{important-context-head-capital-llama2-7b-hb}) and LLaMA2-13B (Figures \ref{important-memory-head-capital-llama2-13b-hb} and \ref{important-context-head-capital-llama2-13b-hb}).
The red squares indicate heads that have a significant positive impact, while the blue squares represent heads that have a negative effect.

\section{Additional Experimental Results}
\label{AExperimental}

We report experimental results in Table \ref{exp1} and \ref{NQ}.

\begin{table*}[t]
\centering
\scalebox{0.85}{
\begin{tabular}{cllcccccccccc}
\toprule
                                 & \multicolumn{2}{l}{}                                  & \multicolumn{2}{c}{\textbf{\texttt{World}}}   & \multicolumn{2}{c}{\textbf{\texttt{World}}}     & \multicolumn{2}{c}{\textbf{\texttt{Official}}} & \multicolumn{2}{c}{}                                   & \multicolumn{2}{c}{}                                     \\
                                 & \multicolumn{2}{l}{}                                  & \multicolumn{2}{c}{\textbf{\texttt{Capital}}} & \multicolumn{2}{c}{\textbf{\texttt{Capital D}}} & \multicolumn{2}{c}{\textbf{\texttt{Language}}} & \multicolumn{2}{c}{\multirow{-2}{*}{\textbf{\texttt{Country}}}} & \multicolumn{2}{c}{\multirow{-2}{*}{ \textbf{\texttt{Continent}}}} \\ \cmidrule{4-13} 

\multirow{-3}{*}{\textbf{Model}} & \multicolumn{2}{c}{\multirow{-3}{*}{\textbf{Method}}} & \textbf{RM}          & \textbf{RC}         & \textbf{RM}           & \textbf{RC}          & \textbf{RM}            & \textbf{RC}           & \textbf{RM}        & \textbf{RC}       &  \textbf{RM}   &  \textbf{RC}  \\ \midrule
                                 \rowcolor[RGB]{246, 246, 246} \cellcolor{white}& \multicolumn{2}{c}{Base}                               & 40.5             & 59.5              & 36.3               & 63.7              & 20.3                 & 79.7                 & 26.8                      & 73.2                      & 19.2                       & 80.8          \\ \cmidrule{2-13} 
                                 \rowcolor[RGB]{255, 231, 231} \cellcolor{white}&     \cellcolor{white}         &              Prompt                & 19.7             & 78.4              & 37.5               & 57.2              & 7.9                  & 91.4                 & 16.9                      & 82.6                      & 7.9                        & 90.1         \\
                                 \rowcolor[RGB]{255, 231, 231} \cellcolor{white}&       \cellcolor{white}  $\uparrow$ Memory                     & Gradient               & 82.7             & 12.2              & 56.4               & 21.2              & 37.1                 & 50.4                 & 38.0                      & 61.1                      & 20.5                       & 55.9                                \\
                                 \rowcolor[RGB]{255, 197, 197}  \cellcolor{white}&       \cellcolor{white}    & PH3 (Ours)                  & \textbf{95.0}    & \textbf{0.2}      & \textbf{87.2}               & \textbf{1.9}               & \textbf{70.3}        & \textbf{16.6}        & \textbf{47.7}                      & \textbf{49.1}                      & \textbf{43.4}                       & \textbf{51.7} \\ \cmidrule{2-13} 
                                 \rowcolor[RGB]{238, 239, 255} \cellcolor{white} OPT-1.3B&       \cellcolor{white}                       & Prompt                 & 17.0             & 81.4              & 38.5               & 57.7              & 9.3                  & 89.5                 & 15.9                      & 83.8                      & 6.3                        & 93.3          \\
                                 \rowcolor[RGB]{238, 239, 255} \cellcolor{white}&              \cellcolor{white}                & CAD                    & 8.1              & 86.5              & 31.6               & 60.0              & \textbf{3.2}         & \textbf{89.6}        & \textbf{0.1}              & \textbf{99.5}             & 5.1                        & 89.1                                \\
                                 \rowcolor[RGB]{238, 239, 255} \cellcolor{white}&         \cellcolor{white} $\uparrow$ Context                   & Gradient               & 22.9             & 73.7              & 35.7               & 63.8              & 12.4                 & 82.5                 & 16.3                      & 82.9                      & 17.6                       & 80.2                                  \\
                                 \rowcolor[RGB]{212, 216, 255} \cellcolor{white}&  \cellcolor{white}  &       PH3 (Ours)                   & \textbf{0.2}     & 97.0              & \textbf{7.9}       & \textbf{69.6}     & 12.8                 & 84.1                 & 4.0                       & 85.4                      & \textbf{1.6}               & 44.1                                 \\
    \rowcolor[RGB]{212, 216, 255}  \cellcolor{white} &    \cellcolor{white}       & \ \ \ \  + Prompt          & 0.4              & \textbf{99.2}     & 10.8               & 68.8              & 9.3                  & 88.3                 & 2.4                       & 92.5                      & 2.5                        & \textbf{92.1}                      \\ \midrule
                              \rowcolor[RGB]{246, 246, 246}  \cellcolor{white} & \multicolumn{2}{c}{Base}                              & 40.2             & 59.8              & 46.6               & 53.4              & 8.8                  & 91.2                 & 26.5                      & 73.5                      & 4.3                        & 95.7                                 \\ \cmidrule{2-13} 
                                \rowcolor[RGB]{255, 231, 231}  \cellcolor{white}&    \cellcolor{white}                           & Prompt                 & 17.7             & 80.3              & 24.2               & 71.6              & 3.4                  & 96.3                 & 12.7                      & 87.2                      & 1.6                        & 98.1                                \\
                                \rowcolor[RGB]{255, 231, 231} \cellcolor{white}&   \cellcolor{white} $\uparrow$ Memory                          & Gradient                & 75.4             & 19.3              & 10.3               & 79.7              & 13.3                 & 57.7                 & 42.9                      & 56.3                      & 5.3                        & 94.1                                \\
                                \rowcolor[RGB]{255, 197, 197} \cellcolor{white} & \cellcolor{white}          & PH3 (Ours)                   & \textbf{93.4}    & \textbf{0.5}      & \textbf{2.8}       & \textbf{87.6}     & \textbf{75.6}        & \textbf{1.5}         & \textbf{56.3}             & \textbf{38.9}             & \textbf{29.3}              & \textbf{55.5}                        \\ \cmidrule{2-13} 
                                \rowcolor[RGB]{238, 239, 255} \cellcolor{white} OPT-2.7B&   \cellcolor{white}                           & Prompt                 & 4.7              & 94.9              & 11.1               & 86.9              & 3.2                  & 96.3                 & 13.2                      & 86.5                      & 0.7                        & 99.0                                \\
                                \rowcolor[RGB]{238, 239, 255} \cellcolor{white}&   \cellcolor{white}                           & CAD                   & 10.8             & 72.2              & 28.8               & 46.3              & 2.7                  & 87.5                 & 9.0                       & 89.1                      & 0.5                        & 99.0                                 \\
                                \rowcolor[RGB]{238, 239, 255} \cellcolor{white}&     \cellcolor{white}  $\uparrow$ Context                       & Gradient               & 36.1             & 62.9              & 43.3               & 53.5              & 7.3                  & 91.6                 & 23.7                      & 76.3                      & 4.1                        & 95.9                                \\
                                 \rowcolor[RGB]{212, 216, 255} \cellcolor{white}&      \cellcolor{white}                        & PH3 (Ours)                  & 1.3              & 97.8              & 3.4                & 81.6              & 4.1                  & 94.9                 & 9.0                       & 90.9                      & 0.9                        & 98.6                                \\
\rowcolor[RGB]{212, 216, 255}     \cellcolor{white}     &    \cellcolor{white}       & \ \ \ \  + Prompt         & \textbf{0.8}     & \textbf{98.3}     & \textbf{1.3}       & \textbf{94.6}     & \textbf{1.5}         & \textbf{98.2}        & \textbf{6.9}              & \textbf{93.1}             & \textbf{0.0}               & \textbf{99.5}                     \\ 
\midrule
                                \rowcolor[RGB]{246, 246, 246} \cellcolor{white} & \multicolumn{2}{c}{Base}                               & 53.3             & 46.6              & 74.8               & 25.2              & 49.7                 & 50.3                 & 41.3                      & 58.7                      & 39.1                       & 60.9          \\ \cmidrule{2-13} 
                                \rowcolor[RGB]{255, 231, 231} \cellcolor{white}&      \cellcolor{white}        &              Prompt                & 44.7             & 51.2              & 41.8               & 37.8              & 16.5                 & 81.4                 & 12.8                      & 87.1                      & 36.2                       & 61.6          \\
                                \rowcolor[RGB]{255, 231, 231} \cellcolor{white}& \cellcolor{white}    $\uparrow$ Memory                          & Gradient               & 56.5             & 35.8              & 72.8               & 22.4              & 56.7                 & 36.3                 & 37.9                      & 62.1                      & 40.1                       & 58.9                                  \\
                                 \rowcolor[RGB]{255, 197, 197} \cellcolor{white}&   \cellcolor{white}       & PH3 (Ours)                  & \textbf{90.2}    & \textbf{6.7}      & \textbf{88.4}      & \textbf{10.2}     & \textbf{71.5}        & \textbf{11.1}        & \textbf{41.8}             & \textbf{57.3}             & \textbf{66.0}              & \textbf{31.0}   \\ \cmidrule{2-13} 
                                \rowcolor[RGB]{238, 239, 255} \cellcolor{white} Pythia-6.9B&                   \cellcolor{white}           & Prompt                 & 32.7             & 63.7              & 32.0               & 44.8              & 8.9                  & 90.5                 & 10.4                      & 89.5                      & 31.7                       & 75.6          \\
                                \rowcolor[RGB]{238, 239, 255} \cellcolor{white}&              \cellcolor{white}                & CAD                    & 14.3             & 55.1              & \textbf{22.0}      & 27.6              & \textbf{3.3}         & 78.1                 & 8.5                       & 91.2                      & 12.3                       & 82.2                               \\
                                \rowcolor[RGB]{238, 239, 255} \cellcolor{white}&                  \cellcolor{white}  $\uparrow$ Context         & Gradient               & 41.4             & 53.7              & 61.8               & 35.6              & 48.2                 & 51.3                 & 34.3                      & 65.6                      & 41.7                       & 53.0                                  \\
                                 \rowcolor[RGB]{212, 216, 255} \cellcolor{white}&   \cellcolor{white}                           & PH3 (Ours)                  & 6.7              & 81.7              & 34.4               & 30.0              & 16.4                 & 70.0                 & 3.4                       & 96.3                      & 3.3                        & 94.5                               \\
\rowcolor[RGB]{212, 216, 255}  \cellcolor{white}     & \cellcolor{white}          & \ \ \ \  + Prompt          & \textbf{0.6}     & \textbf{98.6}     & 24.4               & \textbf{60.4}     & \textbf{3.3}         & \textbf{95.6}        & \textbf{0.3}              & \textbf{99.5}             & \textbf{0.7}               & \textbf{98.8}                       \\ 
\midrule
                                \rowcolor[RGB]{246, 246, 246}  \cellcolor{white}& \multicolumn{2}{c}{Base}                              & 59.7             & 40.3              & 64.6               & 35.4              & 34.3              & 65.7              & 35.0                      & 65.0                      & 43.0                       & 57.0              \\ \cmidrule{2-13} 
                                 \rowcolor[RGB]{255, 231, 231} \cellcolor{white}&       \cellcolor{white}       &              Prompt                & 5.8              & 94.1              & 43.7               & 53.3              & 11.1              & 85.8              & 2.4                       & 97.5                      & 10.1                       & 88.5        \\
                                \rowcolor[RGB]{255, 231, 231} \cellcolor{white}&               \cellcolor{white}   $\uparrow$ Memory            & Gradient               & 62.9             & 27.7              & 63.1               & 30.2              & 39.3              & 37.5              & 46.2                      & 53.3                      & 42.1                       & 56.9                                 \\
                                 \rowcolor[RGB]{255, 197, 197} \cellcolor{white}&    \cellcolor{white}       &  PH3 (Ours)                  & \textbf{95.0}    & \textbf{0.6}      & \textbf{82.4}      & \textbf{2.2}      & \textbf{69.9}     & \textbf{6.9}      & \textbf{57.1}             & \textbf{35.9}             & \textbf{70.1}              & \textbf{9.6}    \\ \cmidrule{2-13} 
                                \rowcolor[RGB]{238, 239, 255} \cellcolor{white} Pythia-12B&                 \cellcolor{white}             & Prompt                 & 6.2              & 93.6              & 34.1               & 62.0              & 21.1              & 77.6              & \textbf{1.7}              & \textbf{98.3}             & 6.3                        & 92.5           \\
                                \rowcolor[RGB]{238, 239, 255} \cellcolor{white}&             \cellcolor{white}                 & CAD                   & 3.1              & 65.7              & 13.6               & 42.9              & 2.0               & 89.4              & 3.7                       & 94.7                      & 11.3                       & 80.0                                \\
                                \rowcolor[RGB]{238, 239, 255} \cellcolor{white}&             \cellcolor{white}    $\uparrow$ Context             & Gradient               & 59.3             & 19.6              & 33.3               & 25.0              & 21.3              & 54.6              & 28.5                      & 71.3                      & 40.2                       & 52.1                                 \\
                                \rowcolor[RGB]{212, 216, 255} \cellcolor{white}&                \cellcolor{white}              & PH3 (Ours)                   & 18.6             & 76.1              & 56.9               & 33.3              & 10.9              & 80.4              & 16.9                      & 76.7                      & 26.9                       & 67.9                                 \\
\rowcolor[RGB]{212, 216, 255}    \cellcolor{white}    &    \cellcolor{white}        & \ \ \ \  + Prompt          & \textbf{1.9}     & \textbf{97.6}     & \textbf{17.7}      & \textbf{75.8}     & \textbf{3.8}      & \textbf{95.5}     & 2.2                       & 97.7                      & \textbf{2.4}               & \textbf{97.0}                           \\ 

\midrule
                                 \rowcolor[RGB]{246, 246, 246} \cellcolor{white} &  \multicolumn{2}{c}{Base}                             & 60.6             & 39.4              & 74.6               & 25.0              & 1.6               & 98.4              & 26.2                      & 73.7                      & 5.6                        & 93.3                                     \\ \cmidrule{2-13} 
                                \rowcolor[RGB]{255, 231, 231} \cellcolor{white}&                \cellcolor{white}              &  Prompt                 & 0.4              & 99.6              & 78.5               & 21.0              & 0.5               & 99.5              & 22.3                      & 77.6                      & 10.7                       & 89.3                               \\
                                \rowcolor[RGB]{255, 231, 231} \cellcolor{white}&           \cellcolor{white}    $\uparrow$ Memory             & Gradient              & 77.7             & 10.0              & 89.9               & 9.5               & 26.1              & 68.3              & \textbf{48.1}             & \textbf{51.4}             & \textbf{30.7}              & 49.0                                  \\
                                 \rowcolor[RGB]{255, 197, 197} \cellcolor{white} &       \cellcolor{white}     &  PH3 (Ours)                   & \textbf{86.3}    & \textbf{12.5}     & \textbf{91.5}      & \textbf{8.1}      & \textbf{71.2}     & \textbf{11.6}     & 47.7                      & 51.6                      & 11.9                       & \textbf{45.4}                                \\ \cmidrule{2-13} 
                                \rowcolor[RGB]{238, 239, 255} \cellcolor{white} LLaMA2-13B &            \cellcolor{white}                   & Prompt                & \textbf{0.0}     & \textbf{100.0}    & 70.9               & 28.7              & \textbf{0.0}      & \textbf{100.0}    & 7.1                       & 92.7                      & 3.6                        & 96.4                               \\
                                \rowcolor[RGB]{238, 239, 255} \cellcolor{white}&                   \cellcolor{white}           & CAD                    & 6.2              & 91.0              & \textbf{1.1}       & \textbf{98.8}     & \textbf{0.0}      & \textbf{100.0}    & 0.5                       & 99.5                      & \textbf{0.0}               & 99.6                                \\
                                \rowcolor[RGB]{238, 239, 255} \cellcolor{white}&             \cellcolor{white}    $\uparrow$ Context             & Gradient               & 46.3             & 33.9              & 74.1               & 25.4              & 16.5              & 83.3              & 20.2                      & 79.5                      & 3.6                        & 96.4                               \\
                                \rowcolor[RGB]{212, 216, 255} \cellcolor{white}&              \cellcolor{white}                & PH3 (Ours)                    & 6.2              & 92.1              & 24.1               & 75.9              & 1.3               & 98.7              & 5.6                       & 94.4                      & 0.4                        & 99.6                                \\
\rowcolor[RGB]{212, 216, 255}    \cellcolor{white}  &     \cellcolor{white}      & \ \ \ \  + Prompt         & \textbf{0.0}     & \textbf{100.0}    & 31.0               & 68.0              & \textbf{0.0}      & \textbf{100.0}    & \textbf{0.0}              & \textbf{100.0}            & \textbf{0.0}               & \textbf{100.0}                                      \\ \bottomrule
\end{tabular}
}
\caption{Experimental results of OPT-1.3B, OPT-2.7B, Pythia-6.9B, Pythia-12B and LLaMA2-7B on five datasets. Bolds denote the best results.}
\label{exp1}
\end{table*}

\begin{table}[t]
\centering
\scalebox{0.95}{
\begin{tabular}{lcccc}
\toprule
\textbf{Method}    & \textbf{GPT-2 XL} & \textbf{GPT-J} & \textbf{OPT-2.7B} \\ \midrule
Base               & 45.6              & 54.8           & 51.4                          \\
Prompt             & 47.6              & 57.4           & 54.1                           \\
CAD                & 44.5              & 55.0           & 50.2                           \\
Gradient           & 45.3              & 55.0           & 50.8                           \\ \midrule
\rowcolor[RGB]{212, 216, 255}PH3 ($k=1$)          & 47.1              & 57.5           & 53.5                           \\
\rowcolor[RGB]{212, 216, 255}PH3 ($k=3$)          & 52.2              & 58.6           & 55.4                           \\
\rowcolor[RGB]{212, 216, 255}\ \ \ \ + Prompt & \textbf{54.0}              & \textbf{59.6}           & \textbf{56.7}                          \\ 

\rowcolor[RGB]{212, 216, 255}PH3 ($k=5$)          & 49.4              & 56.3           & 54.0                       \\ \bottomrule
\end{tabular}
}
\caption{Experimental results (Recall) of GPT-2 XL, GPT-J and OPT-2.7B on the NQ dataset. Bolds denote the best results.}
\label{NQ}
\end{table}

\section{Number of Pruning Heads}
\label{ANumber}
\begin{figure}[h]

    \centering
    \includegraphics[width=0.49\textwidth]{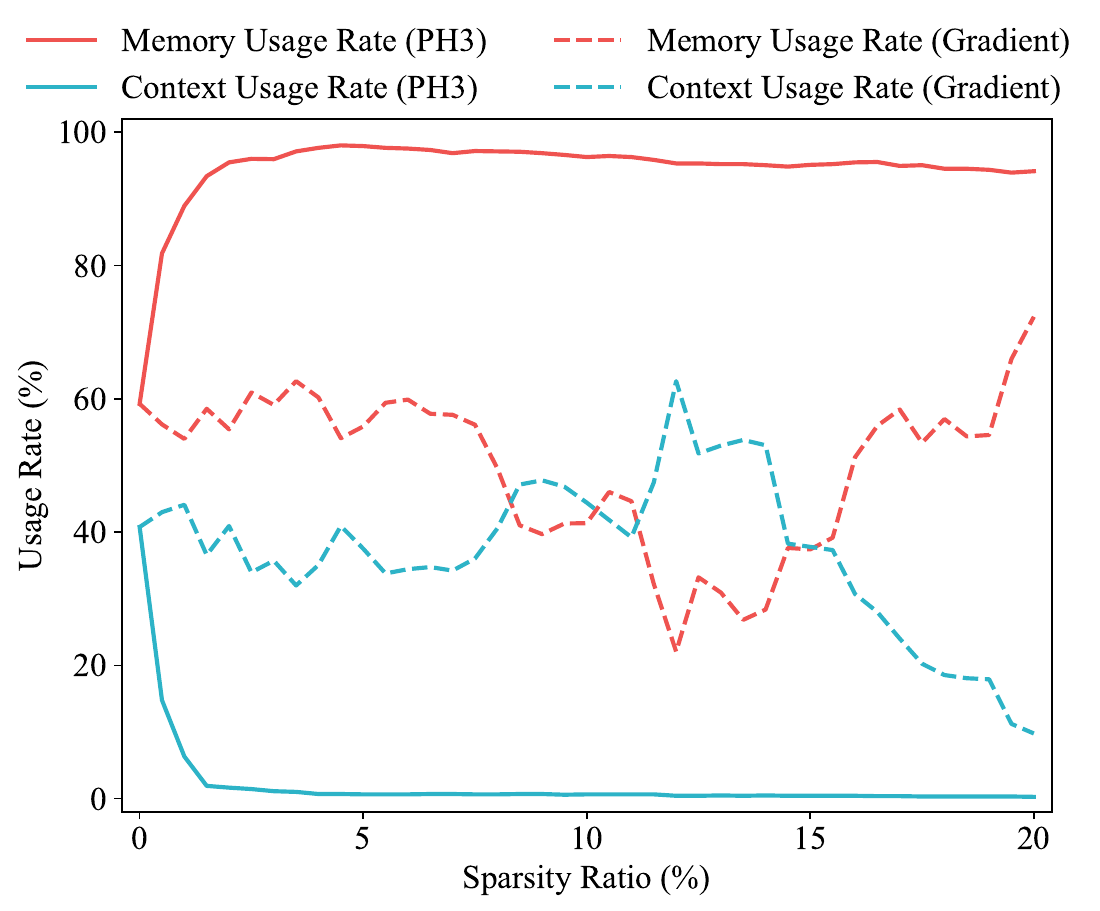}
    \caption{Impact of GPT-2 XL's sparsity ratio on improving internal memory usage rate.}
    \label{ratio-memory-gpt2-xl}

\end{figure}

\begin{figure}[h]

    \centering
    \includegraphics[width=0.49\textwidth]{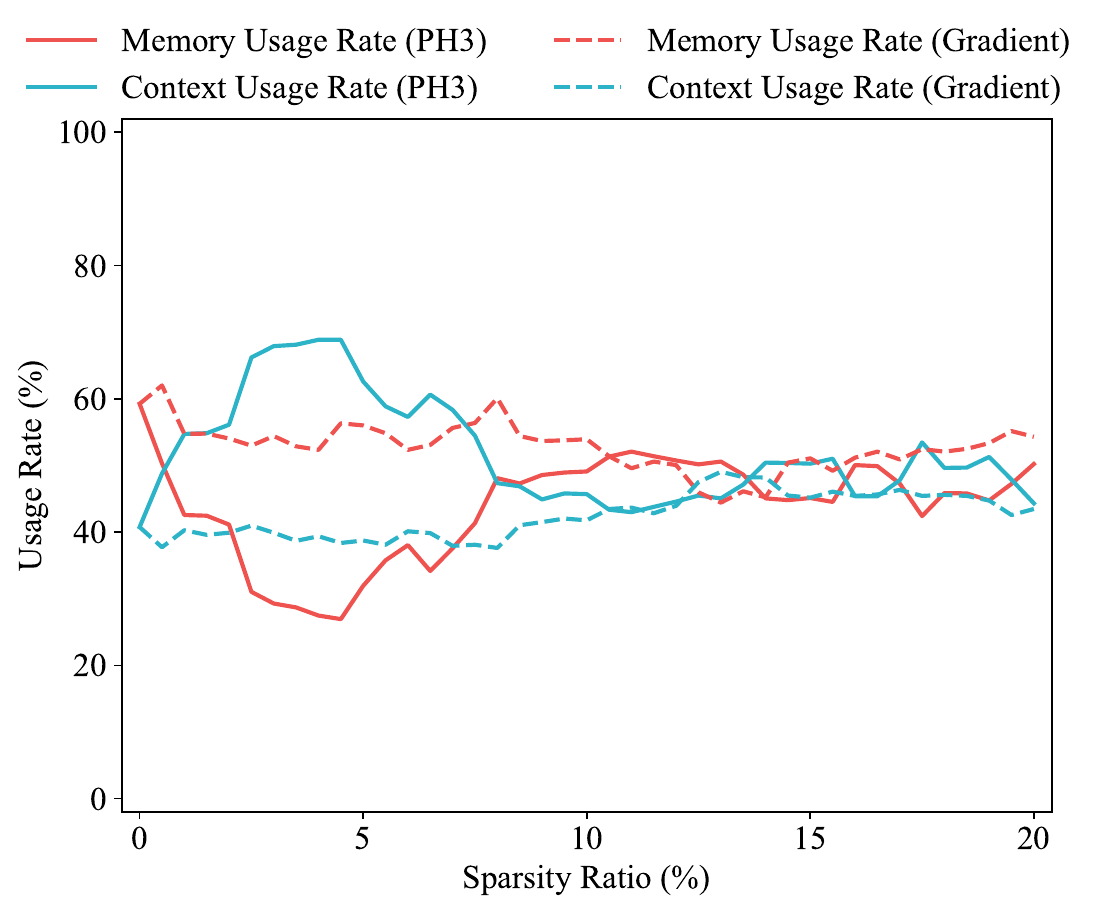}
    \caption{Impact of GPT-2 XL's sparsity ratio on improving external context usage rate.}
    \label{ratio-context-gpt2-xl}

\end{figure}

\begin{figure}[h]

    \centering
    \includegraphics[width=0.49\textwidth]{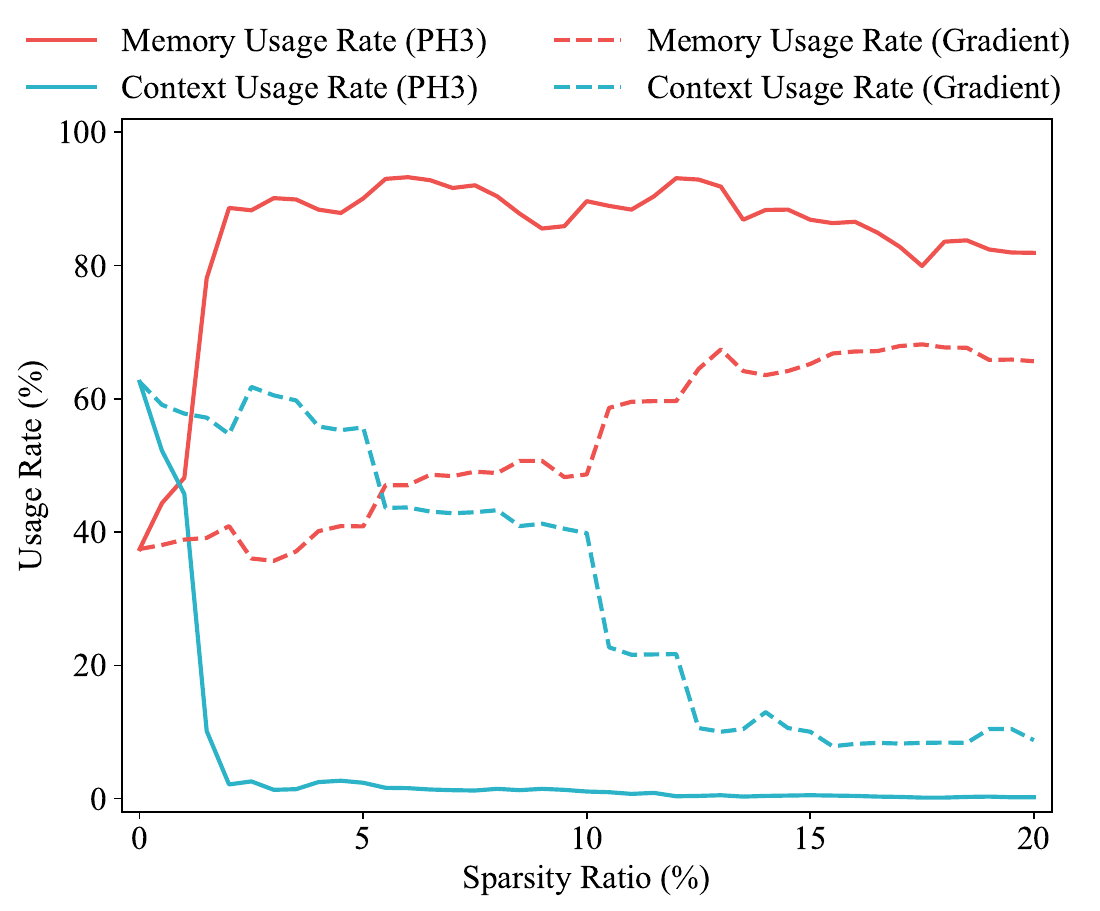}
    \caption{Impact of GPT-J's sparsity ratio on improving internal memory usage rate.}
    \label{ratio-memory-gptj}

\end{figure}

\begin{figure}[h]

    \centering
    \includegraphics[width=0.49\textwidth]{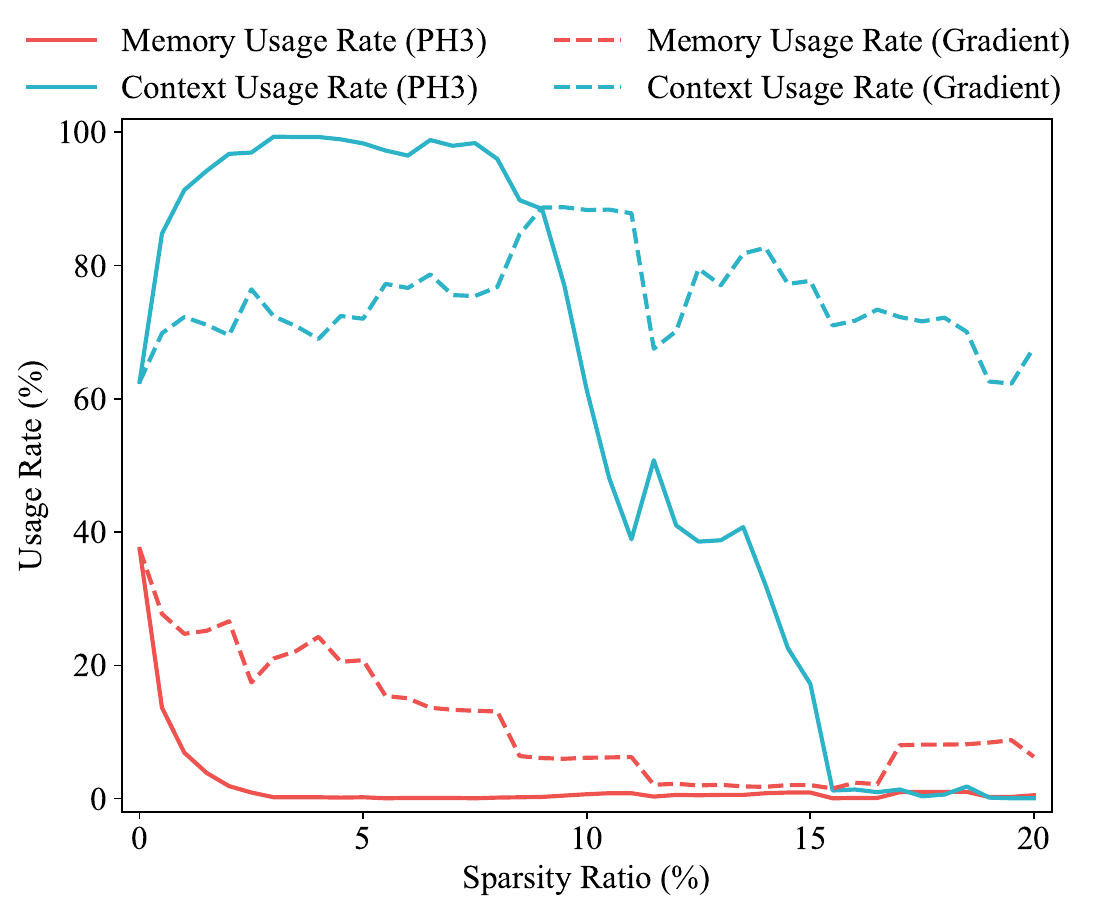}
    \caption{Impact of GPT-J's sparsity ratio on improving external context usage rate.}
    \label{ratio-context-gptj}

\end{figure}

\begin{figure}[h]

    \centering
    \includegraphics[width=0.49\textwidth]{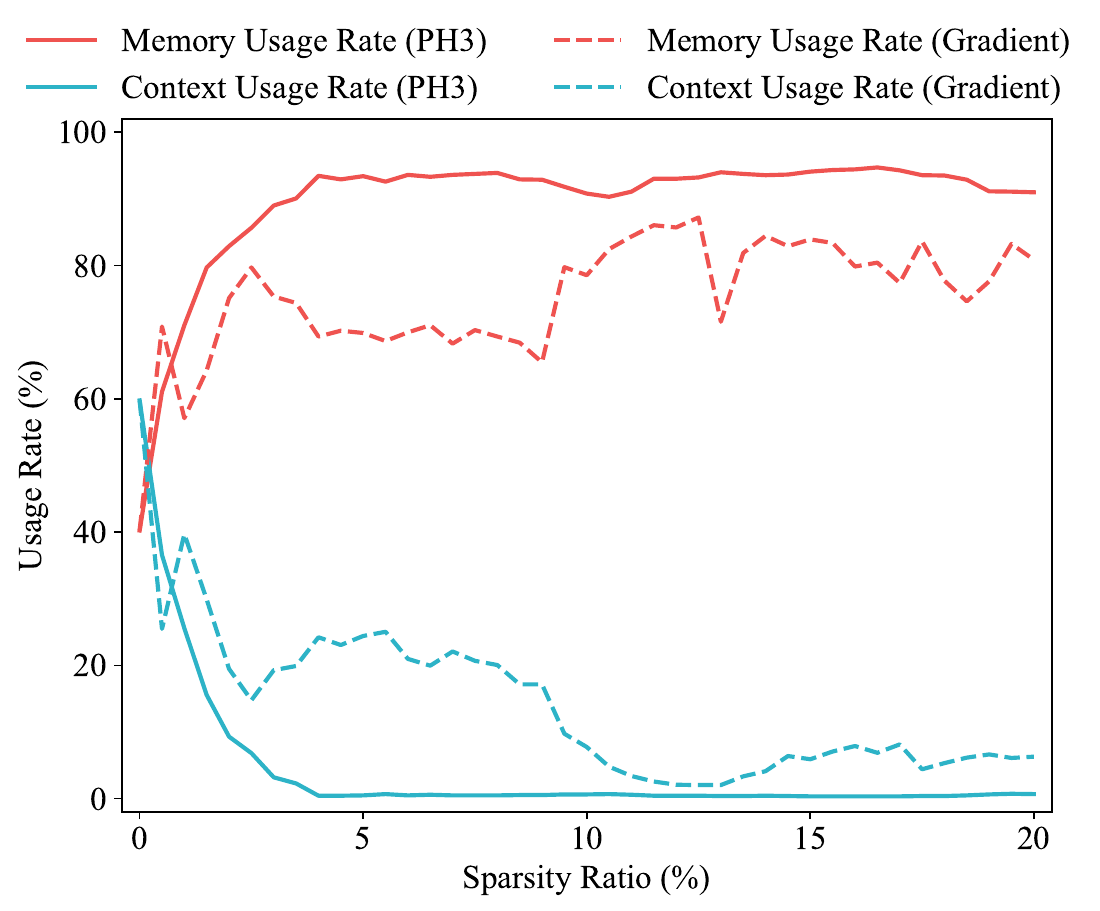}
    \caption{Impact of OPT-2.7B's sparsity ratio on improving internal memory usage rate.}
    \label{ratio-memory-opt-2.7b}

\end{figure}

\begin{figure}[h]

    \centering
    \includegraphics[width=0.49\textwidth]{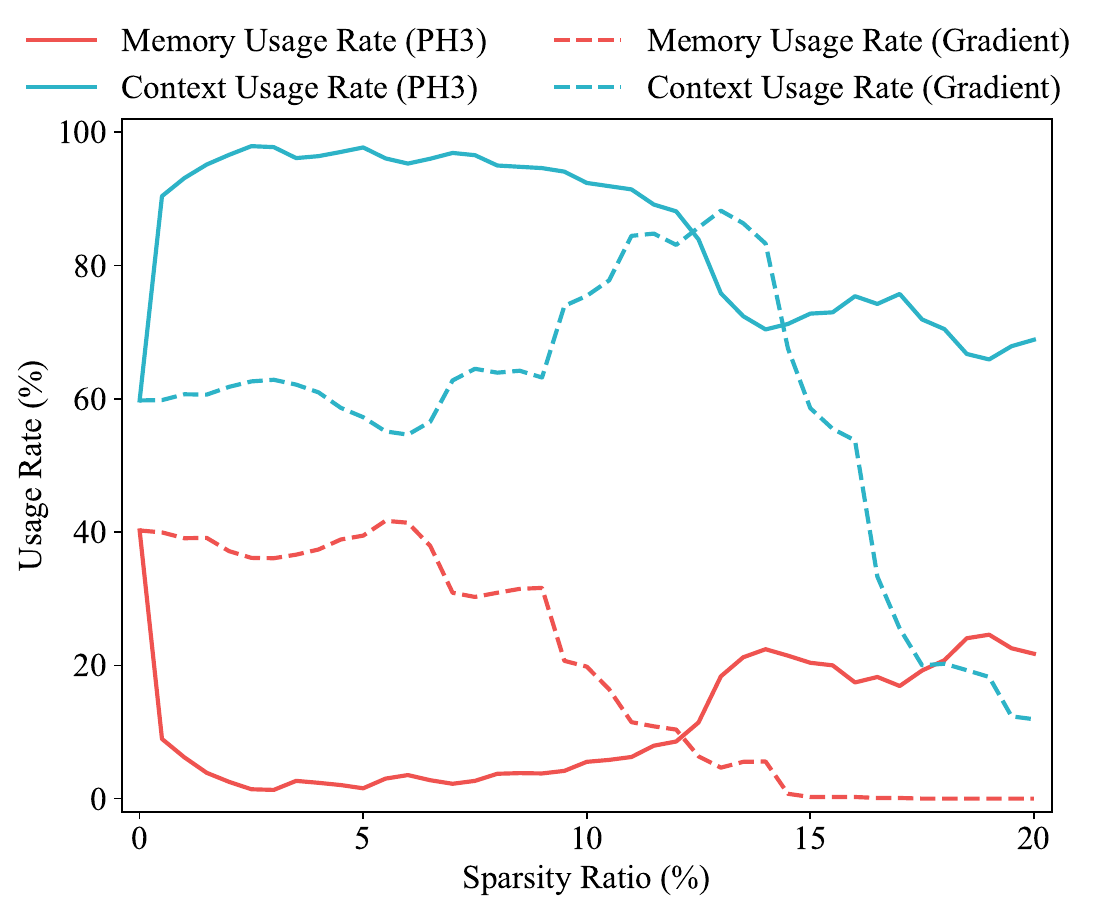}
    \caption{Impact of OPT-2.7B's sparsity ratio on improving external context usage rate.}
    \label{ratio-context-opt-2.7b}

\end{figure}

\begin{figure}[h]

    \centering
    \includegraphics[width=0.49\textwidth]{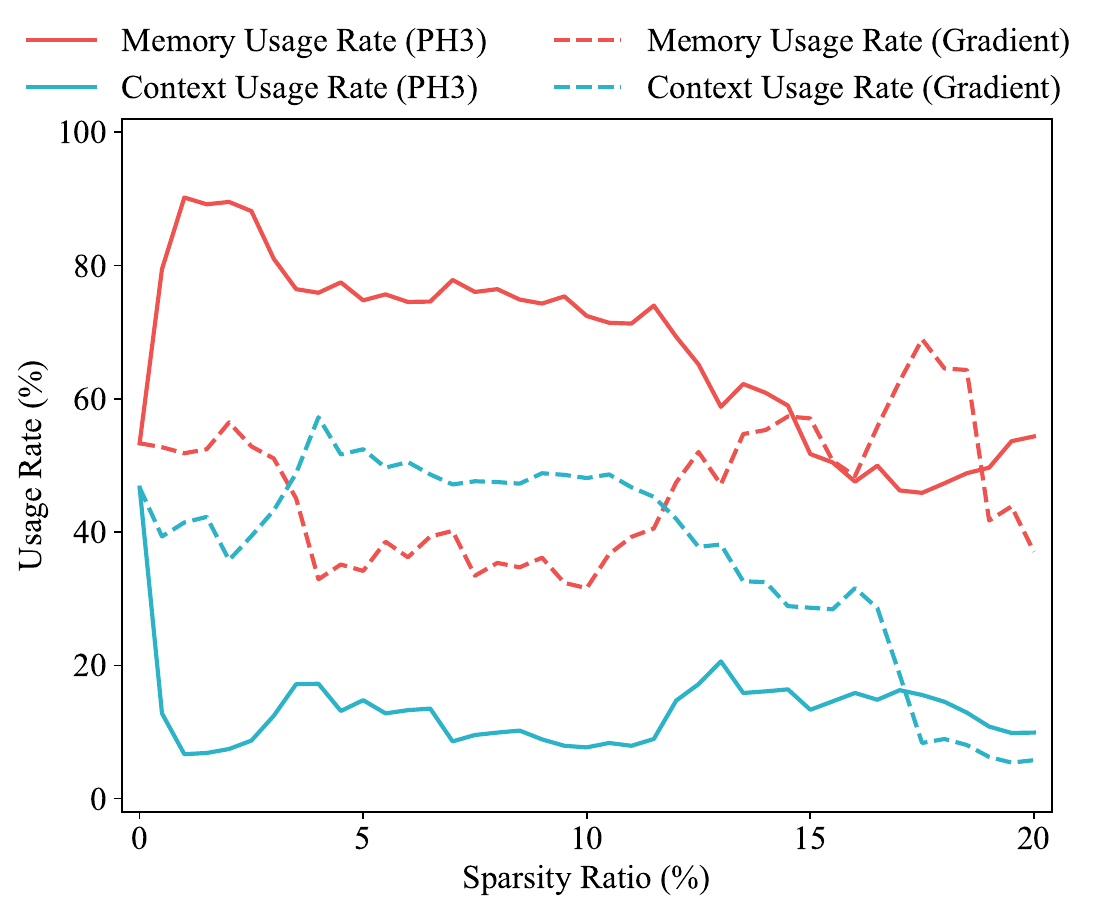}
    \caption{Impact of Pythia-6.9B's sparsity ratio on improving internal memory usage rate.}
    \label{ratio-memory-pythia-6.9b}

\end{figure}

\begin{figure}[h]

    \centering
    \includegraphics[width=0.49\textwidth]{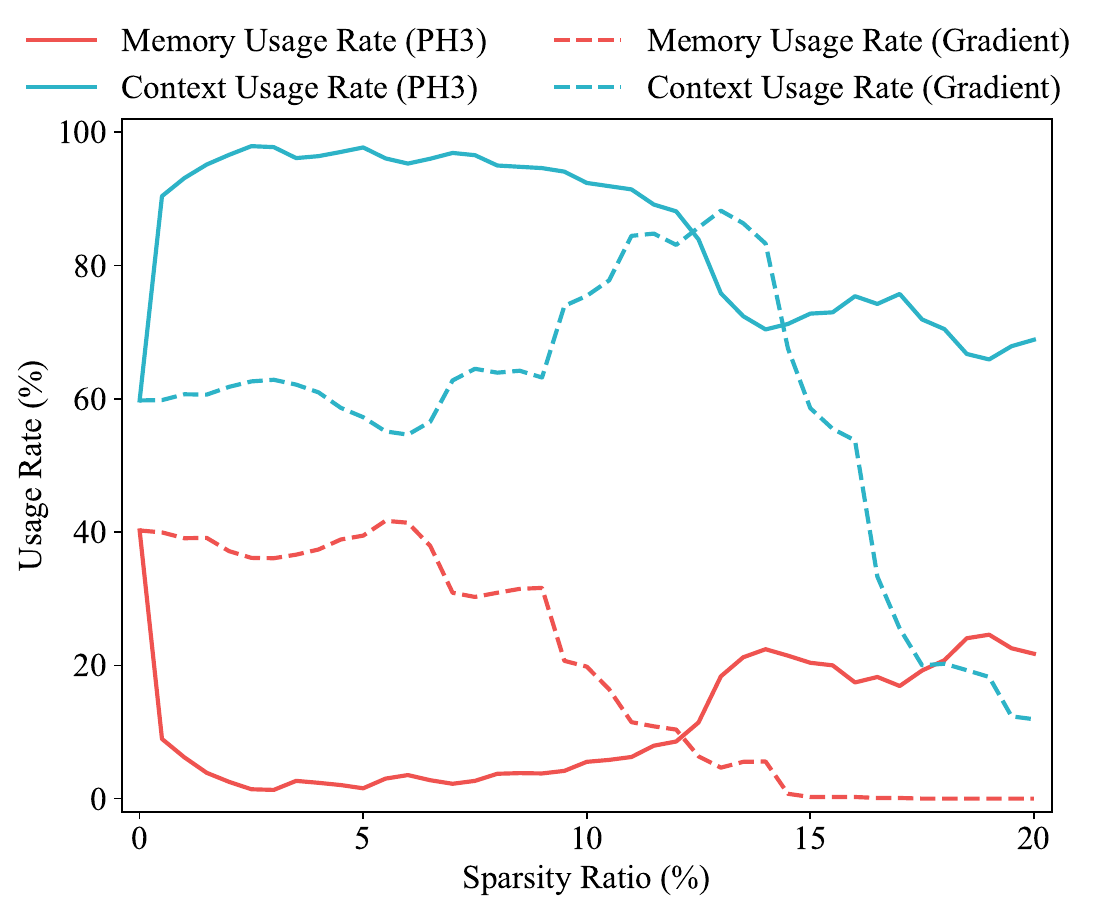}
    \caption{Impact of Pythia-6.9B's sparsity ratio on improving external context usage rate.}
    \label{ratio-context-pythia-6.9b}

\end{figure}

\begin{figure}[h]

    \centering
    \includegraphics[width=0.49\textwidth]{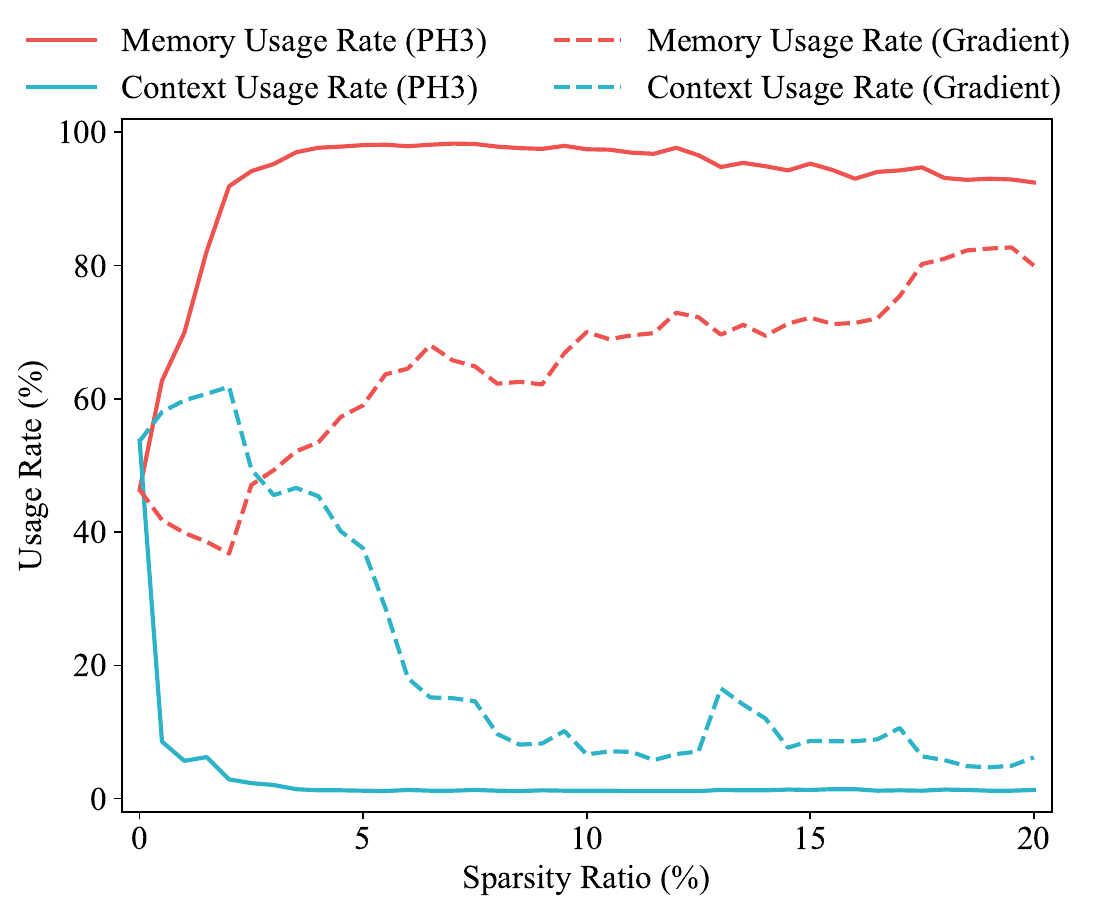}
    \caption{Impact of LLaMA2-7B's sparsity ratio on improving internal memory usage rate.}
    \label{ratio-memory-llama2-7b}

\end{figure}

\begin{figure}[h]

    \centering
    \includegraphics[width=0.49\textwidth]{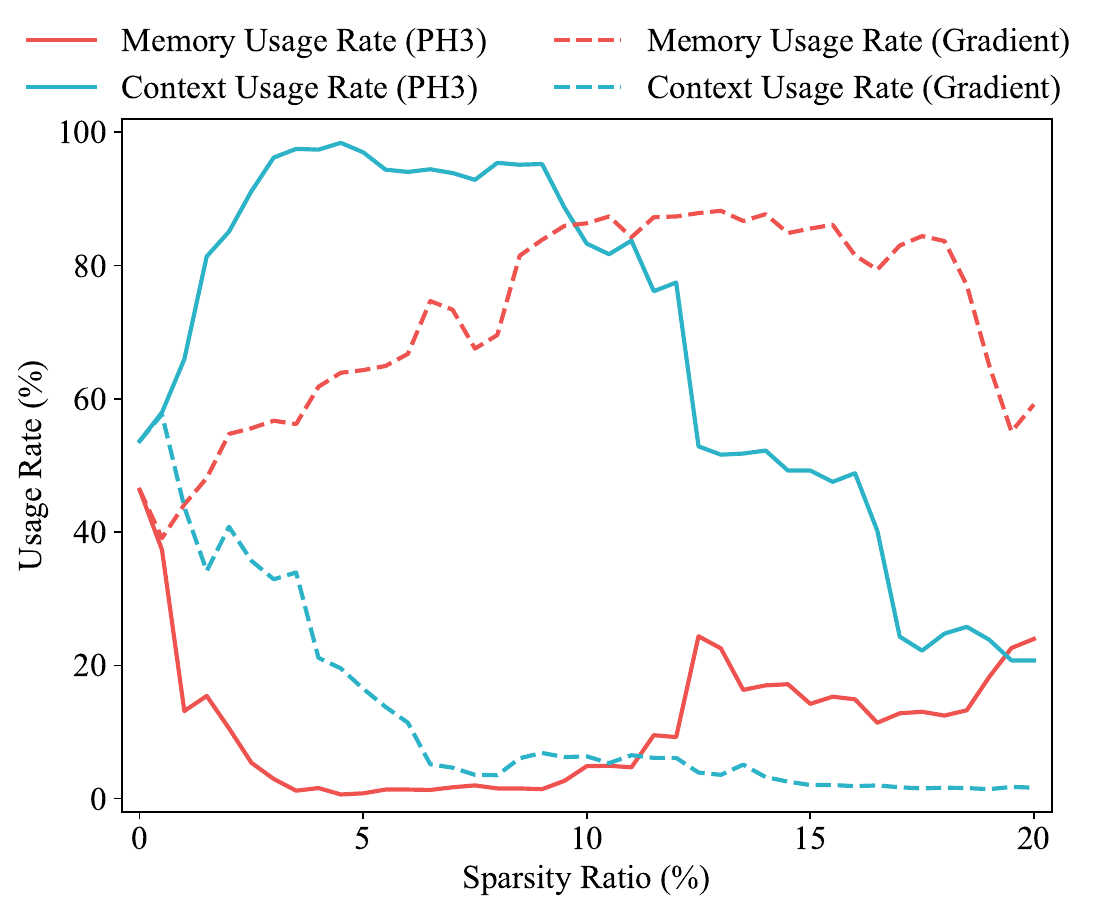}
    \caption{Impact of LLaMA2-7B's sparsity ratio on improving external context usage rate.}
    \label{ratio-context-llama2-7b}

\end{figure}

\begin{figure}[h]

    \centering
    \includegraphics[width=0.49\textwidth]{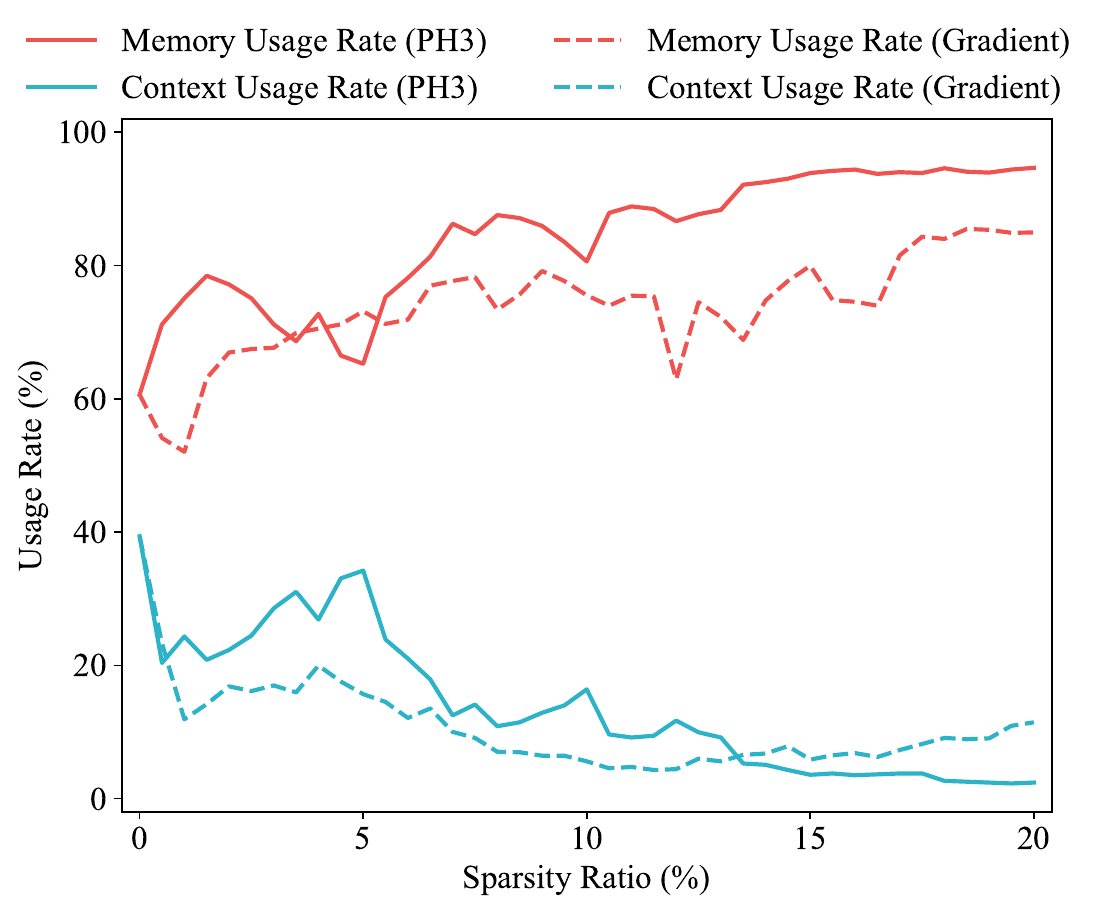}
    \caption{Impact of LLaMA2-13B's sparsity ratio on improving internal memory usage rate.}
    \label{ratio-memory-llama2-13b}

\end{figure}

\begin{figure}[h]

    \centering
    \includegraphics[width=0.49\textwidth]{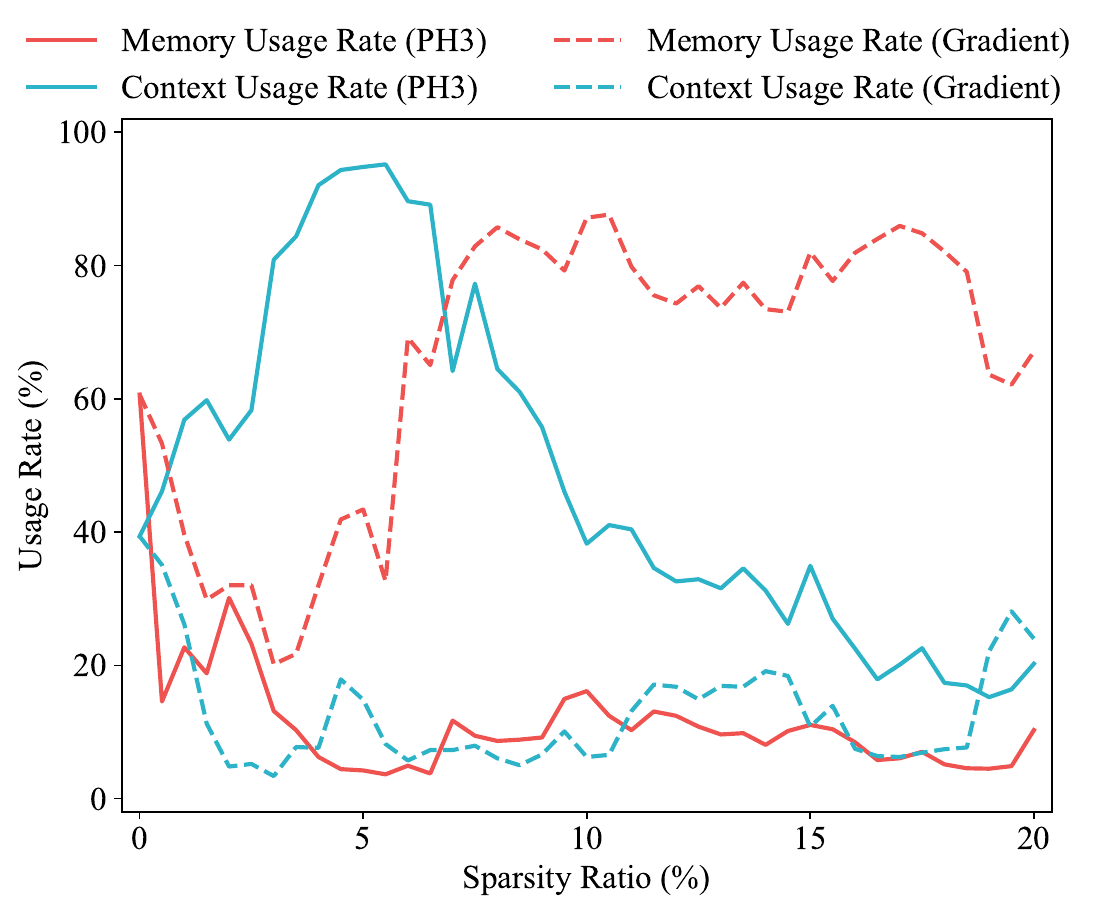}
    \caption{Impact of LLaMA2-13B's sparsity ratio on improving external context usage rate.}
    \label{ratio-context-llama2-13b}

\end{figure}

As shown in Figures \ref{ratio-memory-gpt2-xl}, \ref{ratio-context-gpt2-xl}, \ref{ratio-memory-gptj}, \ref{ratio-context-gptj}, \ref{ratio-memory-opt-2.7b}, \ref{ratio-context-opt-2.7b}, \ref{ratio-memory-pythia-6.9b}, \ref{ratio-context-pythia-6.9b}, \ref{ratio-memory-llama2-7b}, \ref{ratio-context-llama2-7b}, \ref{ratio-memory-llama2-13b} and \ref{ratio-context-llama2-13b}, we analyze the impact of the number of pruning heads (sparsity ratio) on the Gradient and PH3 methods.

\begin{table}[t]
\centering
\begin{tabular}{lcc}
\toprule
\textbf{Model} & \textbf{\#Layer} $L$ & \textbf{\#Head} $M$ \\ \midrule
GPT-2 XL       & 48               & 25              \\
GPT-J          & 28               & 16              \\
OPT-1.3B       & 24               & 32              \\
OPT-2.7B       & 32               & 32              \\
Pythia-6.9B    & 32               & 32              \\
Pythia-12B     & 36               & 40              \\
LLaMA2-7B      & 32               & 32              \\
LLaMA2-13B     & 40               & 40              \\ \bottomrule
\end{tabular}
\caption{Model details.}
\label{model details}

\end{table}

\end{document}